\documentclass[lettersize,journal]{IEEEtran}
\usepackage{amsmath,amsfonts}
\usepackage{array}
\usepackage[caption=false,font=normalsize,labelfont=sf,textfont=sf]{subfig}
\usepackage{textcomp}
\usepackage{stfloats}
\usepackage{url}
\usepackage{verbatim}
\usepackage{graphicx}
\usepackage{cite}
\usepackage{xspace}
\usepackage{algorithm}
\usepackage{algpseudocode}
\usepackage{amsmath}
\usepackage{multirow}
\usepackage{adjustbox}
\usepackage{threeparttable}
\usepackage{authblk}
\usepackage{booktabs}
\usepackage{stackengine}
\usepackage{xcolor}
\usepackage[normalem]{ulem}
\useunder{\uline}{\ul}{}
\usepackage{bbding}
\usepackage{utfsym}
\usepackage{fontawesome}

\makeatletter
\DeclareRobustCommand\onedot{\futurelet\@let@token\@onedot}
\def\@onedot{\ifx\@let@token.\else.\null\fi\xspace}

\def\etal{\emph{et al}\onedot}
\makeatother
\begin{document}

\title{Dual-Hybrid Attention Network for Specular Highlight Removal}

\author[13*]{Xiaojiao Guo}
\author[124*]{Xuhang Chen\thanks{* These authors contribute equally to this work.}}
\author[1]{Shenghong Luo}
\author[2$^\dag$]{Shuqiang Wang\thanks{$^\dag$ Corresponding authors.}}
\author[1$^\dag$]{Chi-Man Pun}

\affil[1]{University of Macau}
\affil[2]{Shenzhen Institutes of Advanced Technology, Chinese Academy of Sciences}
\affil[3]{Baoshan University}
\affil[4]{Huizhou University}


\maketitle

\begin{abstract}
Specular highlight removal plays a pivotal role in multimedia applications, as it enhances the quality and interpretability of images and videos, ultimately improving the performance of downstream tasks such as content-based retrieval, object recognition, and scene understanding. Despite significant advances in deep learning-based methods, current state-of-the-art approaches often rely on additional priors or supervision, limiting their practicality and generalization capability. In this paper, we propose the Dual-Hybrid Attention Network for Specular Highlight Removal (DHAN-SHR), an end-to-end network that introduces novel hybrid attention mechanisms to effectively capture and process information across different scales and domains without relying on additional priors or supervision. DHAN-SHR consists of two key components: the Adaptive Local Hybrid-Domain Dual Attention Transformer (L-HD-DAT) and the Adaptive Global Dual Attention Transformer (G-DAT). The L-HD-DAT captures local inter-channel and inter-pixel dependencies while incorporating spectral domain features, enabling the network to effectively model the complex interactions between specular highlights and the underlying surface properties. The G-DAT models global inter-channel relationships and long-distance pixel dependencies, allowing the network to propagate contextual information across the entire image and generate more coherent and consistent highlight-free results. To evaluate the performance of DHAN-SHR and facilitate future research in this area, we compile a large-scale benchmark dataset comprising a diverse range of images with varying levels of specular highlights. Through extensive experiments, we demonstrate that DHAN-SHR outperforms 18 state-of-the-art methods both quantitatively and qualitatively, setting a new standard for specular highlight removal in multimedia applications. The code and dataset will be available.
\end{abstract}


\section{Introduction}
\IEEEPARstart{S}{pecular} highlights, the intense reflections of light sources on shiny surfaces, pose significant challenges in multimedia and computer vision applications. These reflections disrupt the visual consistency of images and videos, obscuring details and altering color fidelity, which can have a detrimental impact on various multimedia applications, such as video editing, content-based retrieval, and interactive media. Removing specular highlights is crucial for accurate image and video interpretation and processing, yet it remains a complex task due to the variability in light conditions, surface properties, and angles of observation.

Traditional highlight removal techniques, based on models such as the dichromatic reflection model~\cite{shafer1985using}, often fall short in diverse real-world scenarios. Deep learning approaches have shown promise, but most require prior information, such as highlight masks, limiting their practicality and generalization ability. Current methods also struggle to effectively restore highlighted areas, leading to suboptimal results.

To address these challenges, we propose an end-to-end network that eliminates the need for additional priors or supervision, enabling specular highlight removal in a single step while preserving the visual fidelity of the restored diffuse image. Our approach is inspired by the observation that in natural scene photos, only a small portion of the surface produces a highlight effect due to its smoothness and specific reflection angle with the light source, while the majority of the remaining areas are in a diffuse state without highlights. Consequently, for the task of highlight removal, we aim to learn a global relationship on illumination and color. Moreover, to effectively restore details such as texture in the highlight regions, it is crucial to consider local features in addition to the global context. Therefore, in our network design, we incorporate two dimensions of self-attention to capture both global-level dependencies and local-level relationships.

Specifically, to capture the relationships between local features, we introduce the Adaptive Local Hybrid-Domain Dual Attention Transformer, which sets our approach apart from current deep learning dehighlight methods that solely concentrate on the spatial domain (pixel values). Our transformer leverages frequency domain features to aid the learning process of the spatial domain, as frequency domain analysis offers a complementary perspective by uncovering both fine details and overarching patterns.
Building upon this hybrid domain foundation, our transformer employs a window-based dual attention mechanism to efficiently focus on the local inter-channel and inter-pixel relationships within the partitioned windows, respectively. Both attention mechanisms' computations are limited to non-overlapping windows, significantly reducing the computational complexity compared to traditional self-attention mechanisms that consider the entire image. This window-based approach allows for more focused processing of local regions, enabling the transformer to effectively capture fine-grained details and textures.

To further enhance the transformer's ability to capture dependencies across window boundaries, we incorporate a window-shifting mechanism which is inspired by Swin
Transformer~\cite{liu2021swin} and SwinIR~\cite{liang2021swinir}. By shifting the pixels to create new window partitions, our module facilitates information exchange between adjacent windows. This shifting operation allows the transformer to capture local dependencies that span across window boundaries, ensuring a seamless integration of features across the entire image.

For global-level dependencies, we propose the Channel-Wise Contextual Attention Module that employs efficient Transformers to capture inter-channel relationships, rather than inter-patch relationships. This design choice serves two purposes: firstly, it allows the network to focus on more coarse-grained whole features, and secondly, it avoids the high computational and memory demands associated with the original Vision Transformer~\cite{dosovitskiy2020vit}. By attending to the channel-wise context, our module can effectively capture the global dependencies between different feature maps, enabling the network to reason about the overall illumination and color distribution.

To better organize feature learning at different scales, we adopt a network architecture similar to UNet, strategically placing modules that focus on different granularities at various positions within the network. Modules that capture detailed information are placed at higher levels, while modules that focus on global information are positioned at lower levels. This hierarchical arrangement enables our network to effectively learn and process features at different scales, improving its ability to handle complex specular highlight removal tasks.

To provide a standardized basis for comparison and yield meaningful insights into the performance improvements of specular highlight removal methods, we assembled an extensive dataset by combining images from three different highlight removal datasets (PSD \cite{wu2021single}, SHIQ \cite{fu2021multi} and SSHR \cite{fu2023towards}). PSD features high-quality real-world ground truth images, while the other two datasets include generated reference ground truths and fully synthetic data, broadening the diversity of the training samples. We retrained seven state-of-the-art deep learning specular highlight removal methods on this unified benchmark and evaluated their performance, along with 11 traditional methods, on the test sets of the benchmark. The experimental results demonstrate that our approach outperforms 18 other state-of-the-art methods across various test datasets and metrics, showcasing its superiority in specular highlight removal.

Overall, our contributions can be summarized as follows:
\begin{itemize}
	\item We propose the Dual-Hybrid Attention Network for Specular Highlight Removal (DHAN-SHR), an end-to-end specular highlight removal network that introduces novel hybrid attention mechanisms, including the Adaptive Local Hybrid-Domain Dual Attention and the Adaptive Global Dual Attention. These attention mechanisms enable DHAN-SHR to effectively and efficiently capture both spatial and spectral information, as well as contextual relationships at different scales, accurately removing specular highlights while restoring underlying diffuse components.
	\item We compile a comprehensive benchmark dataset for specular highlight removal by combining images from three different datasets, resulting in 29,306 training pairs and 2,947 testing pairs. We retrain and test 18 state-of-the-art methods on this new benchmark, conducting a thorough comparative analysis and laying a solid foundation for future advancements in the field.
	\item Extensive experiments and evaluations demonstrate that our proposed DHAN-SHR outperforms state-of-the-art methods both quantitatively and qualitatively, setting a new standard in the realm of image enhancement and specular highlight removal.
\end{itemize}

\section{Related Work}
The removal of highlights has been a long-standing challenge in the field of image processing and multimedia, with two main approaches emerging: traditional methods and learning-based methods. Traditional methods typically rely on physical models, color, or texture information to capture the relationship between diffuse reflection and specular reflection. These methods utilize optical principles and geometric relationships to derive highlight characteristics and perform detection and removal.

In recent years, learning-based methods have gained significant attention due to their ability to learn complex relationships between input images and desired outputs. These methods often employ techniques such as Generative Adversarial Networks (GANs)~\cite{goodfellow2014generative} and Vision Transformer (ViT)~\cite{dosovitskiy2020vit} to address highlight removal. GANs have been particularly effective in generating realistic images by learning the underlying distribution of the data, while ViTs have proven valuable in capturing long-range dependencies and global context.

\subsection{Traditional Approaches}
The field of traditional methods for highlight removal has primarily focused on separating specular reflections from diffuse reflections, aiming to address the challenges posed by reflections in images and contribute to advancements in image processing and analysis. Early researchers explored various approaches, starting with illumination-based constraints~\cite{quan2003highlight} and methods that relied solely on color recognition of highlight regions~\cite{1374865}. As the field progressed, chromaticity analysis-based methods gained attention, with notable contributions from Shen \etal~\cite{shen2008chromaticity,shen2009simple}.

Researchers have also discovered that leveraging the characteristics of diffuse reflections can be an effective approach for removing specular highlights, as demonstrated by Yang \etal~\cite{yang2010real} and Shen \etal~\cite{shen2013real}. The dichromatic reflection model \cite{shafer1985using} has also been successfully employed in reflection separation methods, as shown by Akashi \etal~\cite{akashi2015separation} and Souza \etal~\cite{souza2018real}.

In addition to these foundational methods, various other approaches have been proposed to tackle the challenge of highlight removal. Nurutdinova \etal~\cite{nurutdinova2017specularity} introduced a semi-automatic algorithm for correction and segmentation of input images, which aids in the highlight removal process. Fu \etal~\cite{fu2019specular} utilized the L0 criterion to enhance the sparsity of encoding coefficients and recover the diffuse reflection component in specular highlight regions, offering a novel perspective on the problem.
Yamamoto \etal~\cite{yamamoto2019general} proposed a method that replaces the separation results of erroneous pixels from a high-pass filter with results from other reference pixels in the image, improving the overall quality of the highlight removal process. Saha \etal~\cite{saha2020combining} took a different approach by combining methods for low-light enhancement and highlight removal to generate high-quality HDR images, demonstrating the potential for integrating multiple techniques to achieve better results.

More recently, Wen \etal~\cite{wen2021polarization} derived a polarization-guided model that incorporates polarization information into an iterative optimization separation strategy. This method effectively separates specular reflections by utilizing polarization information, showcasing the benefits of leveraging additional data sources to improve highlight removal performance.

In summary, traditional methods for highlight removal have evolved from early illumination-based constraints and color recognition approaches to more advanced techniques that leverage chromaticity analysis, diffuse reflection characteristics, and polarization information. These methods have collectively contributed to the development of effective strategies for separating specular reflections from diffuse reflections, paving the way for further advancements in the field of highlight removal and image processing.

\subsection{Deep Learning-based Approaches}
Deep learning methods have gained significant attention in the field of highlight removal due to their potential for higher accuracy and improved generalization, while reducing the need for extensive manual intervention. Various approaches have been proposed to tackle this challenge, each contributing unique insights and techniques.

Guo \etal~\cite{guo2018single} introduced SLRR, a sparse and low-rank reflection model for highlight removal, which laid the groundwork for subsequent deep learning-based methods. Building upon this, Hou \etal~\cite{hou2021text} proposed a hybrid framework that combines a highlight detection network and a highlight removal network, demonstrating promising results in removing specular highlights from text images.

To further improve the modeling of the relationship between diffuse and specular highlight regions, Wu \etal~\cite{wu2021single} developed SpecularityNet, a GAN-based method that incorporates an attention mechanism. Similarly, Fu \etal~\cite{fu2021multi} proposed a multi-task network that integrates the specular highlight image formation model to enhance highlight removal performance.

Liang \etal~\cite{liang2021research} introduced an advanced deep learning method that combines mirror separation and intrinsic decomposition using an adversarial neural network. This approach aims to extract various decomposition results, such as pure diffuse image, normal map, albedo map, visibility map, and residual map, from a single facial image with specular reflections.

Several other deep learning models have been developed to address highlight removal, including Unet-Transformer~\cite{wu2023joint}, which employs a highlight detection module as a mask to guide the removal task, and MG-CycleGAN~\cite{hu2022mask}, which leverages a mask generated by removing specular highlights from unpaired data to guide Cycle-GAN in transforming the problem into an image-to-image translation task.

TSHRNet~\cite{fu2023towards} has demonstrated superior performance in scenarios involving multiple objects and complex lighting conditions, while SHMGAN~\cite{Anwer2023} is a neural network framework capable of effectively separating specular highlight maps and mirror distribution maps without the need for manual input labels.

Despite the advancements made by these state-of-the-art methods, they often encounter issues such as color inconsistency between highlight regions and the background, as well as the generation of unrealistic content within the highlight regions. To address these challenges, Hu \etal~\cite{hu2024highlight} proposed a neural network framework that effectively mitigates these problems, further pushing the boundaries of highlight removal techniques.

In summary, the field of highlight removal has witnessed significant progress through the development of deep learning methods, each contributing novel approaches and techniques to improve accuracy, generalization, and overall performance. However, there remains room for improvement in terms of color consistency and the generation of realistic content within highlight regions, which future research should aim to address.

\begin{figure*}[ht]
\centering
% \vspace*{-0.05in}
\includegraphics[width=\textwidth]{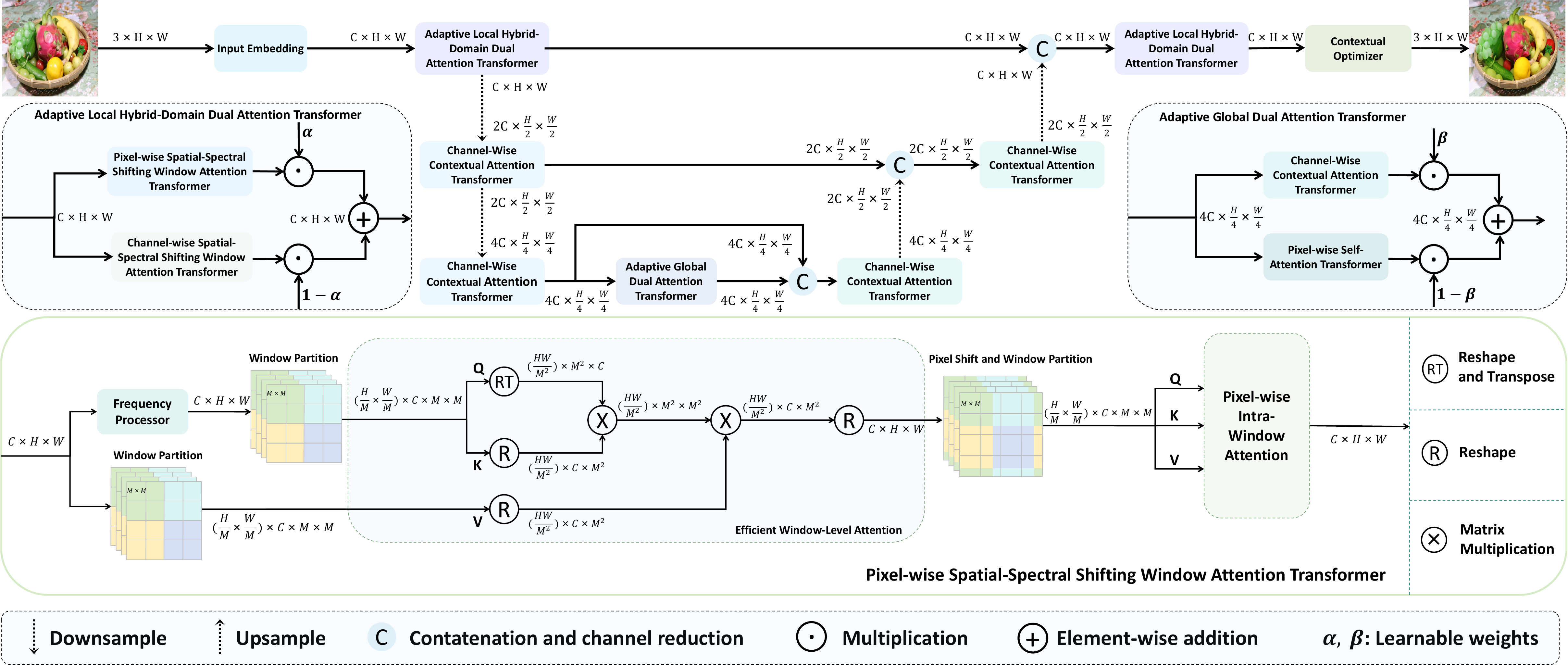}
\caption{{\small The overall architecture of our proposed Dual-Hybrid Attention Network for Specular Highlight Removal (DHAN-SHR).}}
\label{fig:overall}
\end{figure*}
\section{Methodology}
\subsection{Overall Architecture}
Figure \ref{fig:overall} illustrates the architecture of our proposed ``DHAN-SHR'', an end-to-end, one-stage network that takes a single image with specular highlight as input, without requiring any additional information such as highlight masks or priors. The network adopts a U-shape encoder-bottleneck-decoder structure to apply scale-specific feature learning methods at different levels.

The network begins by applying Adaptive Local Hybrid-Domain Dual Attention Transformers (L-HD-DAT) to low-level, high-resolution feature maps, capturing spatial and spectral information with two levels of local attention. The encoder pathway progressively downsamples the image, using Channel-Wise Contextual Attention Transformers (CCAT) to capture contextual information at lower resolutions. At the bottleneck, Adaptive Global Dual Attention Transformers (G-DAT) capture high-level semantic features. 
In the decoder pathway, the encoded features are gradually upsampled, recovering spatial details and fine-grained information. To maintain a balance between the encoder and decoder, we employ the same modules used in the corresponding levels of the encoder. And By concatenating these features via skip connections, the network can effectively combine the strengths of both pathways.

The network's architecture, with its scale-specific feature learning methods, enables effective capture and processing of features at different scales and semantic levels, leading to accurate and efficient specular highlight removal.

\subsection{Adaptive Local Hybrid-Domain Dual Attention Transformer (L-HD-DAT)}
The Adaptive Local Hybrid-Domain Dual Attention Transformer (L-HD-DAT) is a crucial component of our proposed DHAN-SHR network, situated at the topmost position of the U-shaped architecture. Its primary objective is to process feature maps with the highest resolution, matching that of the input image, which contains the most abundant details. In the context of specular highlight removal, accurately detecting and removing specular highlights while restoring the corresponding diffuse visuals with consistent texture and detailed colors is critical for achieving high-quality results. This task is often considered the last mile in determining the visual quality of the highlight removal process.

The L-HD-DAT employs two parallel attention mechanisms to capture both inter-channel and inter-pixel relationships within local windows. These attention mechanisms are implemented through the Pixel-wise Spatial-Spectral Shifting Window Attention Transformer (P\_SSSWAT) and the Channel-wise Spatial-Spectral Shifting Window Attention Transformer (C\_SSSWAT). To adaptively adjust the contribution of each attention mechanism during the training process, we introduce a learnable weight coefficient $\alpha$. The L-HD-DAT can be formulated as follows:
{\small
	\begin{equation}
		\text{L-HD-DAT}(\mathbf{F}) = \alpha \times \text{P\_SSSWAT}(\mathbf{F}) + (1 - \alpha) \times \text{C\_SSSWAT}(\mathbf{F}),
		\label{eq:l-hd-dat}
	\end{equation}
}
where $\mathbf{F}$ represents the input features.

Both P\_SSSWAT and C\_SSSWAT follow the same procedure, which can be described as:
{\small
	\begin{equation}
		\begin{split}
			&\mathbf{Y} = \mathbf{F} + \text{SSSWA}(\text{LN}(\mathbf{F}), \text{LN}(\text{FP}(\mathbf{F}))), \\
			&\text{SSSWAT}(\textbf{F}) = \mathbf{Y} + \text{FFN}(\text{LN}(\mathbf{Y})).
		\end{split}
	\end{equation}
}
In this procedure, $\textbf{F}$ denotes the input features with dimension $C \times H \times W$, $\mathrm{LN}$ represents the LayerNorm operation, and $\mathrm{FP}$ is the Frequency Processor. The Spatial-Spectral Shifting Window Attention ($\mathrm{SSSWA}$) is a key component of both P\_SSSWAT and C\_SSSWAT. The Feed Forward Network ($\mathrm{FFN}$) consists of three convolutional layers that further process the attended features, enabling the network to capture complex spatial relationships and refine the feature representations.

\subsubsection{P\_SSSWA: Pixel-wise Spatial-Spectral Shifting Window Attention}
Both P\_SSSWAT and C\_SSSWAT calculate intra-window attentions twice sequentially: first on the original features and then on the shifted features. 
The Pixel-wise Spatial-Spectral Shifting Window Attention (P\_SSSWA) calculation procedure is illustrated in the lower half of Figure \ref{fig:overall}, with the first cascaded attention depicted in detail. The second attention calculation follows the similar approach, differing only in the input feature maps and the apply of an attention mask.
Given input features $\mathbf{F} \in \mathbb{R}^{C \times H \times W}$, we first obtain the corresponding spectral features $\mathbf{F_s}$ with the same dimension $C \times H \times W$ using the Frequency Processor.
Both $\mathbf{F_s}$ and $\mathbf{F}$ are then partitioned into non-overlapping $M \times M$ windows, similar to SwinIR \cite{liang2021swinir}, where $M$ represents the window height or width (in pixels). To ensure consistent window sizes, we pad the feature maps' right and bottom edges with zeros before partitioning. After partitioning and reshaping, $\mathbf{F_s}$ and $\mathbf{F}$ have dimensions $\frac{HW}{M^2} \times C \times M^2$, where $\frac{HW}{M^2}$ represents the total number of windows in a single channel.

Next, we project the spectral features $\mathbf{F_s}$ to $\mathbf{Q_s}$ (query) and $\mathbf{K_s}$ (key) and compute their self-attention, enabling the model to capture fine details and subtle variations that may be challenging to discern in the spatial domain alone. The resulting spectral attention, with dimensions $\frac{HW}{M^2} \times M^2 \times M^2$, is then multiplied with the spatial features $\mathbf{F}$ (as $\mathbf{V}$ (value)), allowing the model to selectively attend to relevant spatial regions based on insights gained from the frequency domain analysis. In short, the first cascaded Pixel-wise Spatial-Spectral Window Attention is computed as:
{\small
	\begin{equation}
		\text{P\_SSSWA\_1}(\mathbf{Q_s}, \mathbf{K_s}, \mathbf{V}) =\mathbf{V} \cdot \text{Softmax}\left(\frac{\mathbf{Q_s}^T \cdot \mathbf{K_s}}{M}\right).
		\label{eq:psssa1}
	\end{equation}
}
By integrating information from both the spatial and spectral domains, this attention mechanism enables the model to effectively identify and remove specular highlights while preserving the underlying surface details, resulting in improved specular highlight removal performance.

\begin{figure}[ht]
\centering
% \vspace*{-0.05in}
\includegraphics[width=0.46\textwidth]{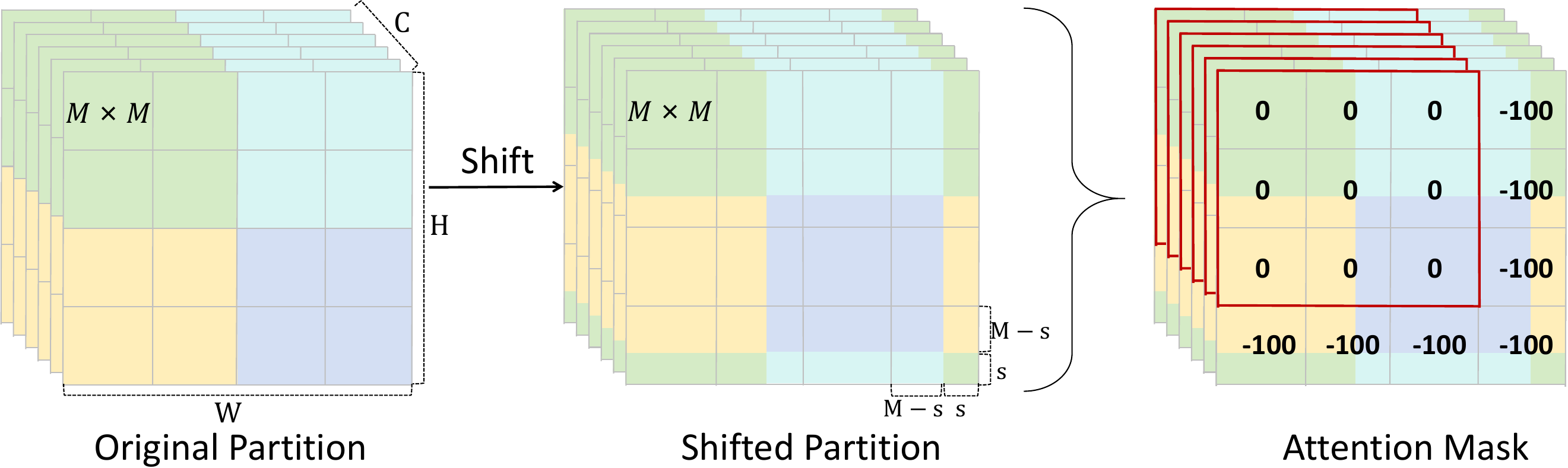}
\caption{{\small Illustration of the window shifting approach and the attention mask applied to the pixel-wise shifting window attention.}}
\label{fig:shift_win}
\end{figure}

Before the partition-reversed output features of $\text{P\_SSSWA\_1}$ are cascadedly input to the second attention block of P\_SSSWAT, we perform a cyclic shift on the pixels of each feature map by $s$ pixels in both horizontal and vertical directions, to cover the areas on the boundaries of the windows from the first partition.
This shifting rule is illustrated using four color blocks in Figure \ref{fig:shift_win}. Unlike Swin Transformer's \cite{liu2021swin} patch-based window partition rules, we directly partition the windows on the pixels themselves. This pixel-level window partitioning allows the window-based attention mechanism to reflect more detailed textures, as it operates on the raw pixel values rather than on abstracted patches.

We then partition and reshape the shifted features $\mathbf{F_{sh}} \in \mathbb{R}^{C \times H \times W}$ into dimensions $\frac{HW}{M^2} \times C \times M^2$, similar to the previous attention block, and project it to obtain $\mathbf{Q_{sh}}$, $\mathbf{K_{sh}}$ and $\mathbf{V_{sh}}$ with the same dimension $\frac{HW}{M^2} \times C \times M^2$. The self-attention of the shifted window is calculated by:
{\small 
	\begin{equation}
		\text{P\_SSSWA\_2}(\mathbf{Q_{sh}}, \mathbf{K_{sh}}, \mathbf{V_{sh}}) =\mathbf{V_{sh}} \cdot \text{Softmax}\left(\frac{\mathbf{Q_{sh}}^T \cdot \mathbf{K_{sh}}}{M} + \mathbf{Mask}\right).
		\label{eq:psssa2}
	\end{equation}
}
In addition to the different input features, the second cascaded attention differs in the inclusion of an attention mask used to distinguish pixels that were originally not adjacent but are now in the same window due to the cyclic shifting. This can be observed on the right and bottom sides of the shifted features in Figure \ref{fig:shift_win}. The mask has the same size as the input shifted feature maps $\mathbf{F_{sh}}$, ensuring a one-to-one position correspondence. We assign a value of $0$ to the positions in the windows where the pixels were originally adjacent, and a value of $-100$ to the positions in the windows where the pixels were originally not adjacent. Figure \ref{fig:shift_win} provides an intuitive understanding of these mask values. By adding this mask to the self-attention of $\mathbf{Q_{sh}}$ and $\mathbf{K_{sh}}$, the model maintains the relationship between originally adjacent pixels while suppressing the influence of pixels that are only adjacent due to cyclic shifting.

Since the attention computation is limited to non-overlapping windows, it significantly reduces the computational complexity compared to traditional self-attention mechanisms that consider the entire image. In traditional approaches, the computational complexity is usually quadratic relative to the image size, and the computational cost is especially tremendous for high-resolution feature maps. However, for one calculation of intra-window pixel-wise attention with input features of dimension $C \times H \times W$ and a window size of $M \times M$, the approximate number of operations, such as element-wise multiplication and addition, is $(HW)M^2(2C-1) + (HW)C(2M^2-1)$. Since both M and C are constants, it means it has a linear computational complexity with respect to the image size. So the intra-window pixel attention not only helps to concentrate on local fine details but also offers computational efficiency.

\subsubsection{C\_SSSWA: Channel-wise Spatial-Spectral Shifting Window Attention}
The working process of the C\_SSSWA is similar to the previously described P\_SSSWA. However, the main difference lies in the inter-channel window range attention calculation approach, which differs from P\_SSSWA's inter-pixel counterpart. The input features remain the same, with $\mathbf{F} \in \mathbb{R}^{C \times H \times W}$ representing the spatial domain features and $\mathbf{F_s} \in \mathbb{R}^{C \times H \times W}$ representing the corresponding spectral domain feature maps. 

Following the same procedure as P\_SSSWA, we perform window partitioning, reshaping, and projection of $\mathbf{F}$ and $\mathbf{F_s}$ to obtain the corresponding window-based features $\mathbf{Q_s}$, $\mathbf{K_s}$, $\mathbf{V}$ in the same dimension $\frac{HW}{M^2} \times C \times M^2$. The first attention block, which hybridizes the spectral and spatial domain features, is calculated as:
{\small 
	\begin{equation}
		\text{C\_SSSWA\_1}(\mathbf{Q_s}, \mathbf{K_s}, \mathbf{V}) =\text{Softmax}\left(\frac{\mathbf{Q_s} \cdot \mathbf{K_s}^T}{\tau}\right) \cdot \mathbf{V},
		\label{eq:csssa1}
	\end{equation}
}
where $\tau$ acts as a learnable temperature parameter that modulates the magnitude of the dot product.
The attention result of $\mathbf{K_s}$ to $\mathbf{Q_s}$ has a dimension of $\frac{HW}{M^2} \times C \times C$, representing the interrelationships between the C channels for each window among the total $\frac{HW}{M^2}$ windows.

In the next step, we reverse the result of equation (\ref{eq:csssa1}) with window partitioned dimension $\frac{HW}{M^2} \times C \times M^2$ back to the input feature maps' dimension $C \times H \times W$. After shifting the feature maps by $s$ pixels, we partition, reshape, and project the new window-based features as $\mathbf{Q_{sh}}$, $\mathbf{K_{sh}}$, and $\mathbf{V_{sh}}$ for the second attention block, which is similar to P\_SSSWA in employing shifted windows to capture dependencies across window boundaries, as shown in equation (\ref{eq:csssa2}).
However, unlike P\_SSSWA, we do not use an attention mask as in equation (\ref{eq:psssa2}). This is because when calculating the inter-channel relationships, we multiply the pixels that are in aligned positions within the window and then add them together. Therefore, the computation is not dependent on the pixels' positional relationships.
{\small 
	\begin{equation}
		\text{C\_SSSWA\_2}(\mathbf{Q_{sh}}, \mathbf{K_{sh}}, \mathbf{V_{sh}}) =\text{Softmax}\left(\frac{\mathbf{Q_{sh}} \cdot \mathbf{K_{sh}}^T}{\tau}\right) \cdot \mathbf{V_{sh}},
		\label{eq:csssa2}
	\end{equation}
}
where $\mathbf{Q_{sh}}, \mathbf{K_{sh}}, \mathbf{V_{sh}} \in \frac{HW}{M^2} \times C \times M^2$, and $\tau$ is still a learnable temperature parameter. The computational complexity of both equation (\ref{eq:csssa1}) and equation (\ref{eq:csssa2}) is approximately $4C^2(HW)$, which is linearly related to the size of the image $H \times W$.

\begin{algorithm}
        \caption{Frequency Processor}
        \label{alg1}
    
        \begin{algorithmic}[1]
            \Require $\mathbf{F}$ (input features)
            \Ensure $\mathbf{F_{s}}$ (frequency processed features)

            \State Apply convolution: $\mathbf{identity_1} \gets Conv2d_{1 \times 1}(\mathbf{F})$
            \State Apply convolution: $\mathbf{identity_2} \gets Conv2d_{1 \times 1}(\mathbf{F})$
            \State Compute FFT of $\mathbf{F}$ and keep the real part: $\mathbf{F_{fft}} \gets \text{FFT}(\mathbf{F}, \text{dim}=(-2, -1)).\text{real}$
            \State Apply convolution to $\mathbf{F_{fft}}$: $\mathbf{F_{fft}} \gets \text{GELU}(Conv2d_{1 \times 1}(\mathbf{F_{fft}}))$
            \State Pass through MLP layers: $\mathbf{F_{fft}} \gets \text{MLPs}(\mathbf{F_{fft}})$
            \State Compute inverse FFT: $\mathbf{F_{ifft}} \gets \text{IFFT}(\mathbf{F_{fft}}, \text{dim}=(-2, -1)).\text{real}$
            \State Add residual connection: $\mathbf{F_s} \gets \mathbf{F_{ifft}} + \mathbf{identity_2}$
            
            \State Apply toning: $\mathbf{F_s} \gets \text{Toning}(\text{Concat}([\mathbf{F_s}, \mathbf{identity_1}], \text{dim}=1))$
 
        \end{algorithmic}
        
\end{algorithm}
\subsubsection{Frequency Processor}

The Frequency Processor employed in both P\_SSSWAT and C\_SSSWAT generates feature maps that have undergone transforms to the frequency domain, as shown in Algorithm \ref{alg1}. 
First, we compute the discrete Fourier transform of the input feature maps. Then, a shallow convolution and GeLU activation are applied to the frequency features to introduce non-linearity and enhance the expressiveness of the frequency-domain representations.
Before executing the inverse Fourier Transform, MLP layers are applied to the frequency-domain features to adapt and refine the spectral representations, capturing more accurate and contextually relevant frequency-domain information essential for reconstructing the diffuse visuals.

To provide conditional guidance during the spectral features' training process, we obtain two identities of the input spatial features, which are then added and concatenated, respectively, to the inverse Fourier Transform of the spectral features.
Finally, a toning operation is applied to the spectral features, which removes the phase information and focuses on the magnitude-based spectral features, helping to suppress noise and outliers while emphasizing relevant frequency ranges for accurate and coherent diffuse visual reconstruction.

\begin{figure*}[ht]
    \begin{minipage}[b]{1.0\linewidth}
        \begin{minipage}[b]{0.12\linewidth}
            \centering
            \centerline{\includegraphics[width=\linewidth]{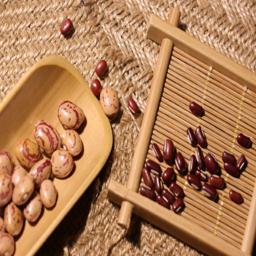}}
        \end{minipage}   
        \begin{minipage}[b]{0.12\linewidth}
            \centering
            \centerline{\includegraphics[width=\linewidth]{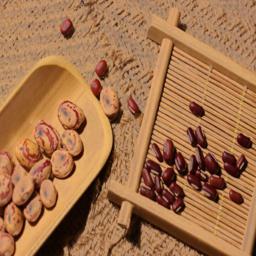}}
        \end{minipage}
        \begin{minipage}[b]{0.12\linewidth}
            \centering
            \centerline{\includegraphics[width=\linewidth]{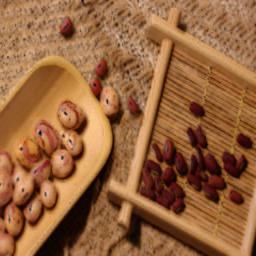}}
        \end{minipage}   
        \begin{minipage}[b]{0.12\linewidth}
            \centering
            \centerline{\includegraphics[width=\linewidth]{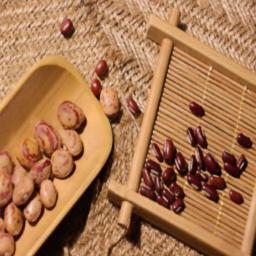}}
        \end{minipage}
        \begin{minipage}[b]{0.12\linewidth}
            \centering
            \centerline{\includegraphics[width=\linewidth]{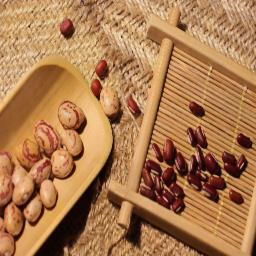}}
        \end{minipage}
        \begin{minipage}[b]{0.12\linewidth}
            \centering
            \centerline{\includegraphics[width=\linewidth]{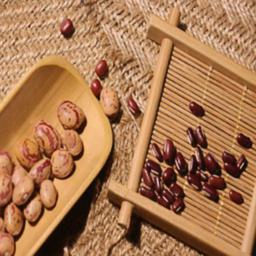}}
        \end{minipage}
        \begin{minipage}[b]{0.12\linewidth}
            \centering
            \centerline{\includegraphics[width=\linewidth]{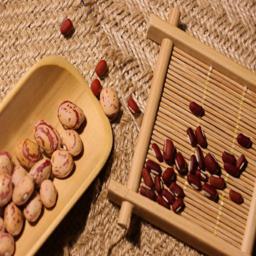}}
        \end{minipage}
        \begin{minipage}[b]{0.12\linewidth}
            \centering
            \centerline{\includegraphics[width=\linewidth]{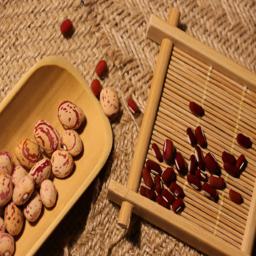}}
        \end{minipage}
    \end{minipage}

    \begin{minipage}[b]{1.0\linewidth}
        \begin{minipage}[b]{0.12\linewidth}
            \centering
            \centerline{\includegraphics[width=\linewidth]{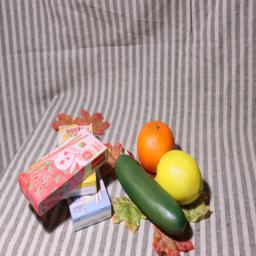}}
        \end{minipage}   
        \begin{minipage}[b]{0.12\linewidth}
            \centering
            \centerline{\includegraphics[width=\linewidth]{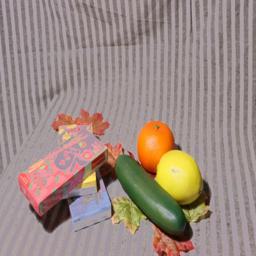}}
        \end{minipage}
        \begin{minipage}[b]{0.12\linewidth}
            \centering
            \centerline{\includegraphics[width=\linewidth]{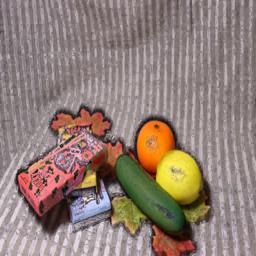}}
        \end{minipage}   
        \begin{minipage}[b]{0.12\linewidth}
            \centering
            \centerline{\includegraphics[width=\linewidth]{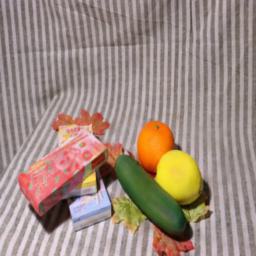}}
        \end{minipage}
        \begin{minipage}[b]{0.12\linewidth}
            \centering
            \centerline{\includegraphics[width=\linewidth]{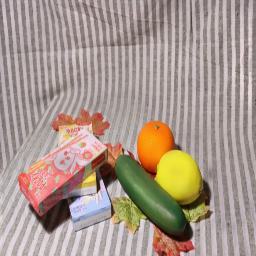}}
        \end{minipage}
        \begin{minipage}[b]{0.12\linewidth}
            \centering
            \centerline{\includegraphics[width=\linewidth]{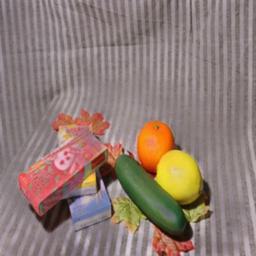}}
        \end{minipage}
        \begin{minipage}[b]{0.12\linewidth}
            \centering
            \centerline{\includegraphics[width=\linewidth]{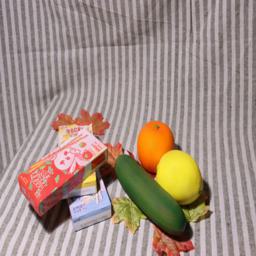}}
        \end{minipage}
        \begin{minipage}[b]{0.12\linewidth}
            \centering
            \centerline{\includegraphics[width=\linewidth]{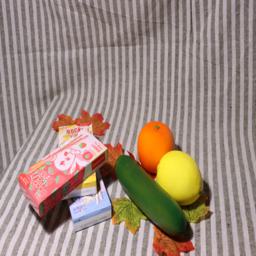}}
        \end{minipage}
    \end{minipage}

    \begin{minipage}[b]{1.0\linewidth}
        \begin{minipage}[b]{0.12\linewidth}
            \centering
            \centerline{\includegraphics[width=\linewidth]{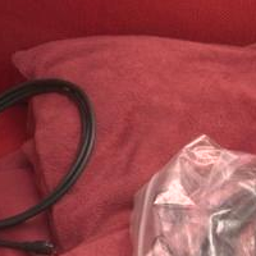}}
        \end{minipage}   
        \begin{minipage}[b]{0.12\linewidth}
            \centering
            \centerline{\includegraphics[width=\linewidth]{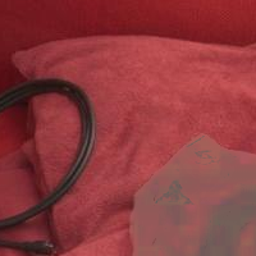}}
        \end{minipage}
        \begin{minipage}[b]{0.12\linewidth}
            \centering
            \centerline{\includegraphics[width=\linewidth]{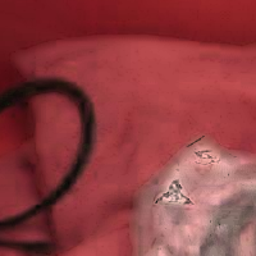}}
        \end{minipage}   
        \begin{minipage}[b]{0.12\linewidth}
            \centering
            \centerline{\includegraphics[width=\linewidth]{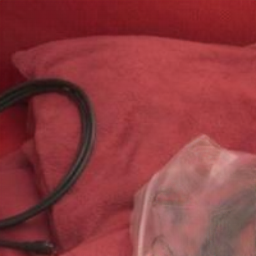}}
        \end{minipage}
        \begin{minipage}[b]{0.12\linewidth}
            \centering
            \centerline{\includegraphics[width=\linewidth]{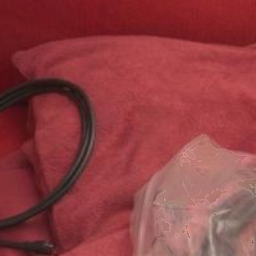}}
        \end{minipage}
        \begin{minipage}[b]{0.12\linewidth}
            \centering
            \centerline{\includegraphics[width=\linewidth]{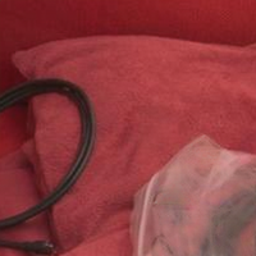}}
        \end{minipage}
        \begin{minipage}[b]{0.12\linewidth}
            \centering
            \centerline{\includegraphics[width=\linewidth]{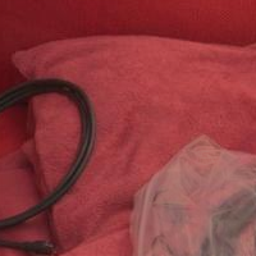}}
        \end{minipage}
        \begin{minipage}[b]{0.12\linewidth}
            \centering
            \centerline{\includegraphics[width=\linewidth]{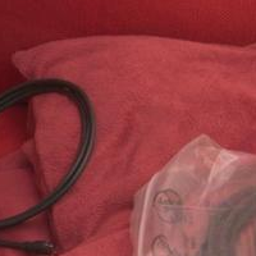}}
        \end{minipage}
    \end{minipage}

    \begin{minipage}[b]{1.0\linewidth}
        \begin{minipage}[b]{0.12\linewidth}
            \centering
            \centerline{\includegraphics[width=\linewidth]{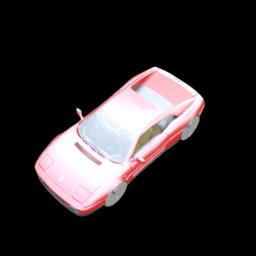}}
        \end{minipage}   
        \begin{minipage}[b]{0.12\linewidth}
            \centering
            \centerline{\includegraphics[width=\linewidth]{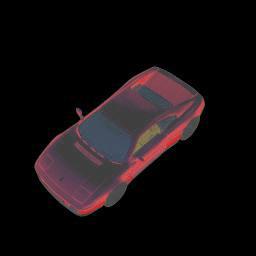}}
        \end{minipage}
        \begin{minipage}[b]{0.12\linewidth}
            \centering
            \centerline{\includegraphics[width=\linewidth]{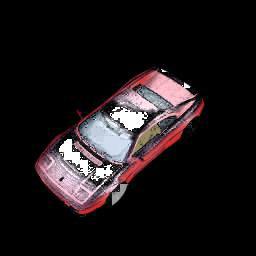}}
        \end{minipage}   
        \begin{minipage}[b]{0.12\linewidth}
            \centering
            \centerline{\includegraphics[width=\linewidth]{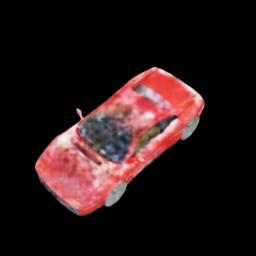}}
        \end{minipage}
        \begin{minipage}[b]{0.12\linewidth}
            \centering
            \centerline{\includegraphics[width=\linewidth]{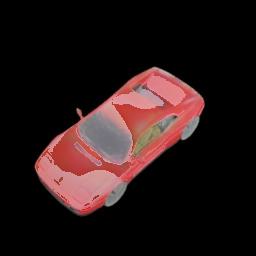}}
        \end{minipage}
        \begin{minipage}[b]{0.12\linewidth}
            \centering
            \centerline{\includegraphics[width=\linewidth]{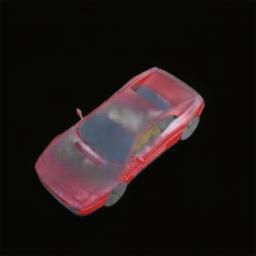}}
        \end{minipage}
        \begin{minipage}[b]{0.12\linewidth}
            \centering
            \centerline{\includegraphics[width=\linewidth]{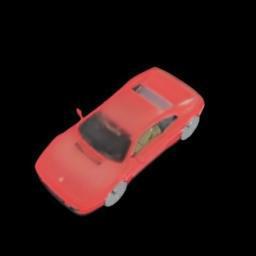}}
        \end{minipage}
        \begin{minipage}[b]{0.12\linewidth}
            \centering
            \centerline{\includegraphics[width=\linewidth]{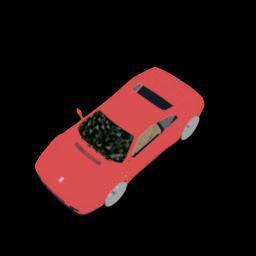}}
        \end{minipage}
    \end{minipage}

    \begin{minipage}[b]{1.0\linewidth}
        \begin{minipage}[b]{0.12\linewidth}
            \centering
            \centerline{\includegraphics[width=\linewidth]{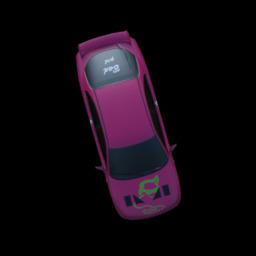}}
            \centerline{{\small (a)Input}}\medskip
        \end{minipage}   
        \begin{minipage}[b]{0.12\linewidth}
            \centering
            \centerline{\includegraphics[width=\linewidth]{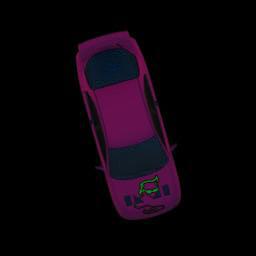}}
            \centerline{{\small (b)Shen \cite{shen2009simple}}}\medskip
        \end{minipage}
        \begin{minipage}[b]{0.12\linewidth}
            \centering
            \centerline{\includegraphics[width=\linewidth]{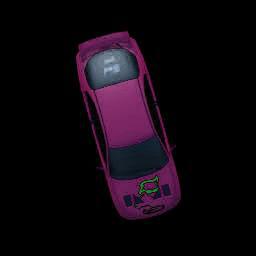}}
            \centerline{{\small (c)Yamamoto \cite{yamamoto2019general}}}\medskip
        \end{minipage}   
        \begin{minipage}[b]{0.12\linewidth}
            \centering
            \centerline{\includegraphics[width=\linewidth]{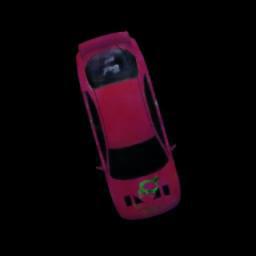}}
            \centerline{{\small (d)TSHRNet \cite{fu2023towards}}}\medskip
        \end{minipage}
        \begin{minipage}[b]{0.12\linewidth}
            \centering
            \centerline{\includegraphics[width=\linewidth]{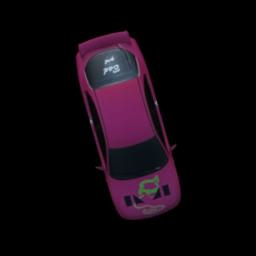}}
            \centerline{{\small (e)Wu \cite{wu2023joint}}}\medskip
        \end{minipage}
        \begin{minipage}[b]{0.12\linewidth}
            \centering
            \centerline{\includegraphics[width=\linewidth]{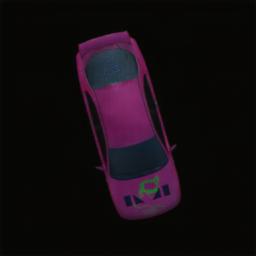}}
            \centerline{{\small (f)JSHDR \cite{fu2021multi}}}\medskip
        \end{minipage}
        \begin{minipage}[b]{0.12\linewidth}
            \centering
            \centerline{\includegraphics[width=\linewidth]{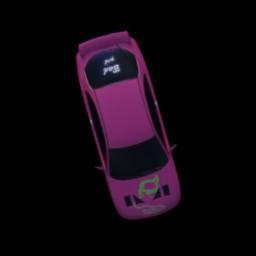}}
            \centerline{{\small (g)Ours}}\medskip
        \end{minipage}
        \begin{minipage}[b]{0.12\linewidth}
            \centering
            \centerline{\includegraphics[width=\linewidth]{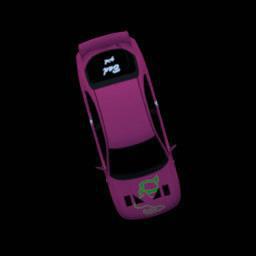}}
            \centerline{{\small (h)GT}}\medskip
        \end{minipage}
    \end{minipage}
    
    \caption{{\small Visual comparative analysis of our method against leading SOTA approaches, highlighting our superior ability to remove specular highlights while preserving the original image's color tone, structure, and crucial details, such as text clarity on reflective surfaces.}}
    \label{fig:compare}
\end{figure*}

\subsection{Channel-Wise Contextual Attention Transformer (CCAT)}
The Channel-Wise Contextual Attention Transformers (CCAT) scale up its concentrated grain compared to L-HD-DAT by their strategic deployment within both the encoder pathway, which progressively downsamples, and the decoder pathway, which conversely upsamples. This positioning allows CCAT to operate on feature maps that possess a higher semantic level than the original input image. To effectively address the challenge of contextual information extraction across channels, CCAT incorporates an attention mechanism that is meticulously designed to weigh the significance of each channel based on its contextual relationship with others. This is achieved through a multi-headed self-attention module as equation (\ref{eq:cca}) that operates on the channel dimension, enabling the network to dynamically adjust the emphasis on different channels according to their contextual relevance.
{\small 
	\begin{equation}
		\text{CCA}(\mathbf{\hat{Q}}, \mathbf{\hat{K}}, \mathbf{\hat{V}}) =\text{Softmax}\left(\frac{\mathbf{\hat{Q}} \cdot \mathbf{\hat{K}}^T}{\sqrt{\hat{C} / h}}\right) \cdot \mathbf{\hat{V}}.
		\label{eq:cca}
	\end{equation}
}
Here, $\mathbf{\hat{Q}}$, $\mathbf{\hat{K}}$ and $\mathbf{\hat{V}}$ denote the projections of the feature maps $\mathbf{\hat{F}} \in \mathbb{R}^{\hat{C} \times \hat{H} \times \hat{W}}$, which are adjusted to $1/2$ or $1/4$ the resolution of the original input image to correspond to different scale levels within the U-shaped architecture. These projections are then reshaped to a dimension of $h \times (\hat{C}/h) \times (\hat{H}\hat{W})$, where $h$ denotes the number of attention heads. The paralleled attention heads make CCAT capable of capturing different aspects of inter-channel relationships. This diversity allows for a more comprehensive understanding of the feature maps, enhancing the network's ability to discern and emphasize the most relevant features for specular highlight removal.

\subsection{Adaptive Global Dual Attention Transformer (G-DAT)}

Adaptive Global Dual Attention Transformers (G-DAT) are deployed at the bottleneck between the encoder and decoder. Given the higher semantic level at this stage, the information each pixel carries is more abstract and globally contextual than at any other layer in the network.  In response to this heightened level of abstraction, we have devised parallel dual attention mechanisms. These are designed to concurrently grasp the intricate inter-channel and inter-pixel dependencies on a global scale, as depicted in the following equation:
{\small
	\begin{equation}
		\text{G-DAT}(\mathbf{\hat{F}}) = \beta \times \text{CCAT}(\mathbf{\hat{F}}) + (1 - \beta) \times \text{PSAT}(\mathbf{\hat{F}}).
		\label{eq:g-dat}
	\end{equation}
}
Here, the considerable reduction in the size of each feature map $\mathbf{\hat{F}}$ due to multiple downsampling stages implies that even the Pixel-wise Self-Attention Transformer (PSAT), which directs attention on a global scale, does not impose a significant computational load. The operational methodology of PSAT is akin to that detailed in Equation (\ref{eq:psssa1}); however, unlike the attention in Equation (\ref{eq:psssa1}) which is confined within a specific window, PSAT expands its focus to cover a global range. Additionally, since the input to the bottleneck module lacks certain visual details, we do not incorporate spectral domain elements in PSAT. This approach ensures that the computational demand is directly proportional to the spatial dimensions of each feature map, maintaining linear computational efficiency.

\subsection{Objective Function}
To optimize the performance of our specular highlight removal methodology, our goal is to align the output as closely as possible with the ground truth diffuse images. This entails not only achieving pixel-level accuracy but also preserving key image attributes such as luminance, contrast, and structure. To accomplish this, we introduce a composite objective function that integrates both Mean Squared Error Loss ($\mathcal{L}_{f}$), which accounts for pixel-level fidelity, and Structural Similarity Loss ($\mathcal{L}_{s}$), focusing on maintaining structural integrity. The formulation of our overall objective function is as follows:
{\small 
	\begin{equation}
		\mathcal{L} = \alpha \cdot \mathcal{L}_{f} + \beta \cdot \mathcal{L}_{s},
		\label{eq:loss}
	\end{equation}
}
where $\alpha$ and $\beta$ have been empirically set to 1 and 0.4, respectively.

We incorporate the Structural Similarity Index Measure (SSIM), as proposed by Wang et al. \cite{wang2004image}, into the structural similarity component $\mathcal{L}_{s}$ of our objective function, as delineated below:
{\small 
	\begin{equation}
		\mathcal{L}_{s} = 1 - \frac{(2 \mu_D \mu_G + C_1)(2 \sigma_{DG} + C_2)}{(\mu_D^2 + \mu_G^2 + C_1)(\sigma_D^2 + \sigma_G^2 + C_2)},
		\label{eq:4}
	\end{equation}
}
where $\mu_D$ and $\mu_G$ represent the average pixel values of the output diffuse image D and the target ground truth image G, $\sigma_D^2$ and $\sigma_G^2$ are the variances of images D and G, ad $\sigma_{DG}$ is the covariance between D and G. $C_1$ and $C_2$ are constants to stabilize the division with weak denominators.

\section{Experiments}
\subsection{Benchmark}
In the pursuit of advancing the real-world applicability of Specular Highlight Removal (SHR) networks, the utilization of real-world datasets is crucial. To date, the PSD \cite{wu2021single} dataset stands out as the only comprehensive collection across various objects where both highlight samples and their corresponding diffuse ground truths are captured in real-world scenarios. However, PSD's limitation lies in its repetition of scenes across different polarization angles, reducing sample diversity despite having nearly 10000 pairs. To overcome the limitations of dataset size and enhance the robustness of deep learning methods for SHR, we propose the creation of a hybrid benchmark. This benchmark amalgamates real-world samples with synthetically generated samples that adhere to optical principles, offering a balanced and comprehensive training and testing environment.

Our hybrid benchmark encompasses data from three distinct datasets: PSD \cite{wu2021single}, SHIQ \cite{fu2021multi}, and SSHR \cite{fu2023towards}, each serving a different purpose. The PSD dataset provides real-world photographs of both specular and diffuse samples, making it a valuable resource for realistic training. Contrastingly, SHIQ's real-world specular samples are paired with diffuse images created via the RPCA method \cite{guo2014robust}, once state-of-the-art. Despite possibly no longer being the leading technique, its use enriches training diversity, highlighting its ongoing relevance. Meanwhile, the SSHR dataset offers a fully synthetic collection, created with open-source rendering software, adding a valuable dimension of diversity.
For training, we selected 9481 pairs from PSD, 9825 pairs from SHIQ, and a random subset of 10000 pairs from SSHR. For testing, our selection comprised all 947 pairs from PSD, 1000 pairs from SHIQ, and 1000 randomly chosen pairs from SSHR. This selection strategy guarantees a diverse and balanced collection of samples for both the training and testing phases, ensuring robustness.

To our knowledge, this is the first instance where a hybrid benchmark combining multiple datasets has been employed for SHR. This approach not only enriches the training and testing environment but also sets a new standard for future research in the field, potentially enhancing the performance and generalizability of SHR methods across a wider range of real-world and synthetic scenarios.

\begin{table}[ht]
\caption{{\small The quantitative comparison results, arranging traditional methods in the upper section and learning-based approaches below. The highest-performing results are emphasized in bold, while the second-best are underscored.}}
\adjustbox{width=\columnwidth}{%
\begin{threeparttable}
\begin{tabular}{lccccccccc}
\toprule
\multirow{2}{*}{{\LARGE Method}} & \multicolumn{3}{c}{{\LARGE PSD (947images)}}       & \multicolumn{3}{c}{{\LARGE SHIQ (1000images)}}     & \multicolumn{3}{c}{{\LARGE SSHR (1000images)}}     \\ \cmidrule(lr){2-4} \cmidrule(lr){5-7} \cmidrule(lr){8-10}
                                  & {\Large PSNR$\uparrow$} & {\Large SSIM$\uparrow$} & {\Large LPIPS$\downarrow$} & {\Large PSNR$\uparrow$} & {\Large SSIM$\uparrow$} & {\Large LPIPS$\downarrow$} & {\Large PSNR$\uparrow$} & {\Large SSIM$\uparrow$} & {\Large LPIPS$\downarrow$} \\ \midrule
{\Large Tan} \cite{1374865}                  & {\LARGE 5.44}      & {\LARGE 0.218}     & {\LARGE 0.746}      & {\LARGE 5.47}       & {\LARGE 0.483}      & {\LARGE 0.823}      & {\LARGE 10.87}      & {\LARGE 0.778}      & {\LARGE 0.357}      \\
{\Large Yoon} \cite{yoon2006fast}            & {\LARGE 16.09}     & {\LARGE 0.498}     & {\LARGE 0.325}      & {\LARGE 19.34}      & {\LARGE 0.679}      & {\LARGE 0.471}      & {\LARGE 28.47}      & {\LARGE 0.916}      & {\LARGE 0.094}      \\
{\Large Shen} \cite{shen2008chromaticity}    & {\LARGE 19.56}     & {\LARGE 0.666}     & {\LARGE 0.238}      & {\LARGE 24.77}      & {\LARGE 0.890}      & {\LARGE 0.200}      & {\LARGE 24.53}      & {\LARGE 0.896}      & {\LARGE 0.101}      \\
{\Large Shen} \cite{shen2009simple}          & {\LARGE 21.33}     & {\LARGE 0.753}     & {\LARGE 0.142 }     & {\LARGE 27.30 }     & {\LARGE 0.917 }     & {\LARGE 0.102 }     & {\LARGE 24.00 }     & {\LARGE 0.891 }     & {\LARGE 0.094}      \\
{\Large Yang} \cite{yang2010real}            & {\LARGE 4.74}      & {\LARGE 0.250}     & {\LARGE 0.893}      & {\LARGE 5.31 }      & {\LARGE 0.556}      & {\LARGE 0.837}      & {\LARGE 10.72}      & {\LARGE 0.781}      & {\LARGE 0.358}      \\
{\Large Shen} \cite{shen2013real}           & {\LARGE 11.51}     & {\LARGE 0.324}     & {\LARGE 0.360}      & {\LARGE 12.24}      & {\LARGE 0.491}      & {\LARGE 0.473}      & {\LARGE 27.13}      & {\LARGE 0.914}      & {\LARGE 0.077}      \\
{\Large Akashi} \cite{akashi2015separation}  & {\LARGE 17.48}     & {\LARGE 0.565}     & {\LARGE 0.334}      & {\LARGE 21.78}      & {\LARGE 0.700}      & {\LARGE 0.460}      & {\LARGE 29.46}      & {\LARGE 0.924}      & {\LARGE 0.076}      \\
{\Large Huo} \cite{huo2015hdr}               & {\LARGE 20.16}     & {\LARGE 0.767}     & {\LARGE 0.182}      & {\LARGE 23.80}      & {\LARGE 0.909}      & {\LARGE 0.154}      & {\LARGE 18.62}      & {\LARGE 0.804}      & {\LARGE 0.281}      \\
{\Large Fu} \cite{fu2019specular}            & {\LARGE 15.24}     & {\LARGE 0.688}     & {\LARGE 0.146}      & {\LARGE 16.40}      & {\LARGE 0.724}      & {\LARGE 0.306}      & {\LARGE 26.15}      & {\LARGE 0.910}      & {\LARGE 0.076}      \\
{\Large Yamamoto} \cite{yamamoto2019general} & {\LARGE 18.37}     & {\LARGE 0.541}     & {\LARGE 0.274}      & {\LARGE 25.49}      & {\LARGE 0.858}      & {\LARGE 0.201}      & {\LARGE 26.95}      & {\LARGE 0.902}      & {\LARGE 0.094}      \\
{\Large Saha} \cite{saha2020combining}       & {\LARGE 15.98}     & {\LARGE 0.455}     & {\LARGE 0.314}      & {\LARGE 22.05}      & {\LARGE 0.832}      & {\LARGE 0.287}      & {\LARGE 23.38}      & {\LARGE 0.886}      & {\LARGE 0.110}      \\  \midrule
{\Large SLRR} \cite{guo2018single}                  & {\LARGE 13.25}     & {\LARGE 0.571}     & {\LARGE 0.235}      & {\LARGE 14.74}      & {\LARGE 0.724}      & {\LARGE 0.283}      & {\LARGE 26.16}      & {\LARGE 0.916}      & {\LARGE 0.060}      \\
{\Large JSHDR\tnote{*}} \cite{fu2021multi}                   & {\LARGE 22.78}     & {\LARGE 0.811}     & {\LARGE 0.089}      & {\LARGE \textbf{37.97}}      & {\LARGE \textbf{0.980}}      & {\LARGE \textbf{0.034}}      & {\LARGE 26.43}      & {\LARGE 0.301}      & {\LARGE 0.059}      \\
{\Large SpecularityNet} \cite{wu2021single}         & {\LARGE 23.58}     & {\LARGE 0.838}     & {\LARGE 0.085}      & {\LARGE 30.92}      & {\LARGE 0.963}      & {\LARGE 0.058}      & {\LARGE 31.07}      & {\LARGE 0.941}      & {\LARGE 0.041}      \\
{\Large MG-CycleGAN} \cite{hu2022mask}              & {\LARGE 22.12}     & {\LARGE 0.815}     & {\LARGE 0.085}      & {\LARGE 26.80}      & {\LARGE 0.935}      & {\LARGE 0.091}      & {\LARGE 28.40}      & {\LARGE 0.874}      & {\LARGE 0.092}      \\
{\Large Wu} \cite{wu2023joint}        & {\LARGE \underline{23.93}}     & {\LARGE \underline{0.863}}     & {\LARGE \underline{0.062}}      & {\LARGE 31.57}      & {\LARGE 0.965}      & {\LARGE 0.059}      & {\LARGE \underline{33.45}}      & {\LARGE \underline{0.951}}      & {\LARGE \underline{0.028}}      \\
{\Large TSHRNet} \cite{fu2023towards}               & {\LARGE 23.30}     & {\LARGE 0.826}     & {\LARGE 0.097}      & {\LARGE \underline{34.57}}      & {\LARGE 0.972}      & {\LARGE 0.044}      & {\LARGE 33.32}      & {\LARGE 0.950}      & {\LARGE 0.036}      \\
{\Large AHA} \cite{hu2024highlight}                 & {\LARGE 20.79}     & {\LARGE 0.845}     & {\LARGE 0.084}      & {\LARGE 21.42}      & {\LARGE 0.903}      & {\LARGE 0.165}      & {\LARGE 31.57}      & {\LARGE 0.944}      & {\LARGE 0.035}      \\
{\Large Ours}      & {\LARGE \textbf{25.28}}     & {\LARGE \textbf{0.883}}     & {\LARGE \textbf{0.049}}      & {\LARGE 33.81}      & {\LARGE \underline{0.975}}      & {\LARGE \underline{0.039}}      & {\LARGE \textbf{36.48}}      & {\LARGE \textbf{0.964}}      & {\LARGE \textbf{0.023}}     \\
\bottomrule
\end{tabular}
    \begin{tablenotes}
      \item[*] {\Large JSHDR's source code is not publicly available; the results are obtained from an executable file provided by its authors.}
    \end{tablenotes}
\end{threeparttable}
}
\label{tab:quan}
\end{table}
% \usepackage{booktabs}
% \usepackage{multirow}
\begin{table}[ht]
\caption{{\small Metric comparison: official SSHR test split vs. randomly selected 1,000-group subset.}}
% \vspace{-1em}
\adjustbox{width=\columnwidth}{%
\begin{tabular}{lccccccc}
\toprule
\multirow{2}{*}{Method} & \multicolumn{3}{c}{SSHR(Subset)} & \multicolumn{3}{c}{SSHR(Full)} & \multirow{2}{*}{\begin{tabular}[c]{@{}c@{}}PSNR \\ Deviation\end{tabular}} \\ \cmidrule(lr){2-4} \cmidrule(lr){5-7}
                         & PSNR$\uparrow$      & SSIM$\uparrow$      & LPIPS$\downarrow$     & PSNR$\uparrow$      & SSIM$\uparrow$      & LPIPS$\downarrow$    &      \\ \midrule
Tan et al.               & 10.87          & 0.778          & 0.357          & 10.78          & 0.775          & 0.359          & 0.82\%                                                                     \\
Yoon et al.              & 28.47          & 0.916          & 0.094          & 28.32          & 0.914          & 0.094          & 0.52\%                                                                     \\
Shen et al.              & 24.53          & 0.896          & 0.101          & 24.59          & 0.895          & 0.100          & 0.26\%                                                                     \\
Shen et al.              & 24.00          & 0.891          & 0.094          & 24.28          & 0.892          & 0.092          & 1.16\%                                                                     \\
Yang et al.              & 10.72          & 0.781          & 0.358          & 10.64          & 0.778          & 0.360          & 0.72\%                                                                     \\
Shen et al.              & 27.13          & 0.914          & 0.077          & 27.20          & 0.913          & 0.077          & 0.25\%                                                                     \\
Akashi et al.            & 29.46          & 0.924          & 0.076          & 29.48          & 0.923          & 0.076          & 0.04\%                                                                     \\
Huo et al.               & 18.62          & 0.804          & 0.281          & 18.59          & 0.802          & 0.280          & 0.17\%                                                                     \\
Fu et al.                & 26.15          & 0.910          & 0.076          & 26.25          & 0.909          & 0.076          & 0.41\%                                                                     \\
Yamamoto et al.          & 26.95          & 0.902          & 0.094          & 27.04          & 0.902          & 0.093          & 0.31\%                                                                     \\
Saha et al.              & 23.38          & 0.886          & 0.110          & 23.48          & 0.886          & 0.108          & 0.45\%                                                                     \\
SLRR                     & 26.16          & 0.916          & 0.060          & 26.34          & 0.916          & 0.059          & 0.67\%                                                                     \\
JSHDR                    & 26.43          & 0.301          & 0.059          & 26.60          & 0.304          & 0.058          & 0.66\%                                                                     \\
SpecularityNet           & 31.07          & 0.941          & 0.041          & 30.92          & 0.940          & 0.042          & 0.47\%                                                                     \\
MG-CycleGAN              & 28.40          & 0.874          & 0.092          & 28.24          & 0.872          & 0.092          & 0.58\%                                                                     \\
Unet-Transformer         & {\ul 33.45}    & {\ul 0.951}    & {\ul 0.028}    & {\ul 33.27}    & {\ul 0.949}    & {\ul 0.029}    & 0.55\%                                                                     \\
TSHRNet                  & 33.32          & 0.950          & 0.036          & 33.14          & 0.948    & 0.036          & 0.55\%                                                                     \\
AHA                      & 31.57          & 0.944          & 0.035          & 31.61          & 0.943          & 0.036          & 0.12\%                                                                     \\ \cmidrule(r){1-8}
Ours & \textbf{36.48} & \textbf{0.964} & \textbf{0.023} & \textbf{36.29} & \textbf{0.962} & \textbf{0.024} & 0.52\%                                                                     \\ \bottomrule
\end{tabular}
}
\label{tab:sshr}
\end{table}
% \usepackage{booktabs}
% \usepackage[normalem]{ulem}
\begin{table}[]
\caption{{\small Computational cost for deep learning methods. The batch size for evaluating the training time per iteration is uniformly set to 2.}}
\adjustbox{width=\columnwidth}{%
\begin{tabular}{@{}lrrrr@{}}
\toprule
\multicolumn{1}{c}{Method}                           & \multicolumn{1}{c}{MACs(G)$\downarrow$} & \multicolumn{1}{c}{Params(M)$\downarrow$} & \multicolumn{1}{c}{Infer Time(ms)$\downarrow$} & \multicolumn{1}{c}{Train Time(ms/iter)$\downarrow$} \\ \midrule
SpecularityNet  & 212.87                       & 17.00                          & 18.33                               & 99.02                                    \\
MG-CycleGAN       & 178.07                       & {\ul 12.78}                    & 115.13                              & 213.24                                   \\
Unet-Transformer & 101.63                       & 53.39                          & 15.45                               & 72.59                                    \\
TSHRNet        & {\ul 72.76}                  & 116.99                         & {\ul 12.00}                         & \textbf{35.90}                           \\
AHA          & 92.40                        & 35.86                          & \textbf{11.26}                      & {\ul 56.07}                              \\
Ours                                                 & \textbf{65.69}               & \textbf{4.53}                  & 43.82                               & 119.84                                   \\ \bottomrule
\end{tabular}
}
\label{tab:cost}
\end{table}

\subsection{Evaluation Metrics}

In our study, we utilize a suite of full-reference evaluation metrics to assess performance, including Peak Signal-to-Noise Ratio (PSNR), Structural Similarity Index (SSIM)~\cite{wang2004image}, and Learned Perceptual Image Patch Similarity (LPIPS)~\cite{zhang2018unreasonable}. For PSNR and SSIM, higher scores denote better performance, indicating a greater similarity between the generated images and the ground truth. On the other hand, a lower score in LPIPS suggests enhanced visual quality, as this metric measures the perceptual similarity between generated and reference images in a way that is more aligned with human visual perception.

\subsection{Implementation Details}
Our model is implemented using PyTorch and trained using the Adam optimizer with default parameters on NVIDIA H800 GPU. To optimize the model, we employ the standard settings of the Adam optimization algorithm, with a batch size of 8 and a learning rate of $1e-4$. To enhance the robustness and generalizability of our model, we incorporate a comprehensive set of data augmentation techniques. These augmentations include random cropping of images, resizing, horizontal and vertical flipping, and the application of the mixup strategy to generate composite images from the original data, thereby exposing the model to a diverse range of variations and improving its ability to handle different data effectively.

\subsection{Comparisons with State-of-the-Art Methods}

To conduct a thorough evaluation of our Specular Highlight Removal (SHR) method relative to the current state-of-the-art, we compared our model against a total of 18 representative SHR techniques, which comprise both 11 traditional and 7 learning-based approaches. For the traditional methods, we processed the test samples directly to obtain their output results. To guarantee a fair comparison, we retrained all the learning-based models on the same benchmark dataset compiled for our study. During this retraining process, we adhered to the training settings (loss, iterations, hyper-parameters, etc.) as specified in the original publications of each method.

\subsubsection{Quantitative Comparison}

Table \ref{tab:quan} showcases the quantitative performance of various specular highlight removal methods across three datasets, utilizing three distinct evaluation metrics. Our model, DHAN-SHR, demonstrates superior performance overall, with the sole exception being the results of JSHDR \cite{fu2021multi} on the SHIQ dataset, which was released in conjunction with JSHDR. It's important to note, however, that JSHDR's source code is not publicly available; our analysis is based on results obtained from an executable file provided by its authors. This limitation means we couldn't retrain JSHDR under the same conditions as other methods, diminishing the comparative value of its performance on the SHIQ dataset. Notably, JSHDR shows significantly lower performance on both the PSD and SSHR datasets compared to DHAN-SHR, with the gap being particularly pronounced outside the SHIQ dataset.

The performance of our DHAN-SHR model on the PSD and SSHR datasets outpaces other methods by a considerable margin, especially highlighting its exceptional performance on the PSD dataset. This dataset, known for its real-world, high-resolution images, underlines the adaptability and effectiveness of our model in real-world application scenarios, suggesting DHAN-SHR's robust capability in addressing specular highlight removal across diverse conditions.

\subsubsection{Qualitative Comparison}

The visual comparisons between our DHAN-SHR and SOTA methods, which include the top 2 traditional and top 3 deep learning methods based on average metric data in Table \ref{tab:quan}, are illustrated in Figures \ref{fig:compare}. For optimal clarity, it is recommended to zoom in.

Observations from Figure \ref{fig:compare} reveal that our method excels not only in removing specular highlights effectively—surpassing even the reference ground truth in the third row—but also in preserving the original tone and consistent color of the entire image. Remarkably, it maintains the detail in diffuse areas and restores clarity to details previously obscured by reflections. In contrast, the methods we compared often fail to fully eliminate highlights, sometimes resulting in black spots within the treated areas. More problematic are the visual effects noted in the fourth row, where these methods disrupt the image's original structure and details, leading to poor visual outcomes. Furthermore, in the fifth row, while competing methods tend to erase or blur text on the car's rear window, our approach successfully retains and sharpens these details.

\subsection{Ablation Studies}

\begin{table}[ht]
\caption{{\small Quantitative results of ablation studies. Key components – Pixel-wise Spatial-Spectral Shifting Window Attention Transformer (P\_SSSWA), Channel-wise Spatial-Spectral Shifting Window Attention Transformer
(C\_SSSWA), Channel-Wise Contextual Attention Transformer (CCAT), Pixel-wise Self-Attention Transformer(PSAT) and FrequencyProcessor (FP) – are individually ablated.}}
\centering
\adjustbox{width=0.37\textwidth}{%
\begin{tabular}{lrrrr}
\toprule
Method       & PSNR$\uparrow$ & SSIM$\uparrow$ & LPIPS$\downarrow$ \\
\midrule
W/O P\_SSSWAT & 30.68          & 0.937          & 0.042          \\
W/O C\_SSSWAT & 30.93          & 0.938          & 0.041          \\
W/O CCAT      & 30.49           & 0.934          & 0.044          \\
W/O PSAT      & 30.80          & 0.935          & 0.043          \\
W/O FP        & 30.92          & 0.938          & 0.041          \\
Full Model    & \textbf{31.86} & \textbf{0.940} & \textbf{0.037} \\

\bottomrule
\end{tabular}
}
\label{tab:ablation}
\end{table}
% \usepackage{booktabs}
\begin{table}[]
\caption{Ablation results that averaged on the PSD, SHIQ, and SSHR test sets.}
\adjustbox{width=\columnwidth}{%
%\begin{threeparttable}
\begin{tabular}{@{}ccccccccc@{}}
\toprule
U-Net & P\_SSSWAT & C\_SSSWAT & FP & CCAT & PSAT & PSNR$\uparrow$          & SSIM$\uparrow$          & LPIPS$\downarrow$         \\ \midrule
\faCheck     & \usym{2715}       & \usym{2715}       & \usym{2715}  & \usym{2715}    & \usym{2715}    & 29.79          & 0.930          & 0.048          \\
\faCheck     & \usym{2715}       & \usym{2715}       & \usym{2715}  & \faCheck    & \usym{2715}    & 30.95          & 0.933          & 0.043          \\
\faCheck     & \usym{2715}       & \usym{2715}       & \usym{2715}  & \faCheck    & \faCheck    & 31.48          & 0.935          & 0.041          \\
\faCheck     & \faCheck       & \usym{2715}       & \usym{2715}  & \usym{2715}    & \usym{2715}    & 30.30          & 0.932          & 0.044          \\
\faCheck     & \usym{2715}       & \faCheck       & \usym{2715}  & \usym{2715}    & \usym{2715}    & 30.36          & 0.931          & 0.045          \\
\faCheck     & \faCheck       & \faCheck       & \usym{2715}  & \usym{2715}    & \usym{2715}    & 30.54          & 0.931          & 0.044          \\
\faCheck     & \faCheck       & \faCheck       & \faCheck  & \usym{2715}    & \usym{2715}    & 30.77          & 0.933          & 0.044          \\
\faCheck     & \faCheck       & \faCheck       & \faCheck  & \faCheck    & \faCheck    & \textbf{31.86} & \textbf{0.940} & \textbf{0.037} \\ \bottomrule
\end{tabular}
%    \begin{tablenotes}
%      \item[*] P\_SSSWAT = P\_SWAT + FP, C\_SSSWAT = C\_SWAT + FP.
%    \end{tablenotes}
%\end{threeparttable}
}
\label{tab:ablation}
\end{table}
To assess the efficacy of our model's integral features, ablation studies were executed, utilizing the averaged metrics from the PSD, SHIQ, and SSHR test sets to ensure a robust evaluation. These studies examined the model's performance with and without its key elements, such as attention mechanisms and processor that handle spatial-spectral data. The summarized findings, detailed in Table \ref{tab:ablation}, underscore our full model's superiority. By selectively removing specific components, we could discern their individual contributions, affirming that the integration of these elements is vital for optimal specular highlight removal performance.

\section{Conclusion}

In this study, we introduce the Dual-Hybrid Attention Network for Specular Highlight Removal (DHAN-SHR), a novel approach that effectively addresses the challenge of specular highlight removal in multimedia applications. DHAN-SHR employs adaptive attention mechanisms that capture both global-level dependencies and local-level relationships, eliminating the need for additional priors or supervision.
We assembled an extensive benchmark dataset combining images from three different highlight removal datasets. Experimental results demonstrate that DHAN-SHR outperforms 18 state-of-the-art methods across various test datasets and metrics.

\bibliographystyle{IEEEtran}
\bibliography{IEEEabrv,ref}

\newpage
\clearpage
\appendices

\section{Full Comparisons}
\begin{figure*}[hb]
    \begin{minipage}[b]{1.0\linewidth}
        \begin{minipage}[b]{0.137\linewidth}
            \centering
            \centerline{
            \stackinset{l}{0pt}{t}{0pt}{%
                \colorbox{gray}{%
                    \textcolor{white}{\textbf{\footnotesize (a)}}%
                }%
            }{\includegraphics[width=\linewidth]{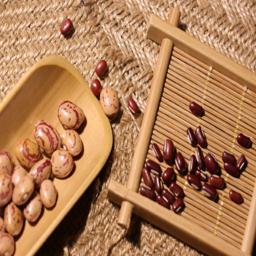}}}
        \end{minipage}   
        \hfill
        \begin{minipage}[b]{0.137\linewidth}
            \centering
            \centerline{
            \stackinset{l}{0pt}{t}{0pt}{%
                \colorbox{gray}{%
                    \textcolor{white}{\textbf{\footnotesize (b)}}%
                }%
            }{\includegraphics[width=\linewidth]{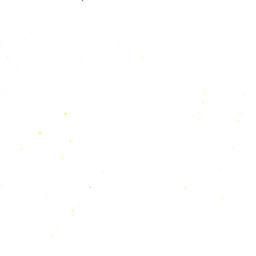}}}
        \end{minipage}   
        \hfill
        \begin{minipage}[b]{0.137\linewidth}
            \centering
            \centerline{
            \stackinset{l}{0pt}{t}{0pt}{%
                \colorbox{gray}{%
                    \textcolor{white}{\textbf{\footnotesize (c)}}%
                }%
            }{\includegraphics[width=\linewidth]{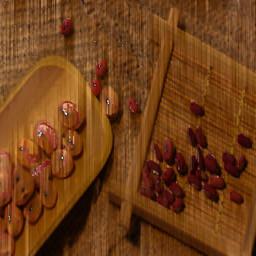}}}
        \end{minipage}   
        \hfill
        \begin{minipage}[b]{0.137\linewidth}
            \centering
            \centerline{
            \stackinset{l}{0pt}{t}{0pt}{%
                \colorbox{gray}{%
                    \textcolor{white}{\textbf{\footnotesize (d)}}%
                }%
            }{\includegraphics[width=\linewidth]{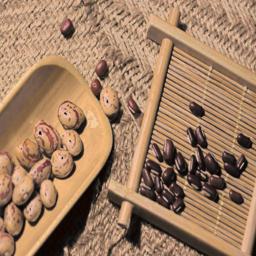}}}
        \end{minipage}   
        \hfill
        \begin{minipage}[b]{0.137\linewidth}
            \centering
            \centerline{
            \stackinset{l}{0pt}{t}{0pt}{%
                \colorbox{gray}{%
                    \textcolor{white}{\textbf{\footnotesize (e)}}%
                }%
            }{\includegraphics[width=\linewidth]{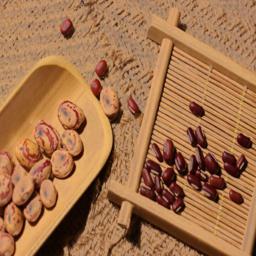}}}
        \end{minipage}
        \hfill
        \begin{minipage}[b]{0.137\linewidth}
            \centering
            \centerline{
            \stackinset{l}{0pt}{t}{0pt}{%
                \colorbox{gray}{%
                    \textcolor{white}{\textbf{\footnotesize (f)}}%
                }%
            }{\includegraphics[width=\linewidth]{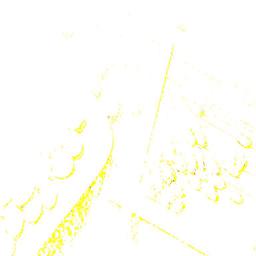}}}
        \end{minipage}
        \hfill
        \begin{minipage}[b]{0.137\linewidth}
            \centering
            \centerline{
            \stackinset{l}{0pt}{t}{0pt}{%
                \colorbox{gray}{%
                    \textcolor{white}{\textbf{\footnotesize (g)}}%
                }%
            }{\includegraphics[width=\linewidth]{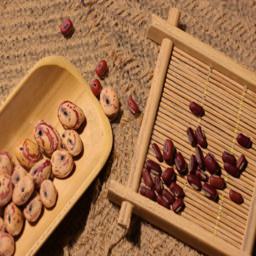}}}
        \end{minipage}   
    \end{minipage}

    \begin{minipage}[b]{1.0\linewidth}
        \begin{minipage}[b]{0.137\linewidth}
            \centering
            \centerline{
            \stackinset{l}{0pt}{t}{0pt}{%
                \colorbox{gray}{%
                    \textcolor{white}{\textbf{\footnotesize (h)}}%
                }%
            }{\includegraphics[width=\linewidth]{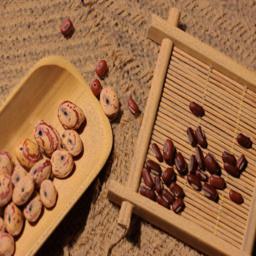}}}
        \end{minipage}   
        \hfill
        \begin{minipage}[b]{0.137\linewidth}
            \centering
            \centerline{
            \stackinset{l}{0pt}{t}{0pt}{%
                \colorbox{gray}{%
                    \textcolor{white}{\textbf{\footnotesize (i)}}%
                }%
            }{\includegraphics[width=\linewidth]{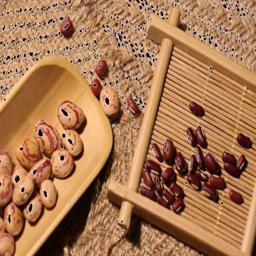}}}
        \end{minipage}   
        \hfill
        \begin{minipage}[b]{0.137\linewidth}
            \centering
            \centerline{
            \stackinset{l}{0pt}{t}{0pt}{%
                \colorbox{gray}{%
                    \textcolor{white}{\textbf{\footnotesize (j)}}%
                }%
            }{\includegraphics[width=\linewidth]{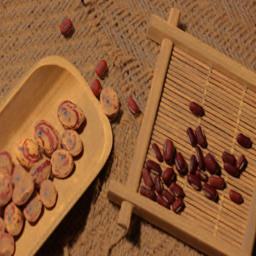}}}
        \end{minipage}   
        \hfill
        \begin{minipage}[b]{0.137\linewidth}
            \centering
            \centerline{
            \stackinset{l}{0pt}{t}{0pt}{%
                \colorbox{gray}{%
                    \textcolor{white}{\textbf{\footnotesize (k)}}%
                }%
            }{\includegraphics[width=\linewidth]{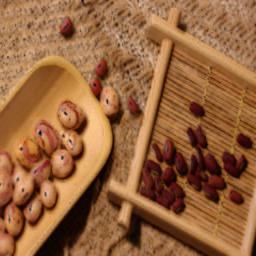}}}
        \end{minipage}
        \hfill
        \begin{minipage}[b]{0.137\linewidth}
            \centering
            \centerline{
            \stackinset{l}{0pt}{t}{0pt}{%
                \colorbox{gray}{%
                    \textcolor{white}{\textbf{\footnotesize (l)}}%
                }%
            }{\includegraphics[width=\linewidth]{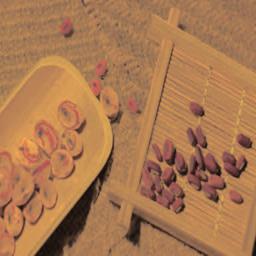}}}
        \end{minipage}
        \hfill
        \begin{minipage}[b]{0.137\linewidth}
            \centering
            \centerline{
            \stackinset{l}{0pt}{t}{0pt}{%
                \colorbox{gray}{%
                    \textcolor{white}{\textbf{\footnotesize (m)}}%
                }%
            }{\includegraphics[width=\linewidth]{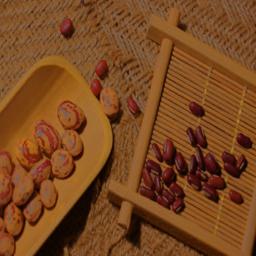}}}
        \end{minipage}   
        \hfill
        \begin{minipage}[b]{0.137\linewidth}
            \centering
            \centerline{
            \stackinset{l}{0pt}{t}{0pt}{%
                \colorbox{gray}{%
                    \textcolor{white}{\textbf{\footnotesize (n)}}%
                }%
            }{\includegraphics[width=\linewidth]{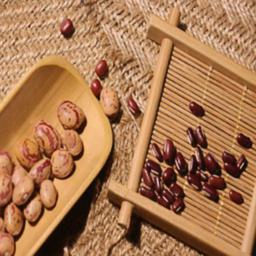}}}
        \end{minipage} 
    \end{minipage}

    \begin{minipage}[b]{1.0\linewidth}
        \begin{minipage}[b]{0.137\linewidth}
            \centering
            \centerline{
            \stackinset{l}{0pt}{t}{0pt}{%
                \colorbox{gray}{%
                    \textcolor{white}{\textbf{\footnotesize (o)}}%
                }%
            }{\includegraphics[width=\linewidth]{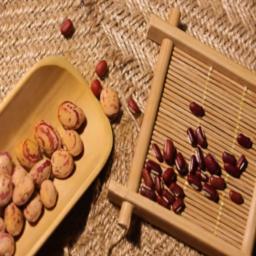}}}
        \end{minipage}   
        \hfill
        \begin{minipage}[b]{0.137\linewidth}
            \centering
            \centerline{
            \stackinset{l}{0pt}{t}{0pt}{%
                \colorbox{gray}{%
                    \textcolor{white}{\textbf{\footnotesize (p)}}%
                }%
            }{\includegraphics[width=\linewidth]{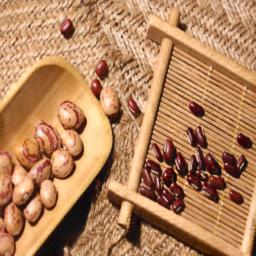}}}
        \end{minipage}   
        \hfill
        \begin{minipage}[b]{0.137\linewidth}
            \centering
            \centerline{
            \stackinset{l}{0pt}{t}{0pt}{%
                \colorbox{gray}{%
                    \textcolor{white}{\textbf{\footnotesize (q)}}%
                }%
            }{\includegraphics[width=\linewidth]{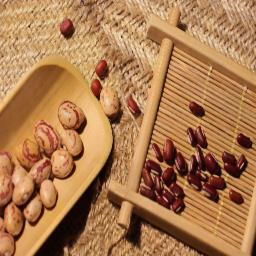}}}
        \end{minipage}
        \hfill
        \begin{minipage}[b]{0.137\linewidth}
            \centering
            \centerline{
            \stackinset{l}{0pt}{t}{0pt}{%
                \colorbox{gray}{%
                    \textcolor{white}{\textbf{\footnotesize (r)}}%
                }%
            }{\includegraphics[width=\linewidth]{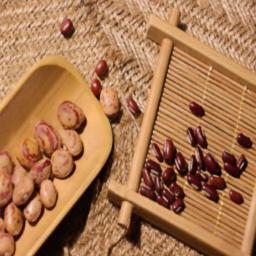}}}
        \end{minipage}
        \hfill
        \begin{minipage}[b]{0.137\linewidth}
            \centering
            \centerline{
            \stackinset{l}{0pt}{t}{0pt}{%
                \colorbox{gray}{%
                    \textcolor{white}{\textbf{\footnotesize (s)}}%
                }%
            }{\includegraphics[width=\linewidth]{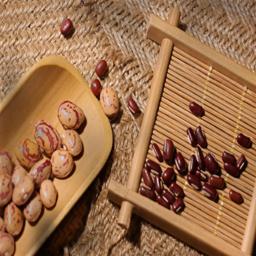}}}
        \end{minipage}
        \hfill
        \begin{minipage}[b]{0.137\linewidth}
            \centering
            \centerline{
            \stackinset{l}{0pt}{t}{0pt}{%
                \colorbox{gray}{%
                    \textcolor{white}{\textbf{\footnotesize (t)}}%
                }%
            }{\includegraphics[width=\linewidth]{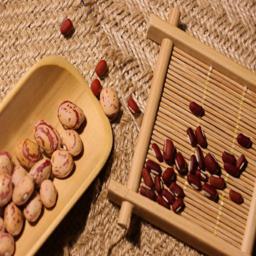}}}
        \end{minipage}
        \hfill
        \begin{minipage}[b]{0.137\linewidth}
            \centering
            \centerline{
            \stackinset{l}{0pt}{t}{0pt}{%
                \colorbox{gray}{%
                    \textcolor{white}{\textbf{\footnotesize (u)}}%
                }%
            }{\includegraphics[width=\linewidth]{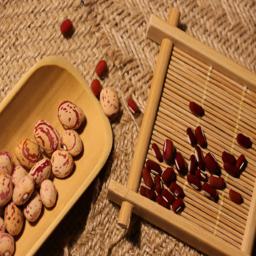}}}
        \end{minipage}
    \end{minipage}

    \begin{minipage}[b]{1.0\linewidth}
        \begin{minipage}[b]{0.137\linewidth}
            \centering
            \centerline{
            \stackinset{l}{0pt}{t}{0pt}{%
                \colorbox{gray}{%
                    \textcolor{white}{\textbf{\footnotesize (a)}}%
                }%
            }{\includegraphics[width=\linewidth]{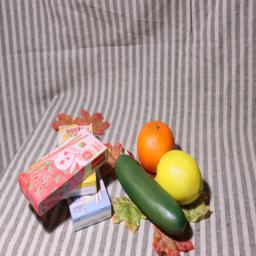}}}
        \end{minipage}   
        \hfill
        \begin{minipage}[b]{0.137\linewidth}
            \centering
            \centerline{
            \stackinset{l}{0pt}{t}{0pt}{%
                \colorbox{gray}{%
                    \textcolor{white}{\textbf{\footnotesize (b)}}%
                }%
            }{\includegraphics[width=\linewidth]{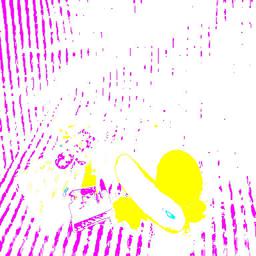}}}
        \end{minipage}   
        \hfill
        \begin{minipage}[b]{0.137\linewidth}
            \centering
            \centerline{
            \stackinset{l}{0pt}{t}{0pt}{%
                \colorbox{gray}{%
                    \textcolor{white}{\textbf{\footnotesize (c)}}%
                }%
            }{\includegraphics[width=\linewidth]{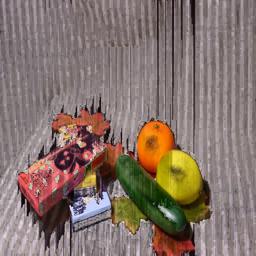}}}
        \end{minipage}   
        \hfill
        \begin{minipage}[b]{0.137\linewidth}
            \centering
            \centerline{
            \stackinset{l}{0pt}{t}{0pt}{%
                \colorbox{gray}{%
                    \textcolor{white}{\textbf{\footnotesize (d)}}%
                }%
            }{\includegraphics[width=\linewidth]{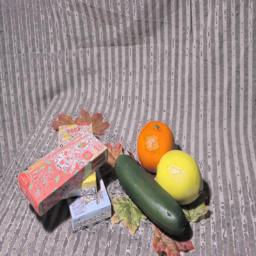}}}
        \end{minipage}   
        \hfill
        \begin{minipage}[b]{0.137\linewidth}
            \centering
            \centerline{
            \stackinset{l}{0pt}{t}{0pt}{%
                \colorbox{gray}{%
                    \textcolor{white}{\textbf{\footnotesize (e)}}%
                }%
            }{\includegraphics[width=\linewidth]{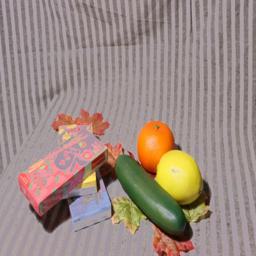}}}
        \end{minipage}
        \hfill
        \begin{minipage}[b]{0.137\linewidth}
            \centering
            \centerline{
            \stackinset{l}{0pt}{t}{0pt}{%
                \colorbox{gray}{%
                    \textcolor{white}{\textbf{\footnotesize (f)}}%
                }%
            }{\includegraphics[width=\linewidth]{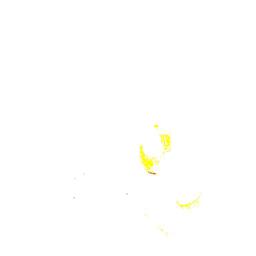}}}
        \end{minipage}
        \hfill
        \begin{minipage}[b]{0.137\linewidth}
            \centering
            \centerline{
            \stackinset{l}{0pt}{t}{0pt}{%
                \colorbox{gray}{%
                    \textcolor{white}{\textbf{\footnotesize (g)}}%
                }%
            }{\includegraphics[width=\linewidth]{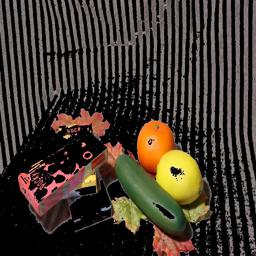}}}
        \end{minipage}   
    \end{minipage}

    \begin{minipage}[b]{1.0\linewidth}
        \begin{minipage}[b]{0.137\linewidth}
            \centering
            \centerline{
            \stackinset{l}{0pt}{t}{0pt}{%
                \colorbox{gray}{%
                    \textcolor{white}{\textbf{\footnotesize (h)}}%
                }%
            }{\includegraphics[width=\linewidth]{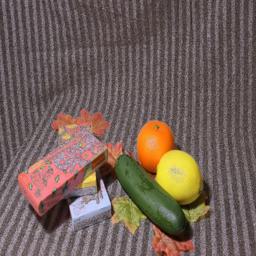}}}
        \end{minipage}   
        \hfill
        \begin{minipage}[b]{0.137\linewidth}
            \centering
            \centerline{
            \stackinset{l}{0pt}{t}{0pt}{%
                \colorbox{gray}{%
                    \textcolor{white}{\textbf{\footnotesize (i)}}%
                }%
            }{\includegraphics[width=\linewidth]{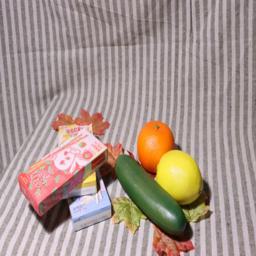}}}
        \end{minipage}   
        \hfill
        \begin{minipage}[b]{0.137\linewidth}
            \centering
            \centerline{
            \stackinset{l}{0pt}{t}{0pt}{%
                \colorbox{gray}{%
                    \textcolor{white}{\textbf{\footnotesize (j)}}%
                }%
            }{\includegraphics[width=\linewidth]{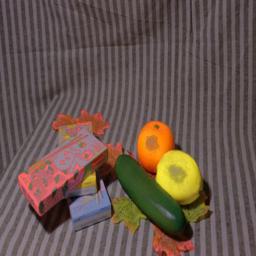}}}
        \end{minipage}   
        \hfill
        \begin{minipage}[b]{0.137\linewidth}
            \centering
            \centerline{
            \stackinset{l}{0pt}{t}{0pt}{%
                \colorbox{gray}{%
                    \textcolor{white}{\textbf{\footnotesize (k)}}%
                }%
            }{\includegraphics[width=\linewidth]{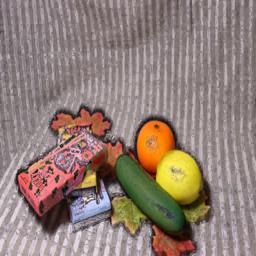}}}
        \end{minipage}
        \hfill
        \begin{minipage}[b]{0.137\linewidth}
            \centering
            \centerline{
            \stackinset{l}{0pt}{t}{0pt}{%
                \colorbox{gray}{%
                    \textcolor{white}{\textbf{\footnotesize (l)}}%
                }%
            }{\includegraphics[width=\linewidth]{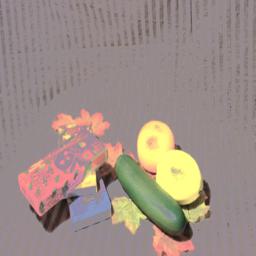}}}
        \end{minipage}
        \hfill
        \begin{minipage}[b]{0.137\linewidth}
            \centering
            \centerline{
            \stackinset{l}{0pt}{t}{0pt}{%
                \colorbox{gray}{%
                    \textcolor{white}{\textbf{\footnotesize (m)}}%
                }%
            }{\includegraphics[width=\linewidth]{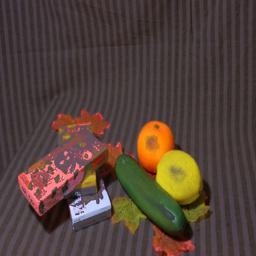}}}
        \end{minipage}   
        \hfill
        \begin{minipage}[b]{0.137\linewidth}
            \centering
            \centerline{
            \stackinset{l}{0pt}{t}{0pt}{%
                \colorbox{gray}{%
                    \textcolor{white}{\textbf{\footnotesize (n)}}%
                }%
            }{\includegraphics[width=\linewidth]{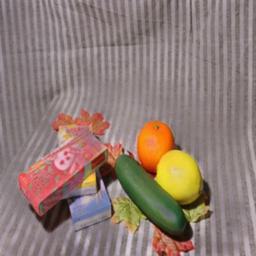}}}
        \end{minipage} 
    \end{minipage}

    \begin{minipage}[b]{1.0\linewidth}
        \begin{minipage}[b]{0.137\linewidth}
            \centering
            \centerline{
            \stackinset{l}{0pt}{t}{0pt}{%
                \colorbox{gray}{%
                    \textcolor{white}{\textbf{\footnotesize (o)}}%
                }%
            }{\includegraphics[width=\linewidth]{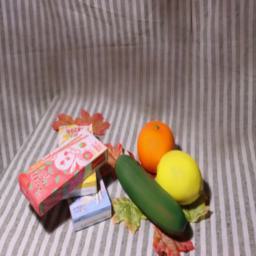}}}
        \end{minipage}   
        \hfill
        \begin{minipage}[b]{0.137\linewidth}
            \centering
            \centerline{
            \stackinset{l}{0pt}{t}{0pt}{%
                \colorbox{gray}{%
                    \textcolor{white}{\textbf{\footnotesize (p)}}%
                }%
            }{\includegraphics[width=\linewidth]{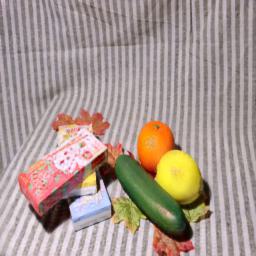}}}
        \end{minipage}   
        \hfill
        \begin{minipage}[b]{0.137\linewidth}
            \centering
            \centerline{
            \stackinset{l}{0pt}{t}{0pt}{%
                \colorbox{gray}{%
                    \textcolor{white}{\textbf{\footnotesize (q)}}%
                }%
            }{\includegraphics[width=\linewidth]{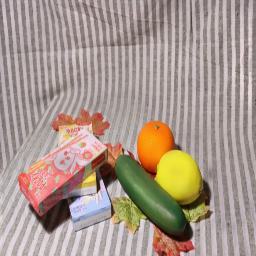}}}
        \end{minipage}
        \hfill
        \begin{minipage}[b]{0.137\linewidth}
            \centering
            \centerline{
            \stackinset{l}{0pt}{t}{0pt}{%
                \colorbox{gray}{%
                    \textcolor{white}{\textbf{\footnotesize (r)}}%
                }%
            }{\includegraphics[width=\linewidth]{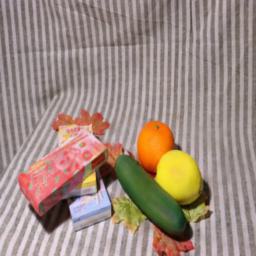}}}
        \end{minipage}
        \hfill
        \begin{minipage}[b]{0.137\linewidth}
            \centering
            \centerline{
            \stackinset{l}{0pt}{t}{0pt}{%
                \colorbox{gray}{%
                    \textcolor{white}{\textbf{\footnotesize (s)}}%
                }%
            }{\includegraphics[width=\linewidth]{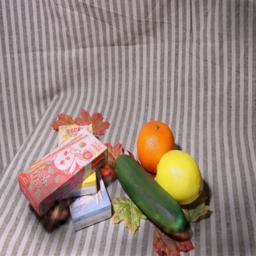}}}
        \end{minipage}
        \hfill
        \begin{minipage}[b]{0.137\linewidth}
            \centering
            \centerline{
            \stackinset{l}{0pt}{t}{0pt}{%
                \colorbox{gray}{%
                    \textcolor{white}{\textbf{\footnotesize (t)}}%
                }%
            }{\includegraphics[width=\linewidth]{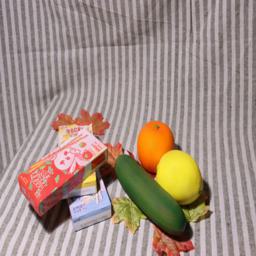}}}
        \end{minipage}
        \hfill
        \begin{minipage}[b]{0.137\linewidth}
            \centering
            \centerline{
            \stackinset{l}{0pt}{t}{0pt}{%
                \colorbox{gray}{%
                    \textcolor{white}{\textbf{\footnotesize (u)}}%
                }%
            }{\includegraphics[width=\linewidth]{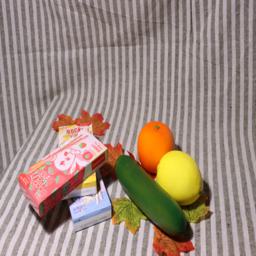}}}
        \end{minipage}
    \end{minipage}

    \begin{minipage}[b]{1.0\linewidth}
        \begin{minipage}[b]{0.137\linewidth}
            \centering
            \centerline{
            \stackinset{l}{0pt}{t}{0pt}{%
                \colorbox{gray}{%
                    \textcolor{white}{\textbf{\footnotesize (a)}}%
                }%
            }{\includegraphics[width=\linewidth]{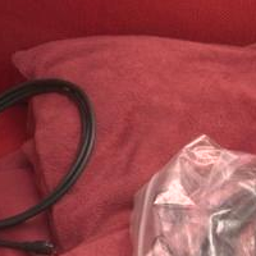}}}
        \end{minipage}   
        \hfill
        \begin{minipage}[b]{0.137\linewidth}
            \centering
            \centerline{
            \stackinset{l}{0pt}{t}{0pt}{%
                \colorbox{gray}{%
                    \textcolor{white}{\textbf{\footnotesize (b)}}%
                }%
            }{\includegraphics[width=\linewidth]{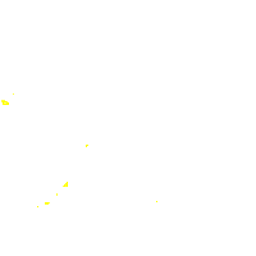}}}
        \end{minipage}   
        \hfill
        \begin{minipage}[b]{0.137\linewidth}
            \centering
            \centerline{
            \stackinset{l}{0pt}{t}{0pt}{%
                \colorbox{gray}{%
                    \textcolor{white}{\textbf{\footnotesize (c)}}%
                }%
            }{\includegraphics[width=\linewidth]{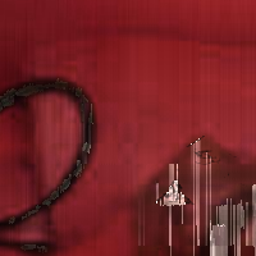}}}
        \end{minipage}   
        \hfill
        \begin{minipage}[b]{0.137\linewidth}
            \centering
            \centerline{
            \stackinset{l}{0pt}{t}{0pt}{%
                \colorbox{gray}{%
                    \textcolor{white}{\textbf{\footnotesize (d)}}%
                }%
            }{\includegraphics[width=\linewidth]{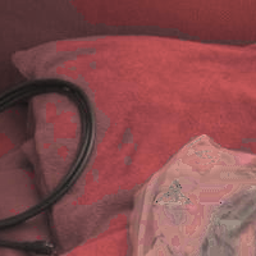}}}
        \end{minipage}   
        \hfill
        \begin{minipage}[b]{0.137\linewidth}
            \centering
            \centerline{
            \stackinset{l}{0pt}{t}{0pt}{%
                \colorbox{gray}{%
                    \textcolor{white}{\textbf{\footnotesize (e)}}%
                }%
            }{\includegraphics[width=\linewidth]{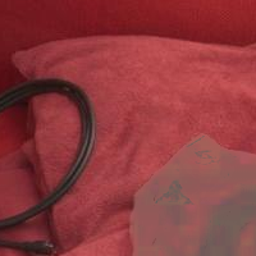}}}
        \end{minipage}
        \hfill
        \begin{minipage}[b]{0.137\linewidth}
            \centering
            \centerline{
            \stackinset{l}{0pt}{t}{0pt}{%
                \colorbox{gray}{%
                    \textcolor{white}{\textbf{\footnotesize (f)}}%
                }%
            }{\includegraphics[width=\linewidth]{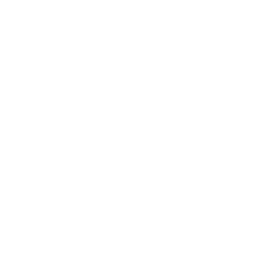}}}
        \end{minipage}
        \hfill
        \begin{minipage}[b]{0.137\linewidth}
            \centering
            \centerline{
            \stackinset{l}{0pt}{t}{0pt}{%
                \colorbox{gray}{%
                    \textcolor{white}{\textbf{\footnotesize (g)}}%
                }%
            }{\includegraphics[width=\linewidth]{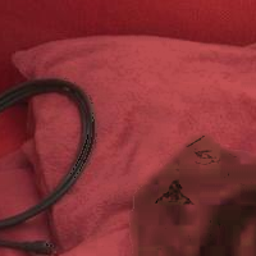}}}
        \end{minipage}   
    \end{minipage}

    \begin{minipage}[b]{1.0\linewidth}
        \begin{minipage}[b]{0.137\linewidth}
            \centering
            \centerline{
            \stackinset{l}{0pt}{t}{0pt}{%
                \colorbox{gray}{%
                    \textcolor{white}{\textbf{\footnotesize (h)}}%
                }%
            }{\includegraphics[width=\linewidth]{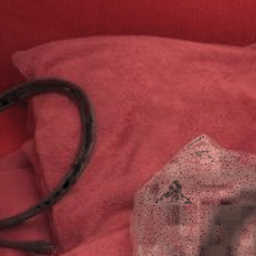}}}
        \end{minipage}   
        \hfill
        \begin{minipage}[b]{0.137\linewidth}
            \centering
            \centerline{
            \stackinset{l}{0pt}{t}{0pt}{%
                \colorbox{gray}{%
                    \textcolor{white}{\textbf{\footnotesize (i)}}%
                }%
            }{\includegraphics[width=\linewidth]{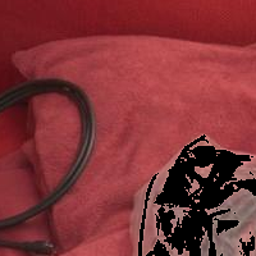}}}
        \end{minipage}   
        \hfill
        \begin{minipage}[b]{0.137\linewidth}
            \centering
            \centerline{
            \stackinset{l}{0pt}{t}{0pt}{%
                \colorbox{gray}{%
                    \textcolor{white}{\textbf{\footnotesize (j)}}%
                }%
            }{\includegraphics[width=\linewidth]{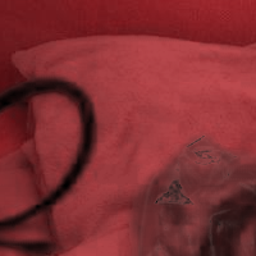}}}
        \end{minipage}   
        \hfill
        \begin{minipage}[b]{0.137\linewidth}
            \centering
            \centerline{
            \stackinset{l}{0pt}{t}{0pt}{%
                \colorbox{gray}{%
                    \textcolor{white}{\textbf{\footnotesize (k)}}%
                }%
            }{\includegraphics[width=\linewidth]{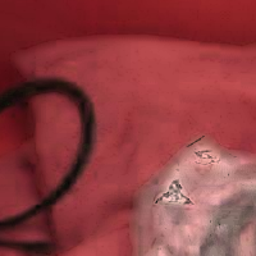}}}
        \end{minipage}
        \hfill
        \begin{minipage}[b]{0.137\linewidth}
            \centering
            \centerline{
            \stackinset{l}{0pt}{t}{0pt}{%
                \colorbox{gray}{%
                    \textcolor{white}{\textbf{\footnotesize (l)}}%
                }%
            }{\includegraphics[width=\linewidth]{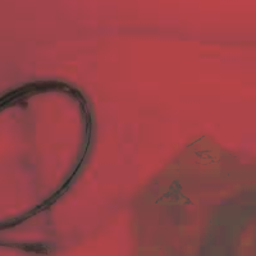}}}
        \end{minipage}
        \hfill
        \begin{minipage}[b]{0.137\linewidth}
            \centering
            \centerline{
            \stackinset{l}{0pt}{t}{0pt}{%
                \colorbox{gray}{%
                    \textcolor{white}{\textbf{\footnotesize (m)}}%
                }%
            }{\includegraphics[width=\linewidth]{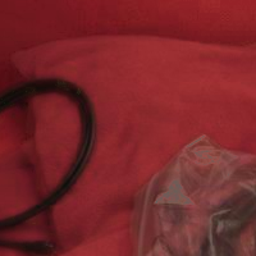}}}
        \end{minipage}   
        \hfill
        \begin{minipage}[b]{0.137\linewidth}
            \centering
            \centerline{
            \stackinset{l}{0pt}{t}{0pt}{%
                \colorbox{gray}{%
                    \textcolor{white}{\textbf{\footnotesize (n)}}%
                }%
            }{\includegraphics[width=\linewidth]{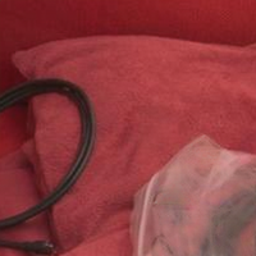}}}
        \end{minipage} 
    \end{minipage}

    \begin{minipage}[b]{1.0\linewidth}
        \begin{minipage}[b]{0.137\linewidth}
            \centering
            \centerline{
            \stackinset{l}{0pt}{t}{0pt}{%
                \colorbox{gray}{%
                    \textcolor{white}{\textbf{\footnotesize (o)}}%
                }%
            }{\includegraphics[width=\linewidth]{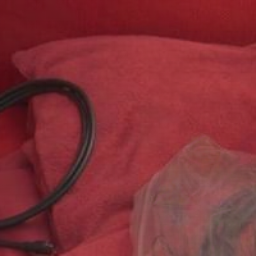}}}
        \end{minipage}   
        \hfill
        \begin{minipage}[b]{0.137\linewidth}
            \centering
            \centerline{
            \stackinset{l}{0pt}{t}{0pt}{%
                \colorbox{gray}{%
                    \textcolor{white}{\textbf{\footnotesize (p)}}%
                }%
            }{\includegraphics[width=\linewidth]{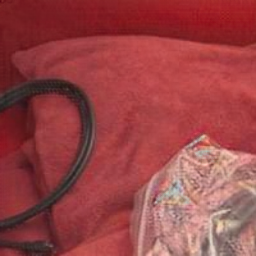}}}
        \end{minipage}   
        \hfill
        \begin{minipage}[b]{0.137\linewidth}
            \centering
            \centerline{
            \stackinset{l}{0pt}{t}{0pt}{%
                \colorbox{gray}{%
                    \textcolor{white}{\textbf{\footnotesize (q)}}%
                }%
            }{\includegraphics[width=\linewidth]{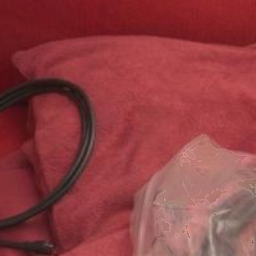}}}
        \end{minipage}
        \hfill
        \begin{minipage}[b]{0.137\linewidth}
            \centering
            \centerline{
            \stackinset{l}{0pt}{t}{0pt}{%
                \colorbox{gray}{%
                    \textcolor{white}{\textbf{\footnotesize (r)}}%
                }%
            }{\includegraphics[width=\linewidth]{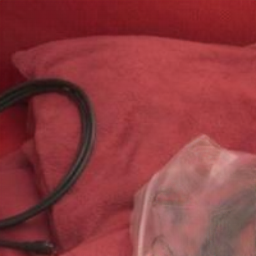}}}
        \end{minipage}
        \hfill
        \begin{minipage}[b]{0.137\linewidth}
            \centering
            \centerline{
            \stackinset{l}{0pt}{t}{0pt}{%
                \colorbox{gray}{%
                    \textcolor{white}{\textbf{\footnotesize (s)}}%
                }%
            }{\includegraphics[width=\linewidth]{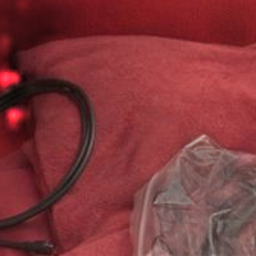}}}
        \end{minipage}
        \hfill
        \begin{minipage}[b]{0.137\linewidth}
            \centering
            \centerline{
            \stackinset{l}{0pt}{t}{0pt}{%
                \colorbox{gray}{%
                    \textcolor{white}{\textbf{\footnotesize (t)}}%
                }%
            }{\includegraphics[width=\linewidth]{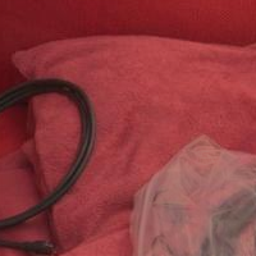}}}
        \end{minipage}
        \hfill
        \begin{minipage}[b]{0.137\linewidth}
            \centering
            \centerline{
            \stackinset{l}{0pt}{t}{0pt}{%
                \colorbox{gray}{%
                    \textcolor{white}{\textbf{\footnotesize (u)}}%
                }%
            }{\includegraphics[width=\linewidth]{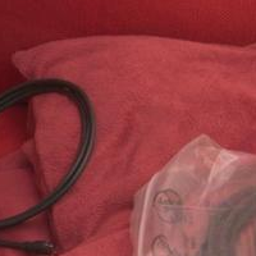}}}
        \end{minipage}
    \end{minipage}

    \caption{{\small Comprehensive visual comparison. (a) Input specular highlight image, (b) Tan [10], (c) Yoon [31], (d) Shen [11], (e) Shen [12], (f) Yang [13], (g) Shen [14], (h) Akashi [15], (i) Huo [32], (j) Fu [18], (k) Yamamoto [19], (l) Saha [20], (m) SLRR [22], (n) JSHDR [6], (o) SpecularityNet [5], (p) MG-CycleGAN [26], (q) Wu [25], (r) TSHRNet [7], (s) AHA [28], (t) Ours, (u) GT diffuse image. The reader is encouraged to zoom-in.}}
    \label{fig:full_compare1}
\end{figure*}

\begin{figure*}[hb]
    \begin{minipage}[b]{1.0\linewidth}
        \begin{minipage}[b]{0.137\linewidth}
            \centering
            \centerline{
            \stackinset{l}{0pt}{t}{0pt}{%
                \colorbox{gray}{%
                    \textcolor{white}{\textbf{\footnotesize (a)}}%
                }%
            }{\includegraphics[width=\linewidth]{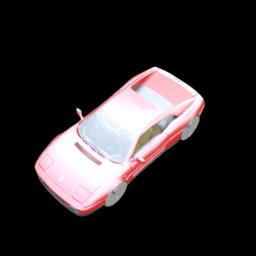}}}
        \end{minipage}   
        \hfill
        \begin{minipage}[b]{0.137\linewidth}
            \centering
            \centerline{
            \stackinset{l}{0pt}{t}{0pt}{%
                \colorbox{gray}{%
                    \textcolor{white}{\textbf{\footnotesize (b)}}%
                }%
            }{\includegraphics[width=\linewidth]{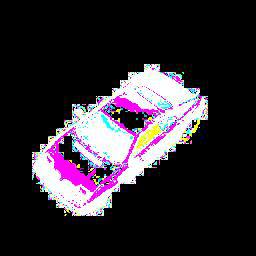}}}
        \end{minipage}   
        \hfill
        \begin{minipage}[b]{0.137\linewidth}
            \centering
            \centerline{
            \stackinset{l}{0pt}{t}{0pt}{%
                \colorbox{gray}{%
                    \textcolor{white}{\textbf{\footnotesize (c)}}%
                }%
            }{\includegraphics[width=\linewidth]{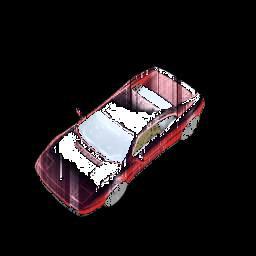}}}
        \end{minipage}   
        \hfill
        \begin{minipage}[b]{0.137\linewidth}
            \centering
            \centerline{
            \stackinset{l}{0pt}{t}{0pt}{%
                \colorbox{gray}{%
                    \textcolor{white}{\textbf{\footnotesize (d)}}%
                }%
            }{\includegraphics[width=\linewidth]{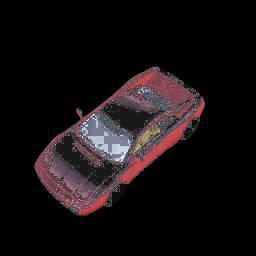}}}
        \end{minipage}   
        \hfill
        \begin{minipage}[b]{0.137\linewidth}
            \centering
            \centerline{
            \stackinset{l}{0pt}{t}{0pt}{%
                \colorbox{gray}{%
                    \textcolor{white}{\textbf{\footnotesize (e)}}%
                }%
            }{\includegraphics[width=\linewidth]{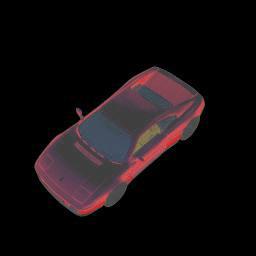}}}
        \end{minipage}
        \hfill
        \begin{minipage}[b]{0.137\linewidth}
            \centering
            \centerline{
            \stackinset{l}{0pt}{t}{0pt}{%
                \colorbox{gray}{%
                    \textcolor{white}{\textbf{\footnotesize (f)}}%
                }%
            }{\includegraphics[width=\linewidth]{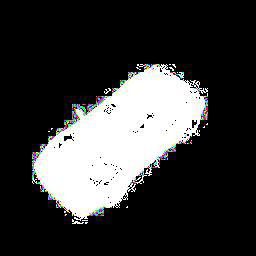}}}
        \end{minipage}
        \hfill
        \begin{minipage}[b]{0.137\linewidth}
            \centering
            \centerline{
            \stackinset{l}{0pt}{t}{0pt}{%
                \colorbox{gray}{%
                    \textcolor{white}{\textbf{\footnotesize (g)}}%
                }%
            }{\includegraphics[width=\linewidth]{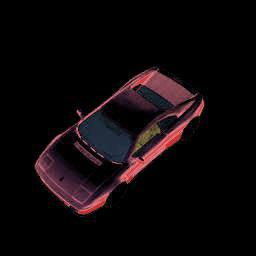}}}
        \end{minipage}   
    \end{minipage}

    \begin{minipage}[b]{1.0\linewidth}
        \begin{minipage}[b]{0.137\linewidth}
            \centering
            \centerline{
            \stackinset{l}{0pt}{t}{0pt}{%
                \colorbox{gray}{%
                    \textcolor{white}{\textbf{\footnotesize (h)}}%
                }%
            }{\includegraphics[width=\linewidth]{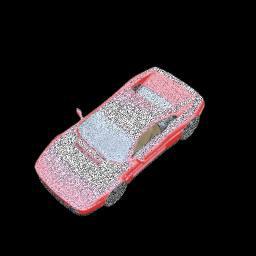}}}
        \end{minipage}   
        \hfill
        \begin{minipage}[b]{0.137\linewidth}
            \centering
            \centerline{
            \stackinset{l}{0pt}{t}{0pt}{%
                \colorbox{gray}{%
                    \textcolor{white}{\textbf{\footnotesize (i)}}%
                }%
            }{\includegraphics[width=\linewidth]{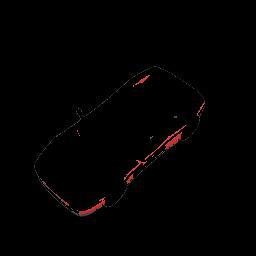}}}
        \end{minipage}   
        \hfill
        \begin{minipage}[b]{0.137\linewidth}
            \centering
            \centerline{
            \stackinset{l}{0pt}{t}{0pt}{%
                \colorbox{gray}{%
                    \textcolor{white}{\textbf{\footnotesize (j)}}%
                }%
            }{\includegraphics[width=\linewidth]{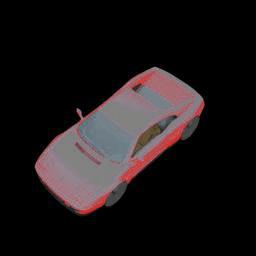}}}
        \end{minipage}   
        \hfill
        \begin{minipage}[b]{0.137\linewidth}
            \centering
            \centerline{
            \stackinset{l}{0pt}{t}{0pt}{%
                \colorbox{gray}{%
                    \textcolor{white}{\textbf{\footnotesize (k)}}%
                }%
            }{\includegraphics[width=\linewidth]{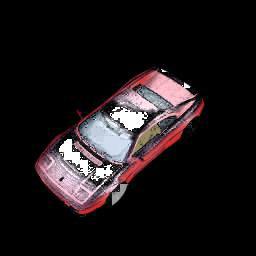}}}
        \end{minipage}
        \hfill
        \begin{minipage}[b]{0.137\linewidth}
            \centering
            \centerline{
            \stackinset{l}{0pt}{t}{0pt}{%
                \colorbox{gray}{%
                    \textcolor{white}{\textbf{\footnotesize (l)}}%
                }%
            }{\includegraphics[width=\linewidth]{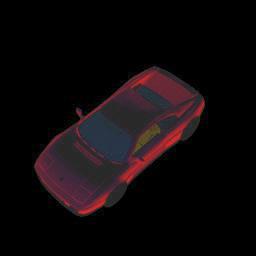}}}
        \end{minipage}
        \hfill
        \begin{minipage}[b]{0.137\linewidth}
            \centering
            \centerline{
            \stackinset{l}{0pt}{t}{0pt}{%
                \colorbox{gray}{%
                    \textcolor{white}{\textbf{\footnotesize (m)}}%
                }%
            }{\includegraphics[width=\linewidth]{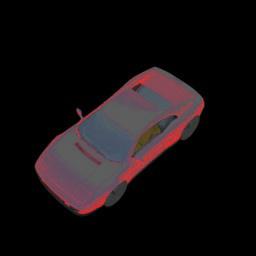}}}
        \end{minipage}   
        \hfill
        \begin{minipage}[b]{0.137\linewidth}
            \centering
            \centerline{
            \stackinset{l}{0pt}{t}{0pt}{%
                \colorbox{gray}{%
                    \textcolor{white}{\textbf{\footnotesize (n)}}%
                }%
            }{\includegraphics[width=\linewidth]{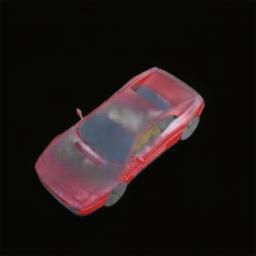}}}
        \end{minipage} 
    \end{minipage}

    \begin{minipage}[b]{1.0\linewidth}
        \begin{minipage}[b]{0.137\linewidth}
            \centering
            \centerline{
            \stackinset{l}{0pt}{t}{0pt}{%
                \colorbox{gray}{%
                    \textcolor{white}{\textbf{\footnotesize (o)}}%
                }%
            }{\includegraphics[width=\linewidth]{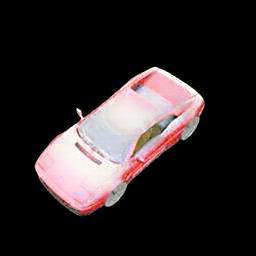}}}
        \end{minipage}   
        \hfill
        \begin{minipage}[b]{0.137\linewidth}
            \centering
            \centerline{
            \stackinset{l}{0pt}{t}{0pt}{%
                \colorbox{gray}{%
                    \textcolor{white}{\textbf{\footnotesize (p)}}%
                }%
            }{\includegraphics[width=\linewidth]{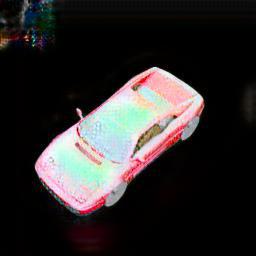}}}
        \end{minipage}   
        \hfill
        \begin{minipage}[b]{0.137\linewidth}
            \centering
            \centerline{
            \stackinset{l}{0pt}{t}{0pt}{%
                \colorbox{gray}{%
                    \textcolor{white}{\textbf{\footnotesize (q)}}%
                }%
            }{\includegraphics[width=\linewidth]{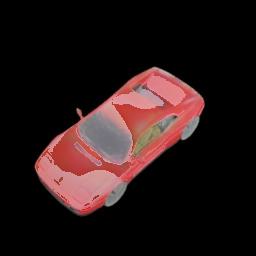}}}
        \end{minipage}
        \hfill
        \begin{minipage}[b]{0.137\linewidth}
            \centering
            \centerline{
            \stackinset{l}{0pt}{t}{0pt}{%
                \colorbox{gray}{%
                    \textcolor{white}{\textbf{\footnotesize (r)}}%
                }%
            }{\includegraphics[width=\linewidth]{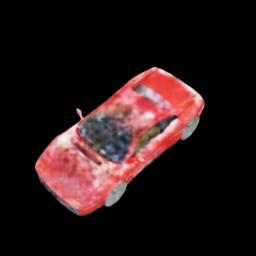}}}
        \end{minipage}
        \hfill
        \begin{minipage}[b]{0.137\linewidth}
            \centering
            \centerline{
            \stackinset{l}{0pt}{t}{0pt}{%
                \colorbox{gray}{%
                    \textcolor{white}{\textbf{\footnotesize (s)}}%
                }%
            }{\includegraphics[width=\linewidth]{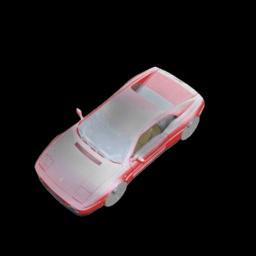}}}
        \end{minipage}
        \hfill
        \begin{minipage}[b]{0.137\linewidth}
            \centering
            \centerline{
            \stackinset{l}{0pt}{t}{0pt}{%
                \colorbox{gray}{%
                    \textcolor{white}{\textbf{\footnotesize (t)}}%
                }%
            }{\includegraphics[width=\linewidth]{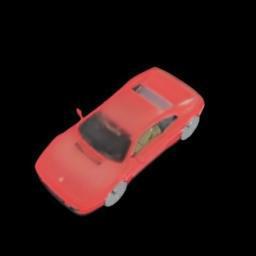}}}
        \end{minipage}
        \hfill
        \begin{minipage}[b]{0.137\linewidth}
            \centering
            \centerline{
            \stackinset{l}{0pt}{t}{0pt}{%
                \colorbox{gray}{%
                    \textcolor{white}{\textbf{\footnotesize (u)}}%
                }%
            }{\includegraphics[width=\linewidth]{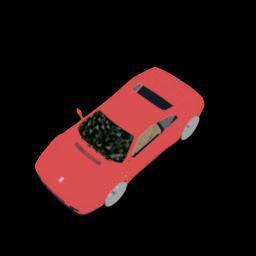}}}
        \end{minipage}
    \end{minipage}

    \begin{minipage}[b]{1.0\linewidth}
        \begin{minipage}[b]{0.137\linewidth}
            \centering
            \centerline{
            \stackinset{l}{0pt}{t}{0pt}{%
                \colorbox{gray}{%
                    \textcolor{white}{\textbf{\footnotesize (a)}}%
                }%
            }{\includegraphics[width=\linewidth]{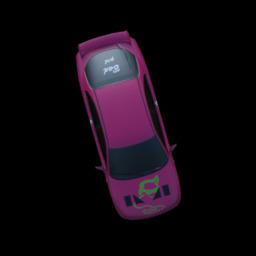}}}
        \end{minipage}   
        \hfill
        \begin{minipage}[b]{0.137\linewidth}
            \centering
            \centerline{
            \stackinset{l}{0pt}{t}{0pt}{%
                \colorbox{gray}{%
                    \textcolor{white}{\textbf{\footnotesize (b)}}%
                }%
            }{\includegraphics[width=\linewidth]{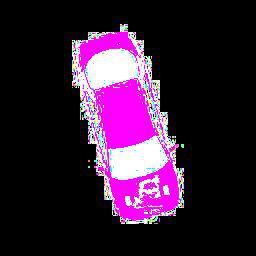}}}
        \end{minipage}   
        \hfill
        \begin{minipage}[b]{0.137\linewidth}
            \centering
            \centerline{
            \stackinset{l}{0pt}{t}{0pt}{%
                \colorbox{gray}{%
                    \textcolor{white}{\textbf{\footnotesize (c)}}%
                }%
            }{\includegraphics[width=\linewidth]{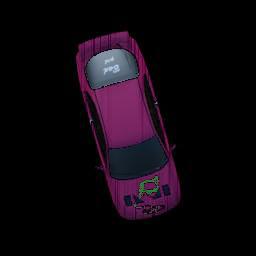}}}
        \end{minipage}   
        \hfill
        \begin{minipage}[b]{0.137\linewidth}
            \centering
            \centerline{
            \stackinset{l}{0pt}{t}{0pt}{%
                \colorbox{gray}{%
                    \textcolor{white}{\textbf{\footnotesize (d)}}%
                }%
            }{\includegraphics[width=\linewidth]{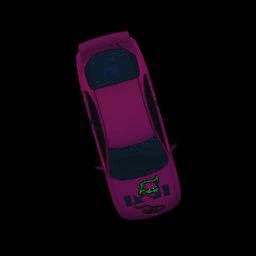}}}
        \end{minipage}   
        \hfill
        \begin{minipage}[b]{0.137\linewidth}
            \centering
            \centerline{
            \stackinset{l}{0pt}{t}{0pt}{%
                \colorbox{gray}{%
                    \textcolor{white}{\textbf{\footnotesize (e)}}%
                }%
            }{\includegraphics[width=\linewidth]{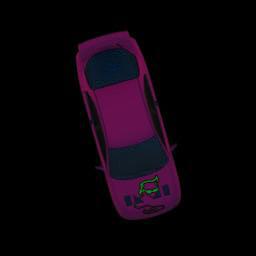}}}
        \end{minipage}
        \hfill
        \begin{minipage}[b]{0.137\linewidth}
            \centering
            \centerline{
            \stackinset{l}{0pt}{t}{0pt}{%
                \colorbox{gray}{%
                    \textcolor{white}{\textbf{\footnotesize (f)}}%
                }%
            }{\includegraphics[width=\linewidth]{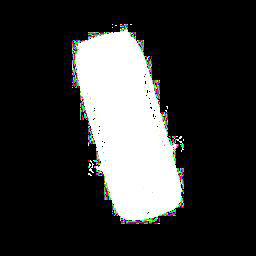}}}
        \end{minipage}
        \hfill
        \begin{minipage}[b]{0.137\linewidth}
            \centering
            \centerline{
            \stackinset{l}{0pt}{t}{0pt}{%
                \colorbox{gray}{%
                    \textcolor{white}{\textbf{\footnotesize (g)}}%
                }%
            }{\includegraphics[width=\linewidth]{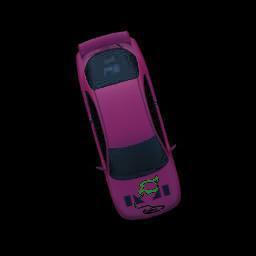}}}
        \end{minipage}   
    \end{minipage}

    \begin{minipage}[b]{1.0\linewidth}
        \begin{minipage}[b]{0.137\linewidth}
            \centering
            \centerline{
            \stackinset{l}{0pt}{t}{0pt}{%
                \colorbox{gray}{%
                    \textcolor{white}{\textbf{\footnotesize (h)}}%
                }%
            }{\includegraphics[width=\linewidth]{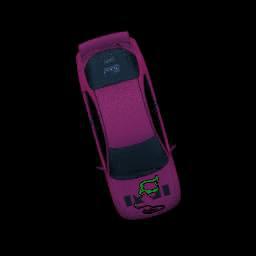}}}
        \end{minipage}   
        \hfill
        \begin{minipage}[b]{0.137\linewidth}
            \centering
            \centerline{
            \stackinset{l}{0pt}{t}{0pt}{%
                \colorbox{gray}{%
                    \textcolor{white}{\textbf{\footnotesize (i)}}%
                }%
            }{\includegraphics[width=\linewidth]{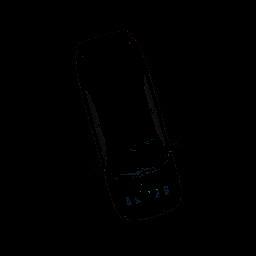}}}
        \end{minipage}   
        \hfill
        \begin{minipage}[b]{0.137\linewidth}
            \centering
            \centerline{
            \stackinset{l}{0pt}{t}{0pt}{%
                \colorbox{gray}{%
                    \textcolor{white}{\textbf{\footnotesize (j)}}%
                }%
            }{\includegraphics[width=\linewidth]{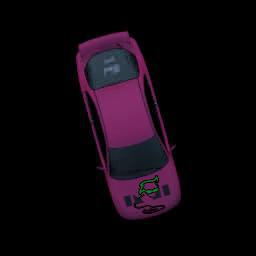}}}
        \end{minipage}   
        \hfill
        \begin{minipage}[b]{0.137\linewidth}
            \centering
            \centerline{
            \stackinset{l}{0pt}{t}{0pt}{%
                \colorbox{gray}{%
                    \textcolor{white}{\textbf{\footnotesize (k)}}%
                }%
            }{\includegraphics[width=\linewidth]{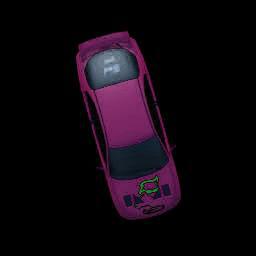}}}
        \end{minipage}
        \hfill
        \begin{minipage}[b]{0.137\linewidth}
            \centering
            \centerline{
            \stackinset{l}{0pt}{t}{0pt}{%
                \colorbox{gray}{%
                    \textcolor{white}{\textbf{\footnotesize (l)}}%
                }%
            }{\includegraphics[width=\linewidth]{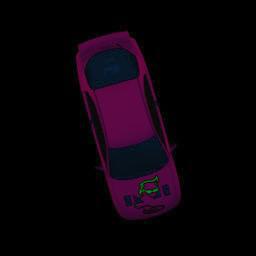}}}
        \end{minipage}
        \hfill
        \begin{minipage}[b]{0.137\linewidth}
            \centering
            \centerline{
            \stackinset{l}{0pt}{t}{0pt}{%
                \colorbox{gray}{%
                    \textcolor{white}{\textbf{\footnotesize (m)}}%
                }%
            }{\includegraphics[width=\linewidth]{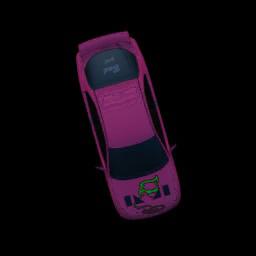}}}
        \end{minipage}   
        \hfill
        \begin{minipage}[b]{0.137\linewidth}
            \centering
            \centerline{
            \stackinset{l}{0pt}{t}{0pt}{%
                \colorbox{gray}{%
                    \textcolor{white}{\textbf{\footnotesize (n)}}%
                }%
            }{\includegraphics[width=\linewidth]{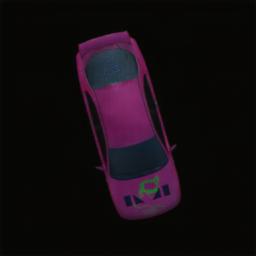}}}
        \end{minipage} 
    \end{minipage}

    \begin{minipage}[b]{1.0\linewidth}
        \begin{minipage}[b]{0.137\linewidth}
            \centering
            \centerline{
            \stackinset{l}{0pt}{t}{0pt}{%
                \colorbox{gray}{%
                    \textcolor{white}{\textbf{\footnotesize (o)}}%
                }%
            }{\includegraphics[width=\linewidth]{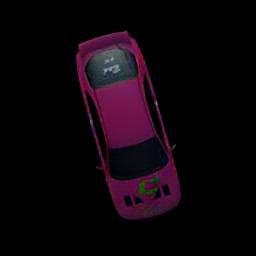}}}
        \end{minipage}   
        \hfill
        \begin{minipage}[b]{0.137\linewidth}
            \centering
            \centerline{
            \stackinset{l}{0pt}{t}{0pt}{%
                \colorbox{gray}{%
                    \textcolor{white}{\textbf{\footnotesize (p)}}%
                }%
            }{\includegraphics[width=\linewidth]{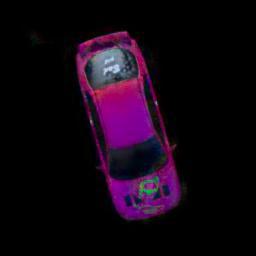}}}
        \end{minipage}   
        \hfill
        \begin{minipage}[b]{0.137\linewidth}
            \centering
            \centerline{
            \stackinset{l}{0pt}{t}{0pt}{%
                \colorbox{gray}{%
                    \textcolor{white}{\textbf{\footnotesize (q)}}%
                }%
            }{\includegraphics[width=\linewidth]{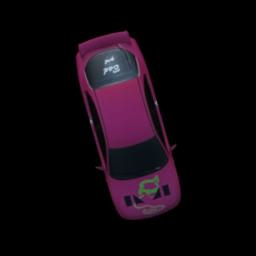}}}
        \end{minipage}
        \hfill
        \begin{minipage}[b]{0.137\linewidth}
            \centering
            \centerline{
            \stackinset{l}{0pt}{t}{0pt}{%
                \colorbox{gray}{%
                    \textcolor{white}{\textbf{\footnotesize (r)}}%
                }%
            }{\includegraphics[width=\linewidth]{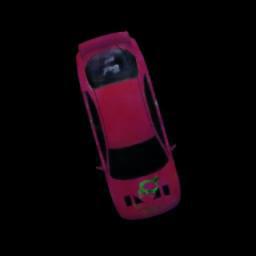}}}
        \end{minipage}
        \hfill
        \begin{minipage}[b]{0.137\linewidth}
            \centering
            \centerline{
            \stackinset{l}{0pt}{t}{0pt}{%
                \colorbox{gray}{%
                    \textcolor{white}{\textbf{\footnotesize (s)}}%
                }%
            }{\includegraphics[width=\linewidth]{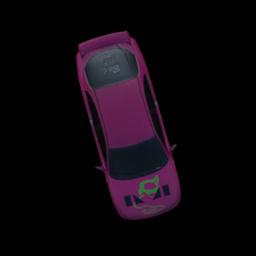}}}
        \end{minipage}
        \hfill
        \begin{minipage}[b]{0.137\linewidth}
            \centering
            \centerline{
            \stackinset{l}{0pt}{t}{0pt}{%
                \colorbox{gray}{%
                    \textcolor{white}{\textbf{\footnotesize (t)}}%
                }%
            }{\includegraphics[width=\linewidth]{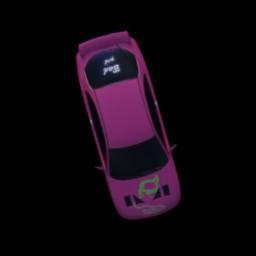}}}
        \end{minipage}
        \hfill
        \begin{minipage}[b]{0.137\linewidth}
            \centering
            \centerline{
            \stackinset{l}{0pt}{t}{0pt}{%
                \colorbox{gray}{%
                    \textcolor{white}{\textbf{\footnotesize (u)}}%
                }%
            }{\includegraphics[width=\linewidth]{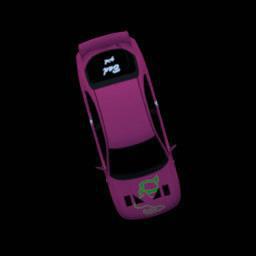}}}
        \end{minipage}
    \end{minipage}

    \begin{minipage}[b]{1.0\linewidth}
        \begin{minipage}[b]{0.137\linewidth}
            \centering
            \centerline{
            \stackinset{l}{0pt}{t}{0pt}{%
                \colorbox{gray}{%
                    \textcolor{white}{\textbf{\footnotesize (a)}}%
                }%
            }{\includegraphics[width=\linewidth]{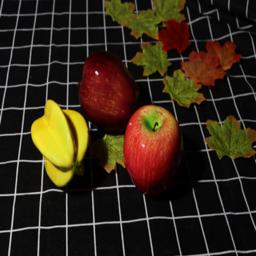}}}
        \end{minipage}   
        \hfill
        \begin{minipage}[b]{0.137\linewidth}
            \centering
            \centerline{
            \stackinset{l}{0pt}{t}{0pt}{%
                \colorbox{gray}{%
                    \textcolor{white}{\textbf{\footnotesize (b)}}%
                }%
            }{\includegraphics[width=\linewidth]{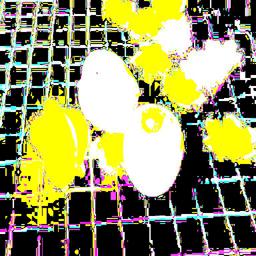}}}
        \end{minipage}   
        \hfill
        \begin{minipage}[b]{0.137\linewidth}
            \centering
            \centerline{
            \stackinset{l}{0pt}{t}{0pt}{%
                \colorbox{gray}{%
                    \textcolor{white}{\textbf{\footnotesize (c)}}%
                }%
            }{\includegraphics[width=\linewidth]{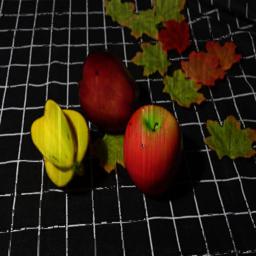}}}
        \end{minipage}   
        \hfill
        \begin{minipage}[b]{0.137\linewidth}
            \centering
            \centerline{
            \stackinset{l}{0pt}{t}{0pt}{%
                \colorbox{gray}{%
                    \textcolor{white}{\textbf{\footnotesize (d)}}%
                }%
            }{\includegraphics[width=\linewidth]{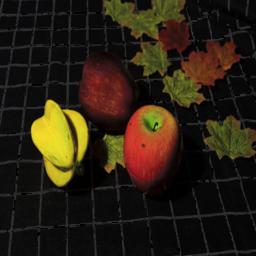}}}
        \end{minipage}   
        \hfill
        \begin{minipage}[b]{0.137\linewidth}
            \centering
            \centerline{
            \stackinset{l}{0pt}{t}{0pt}{%
                \colorbox{gray}{%
                    \textcolor{white}{\textbf{\footnotesize (e)}}%
                }%
            }{\includegraphics[width=\linewidth]{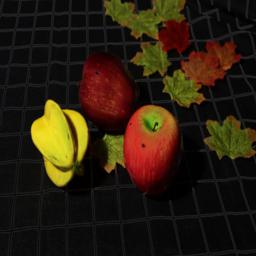}}}
        \end{minipage}
        \hfill
        \begin{minipage}[b]{0.137\linewidth}
            \centering
            \centerline{
            \stackinset{l}{0pt}{t}{0pt}{%
                \colorbox{gray}{%
                    \textcolor{white}{\textbf{\footnotesize (f)}}%
                }%
            }{\includegraphics[width=\linewidth]{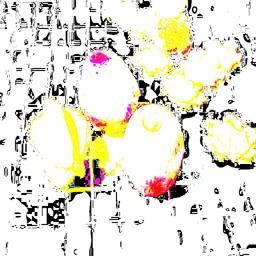}}}
        \end{minipage}
        \hfill
        \begin{minipage}[b]{0.137\linewidth}
            \centering
            \centerline{
            \stackinset{l}{0pt}{t}{0pt}{%
                \colorbox{gray}{%
                    \textcolor{white}{\textbf{\footnotesize (g)}}%
                }%
            }{\includegraphics[width=\linewidth]{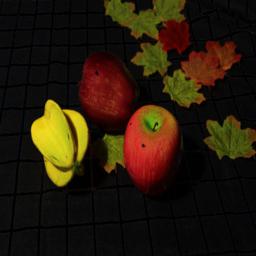}}}
        \end{minipage}   
    \end{minipage}

    \begin{minipage}[b]{1.0\linewidth}
        \begin{minipage}[b]{0.137\linewidth}
            \centering
            \centerline{
            \stackinset{l}{0pt}{t}{0pt}{%
                \colorbox{gray}{%
                    \textcolor{white}{\textbf{\footnotesize (h)}}%
                }%
            }{\includegraphics[width=\linewidth]{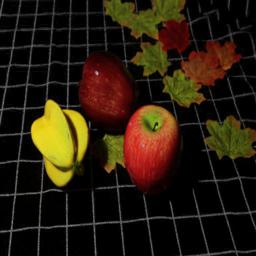}}}
        \end{minipage}   
        \hfill
        \begin{minipage}[b]{0.137\linewidth}
            \centering
            \centerline{
            \stackinset{l}{0pt}{t}{0pt}{%
                \colorbox{gray}{%
                    \textcolor{white}{\textbf{\footnotesize (i)}}%
                }%
            }{\includegraphics[width=\linewidth]{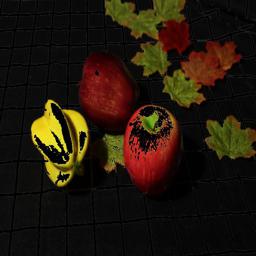}}}
        \end{minipage}   
        \hfill
        \begin{minipage}[b]{0.137\linewidth}
            \centering
            \centerline{
            \stackinset{l}{0pt}{t}{0pt}{%
                \colorbox{gray}{%
                    \textcolor{white}{\textbf{\footnotesize (j)}}%
                }%
            }{\includegraphics[width=\linewidth]{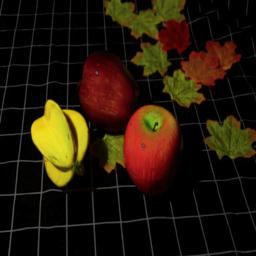}}}
        \end{minipage}   
        \hfill
        \begin{minipage}[b]{0.137\linewidth}
            \centering
            \centerline{
            \stackinset{l}{0pt}{t}{0pt}{%
                \colorbox{gray}{%
                    \textcolor{white}{\textbf{\footnotesize (k)}}%
                }%
            }{\includegraphics[width=\linewidth]{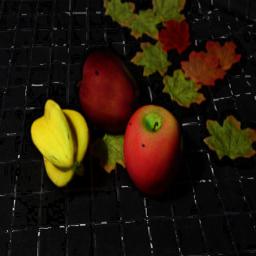}}}
        \end{minipage}
        \hfill
        \begin{minipage}[b]{0.137\linewidth}
            \centering
            \centerline{
            \stackinset{l}{0pt}{t}{0pt}{%
                \colorbox{gray}{%
                    \textcolor{white}{\textbf{\footnotesize (l)}}%
                }%
            }{\includegraphics[width=\linewidth]{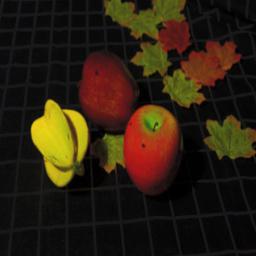}}}
        \end{minipage}
        \hfill
        \begin{minipage}[b]{0.137\linewidth}
            \centering
            \centerline{
            \stackinset{l}{0pt}{t}{0pt}{%
                \colorbox{gray}{%
                    \textcolor{white}{\textbf{\footnotesize (m)}}%
                }%
            }{\includegraphics[width=\linewidth]{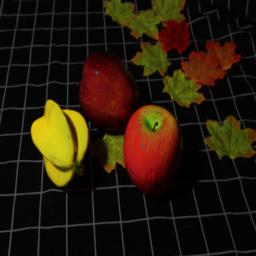}}}
        \end{minipage}   
        \hfill
        \begin{minipage}[b]{0.137\linewidth}
            \centering
            \centerline{
            \stackinset{l}{0pt}{t}{0pt}{%
                \colorbox{gray}{%
                    \textcolor{white}{\textbf{\footnotesize (n)}}%
                }%
            }{\includegraphics[width=\linewidth]{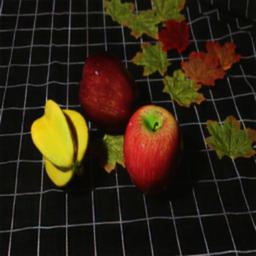}}}
        \end{minipage} 
    \end{minipage}

    \begin{minipage}[b]{1.0\linewidth}
        \begin{minipage}[b]{0.137\linewidth}
            \centering
            \centerline{
            \stackinset{l}{0pt}{t}{0pt}{%
                \colorbox{gray}{%
                    \textcolor{white}{\textbf{\footnotesize (o)}}%
                }%
            }{\includegraphics[width=\linewidth]{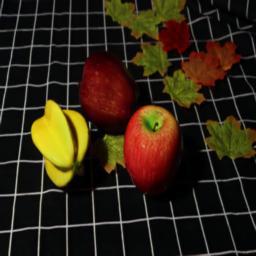}}}
        \end{minipage}   
        \hfill
        \begin{minipage}[b]{0.137\linewidth}
            \centering
            \centerline{
            \stackinset{l}{0pt}{t}{0pt}{%
                \colorbox{gray}{%
                    \textcolor{white}{\textbf{\footnotesize (p)}}%
                }%
            }{\includegraphics[width=\linewidth]{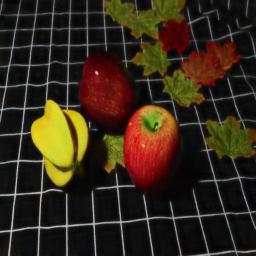}}}
        \end{minipage}   
        \hfill
        \begin{minipage}[b]{0.137\linewidth}
            \centering
            \centerline{
            \stackinset{l}{0pt}{t}{0pt}{%
                \colorbox{gray}{%
                    \textcolor{white}{\textbf{\footnotesize (q)}}%
                }%
            }{\includegraphics[width=\linewidth]{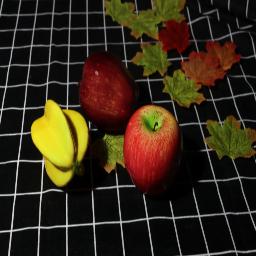}}}
        \end{minipage}
        \hfill
        \begin{minipage}[b]{0.137\linewidth}
            \centering
            \centerline{
            \stackinset{l}{0pt}{t}{0pt}{%
                \colorbox{gray}{%
                    \textcolor{white}{\textbf{\footnotesize (r)}}%
                }%
            }{\includegraphics[width=\linewidth]{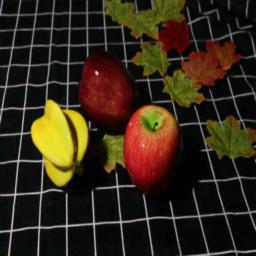}}}
        \end{minipage}
        \hfill
        \begin{minipage}[b]{0.137\linewidth}
            \centering
            \centerline{
            \stackinset{l}{0pt}{t}{0pt}{%
                \colorbox{gray}{%
                    \textcolor{white}{\textbf{\footnotesize (s)}}%
                }%
            }{\includegraphics[width=\linewidth]{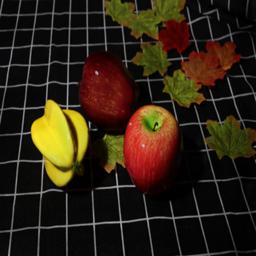}}}
        \end{minipage}
        \hfill
        \begin{minipage}[b]{0.137\linewidth}
            \centering
            \centerline{
            \stackinset{l}{0pt}{t}{0pt}{%
                \colorbox{gray}{%
                    \textcolor{white}{\textbf{\footnotesize (t)}}%
                }%
            }{\includegraphics[width=\linewidth]{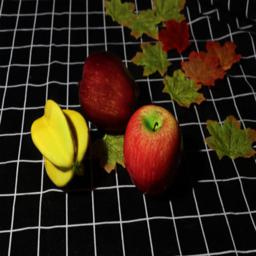}}}
        \end{minipage}
        \hfill
        \begin{minipage}[b]{0.137\linewidth}
            \centering
            \centerline{
            \stackinset{l}{0pt}{t}{0pt}{%
                \colorbox{gray}{%
                    \textcolor{white}{\textbf{\footnotesize (u)}}%
                }%
            }{\includegraphics[width=\linewidth]{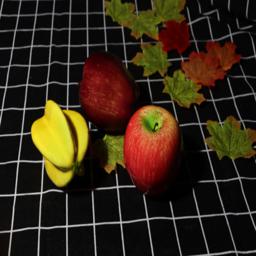}}}
        \end{minipage}
    \end{minipage}

    \caption{{\small Comprehensive visual comparison. (a) Input specular highlight image, (b) Tan [10], (c) Yoon [31], (d) Shen [11], (e) Shen [12], (f) Yang [13], (g) Shen [14], (h) Akashi [15], (i) Huo [32], (j) Fu [18], (k) Yamamoto [19], (l) Saha [20], (m) SLRR [22], (n) JSHDR [6], (o) SpecularityNet [5], (p) MG-CycleGAN [26], (q) Wu [25], (r) TSHRNet [7], (s) AHA [28], (t) Ours, (u) GT diffuse image. The reader is encouraged to zoom-in.}}
    \label{fig:full_compare2}
\end{figure*}

\begin{figure*}[hb]
    \begin{minipage}[b]{1.0\linewidth}
        \begin{minipage}[b]{0.137\linewidth}
            \centering
            \centerline{
            \stackinset{l}{0pt}{t}{0pt}{%
                \colorbox{gray}{%
                    \textcolor{white}{\textbf{\footnotesize (a)}}%
                }%
            }{\includegraphics[width=\linewidth]{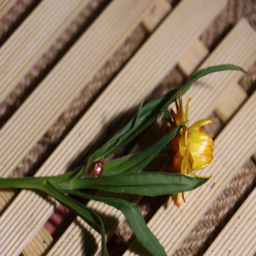}}}
        \end{minipage}   
        \hfill
        \begin{minipage}[b]{0.137\linewidth}
            \centering
            \centerline{
            \stackinset{l}{0pt}{t}{0pt}{%
                \colorbox{gray}{%
                    \textcolor{white}{\textbf{\footnotesize (b)}}%
                }%
            }{\includegraphics[width=\linewidth]{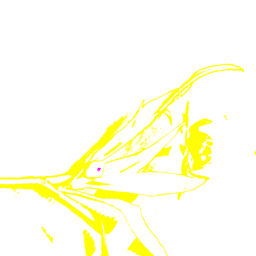}}}
        \end{minipage}   
        \hfill
        \begin{minipage}[b]{0.137\linewidth}
            \centering
            \centerline{
            \stackinset{l}{0pt}{t}{0pt}{%
                \colorbox{gray}{%
                    \textcolor{white}{\textbf{\footnotesize (c)}}%
                }%
            }{\includegraphics[width=\linewidth]{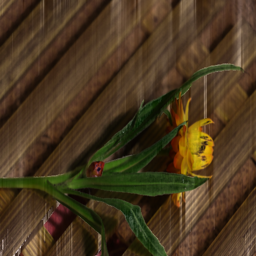}}}
        \end{minipage}   
        \hfill
        \begin{minipage}[b]{0.137\linewidth}
            \centering
            \centerline{
            \stackinset{l}{0pt}{t}{0pt}{%
                \colorbox{gray}{%
                    \textcolor{white}{\textbf{\footnotesize (d)}}%
                }%
            }{\includegraphics[width=\linewidth]{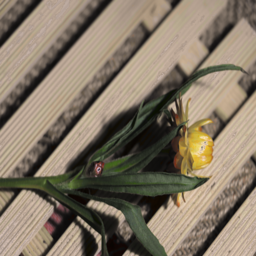}}}
        \end{minipage}   
        \hfill
        \begin{minipage}[b]{0.137\linewidth}
            \centering
            \centerline{
            \stackinset{l}{0pt}{t}{0pt}{%
                \colorbox{gray}{%
                    \textcolor{white}{\textbf{\footnotesize (e)}}%
                }%
            }{\includegraphics[width=\linewidth]{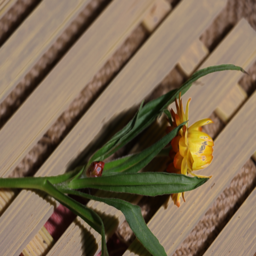}}}
        \end{minipage}
        \hfill
        \begin{minipage}[b]{0.137\linewidth}
            \centering
            \centerline{
            \stackinset{l}{0pt}{t}{0pt}{%
                \colorbox{gray}{%
                    \textcolor{white}{\textbf{\footnotesize (f)}}%
                }%
            }{\includegraphics[width=\linewidth]{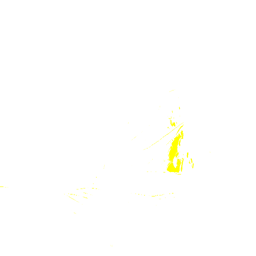}}}
        \end{minipage}
        \hfill
        \begin{minipage}[b]{0.137\linewidth}
            \centering
            \centerline{
            \stackinset{l}{0pt}{t}{0pt}{%
                \colorbox{gray}{%
                    \textcolor{white}{\textbf{\footnotesize (g)}}%
                }%
            }{\includegraphics[width=\linewidth]{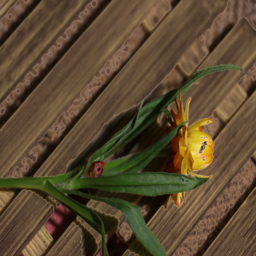}}}
        \end{minipage}   
    \end{minipage}

    \begin{minipage}[b]{1.0\linewidth}
        \begin{minipage}[b]{0.137\linewidth}
            \centering
            \centerline{
            \stackinset{l}{0pt}{t}{0pt}{%
                \colorbox{gray}{%
                    \textcolor{white}{\textbf{\footnotesize (h)}}%
                }%
            }{\includegraphics[width=\linewidth]{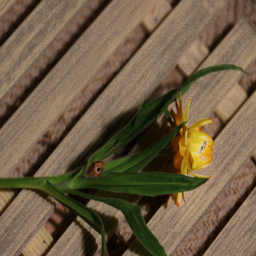}}}
        \end{minipage}   
        \hfill
        \begin{minipage}[b]{0.137\linewidth}
            \centering
            \centerline{
            \stackinset{l}{0pt}{t}{0pt}{%
                \colorbox{gray}{%
                    \textcolor{white}{\textbf{\footnotesize (i)}}%
                }%
            }{\includegraphics[width=\linewidth]{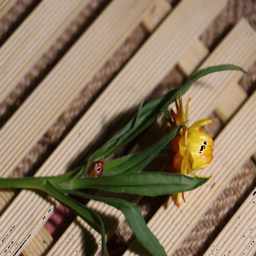}}}
        \end{minipage}   
        \hfill
        \begin{minipage}[b]{0.137\linewidth}
            \centering
            \centerline{
            \stackinset{l}{0pt}{t}{0pt}{%
                \colorbox{gray}{%
                    \textcolor{white}{\textbf{\footnotesize (j)}}%
                }%
            }{\includegraphics[width=\linewidth]{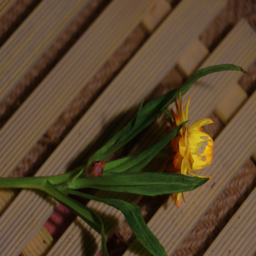}}}
        \end{minipage}   
        \hfill
        \begin{minipage}[b]{0.137\linewidth}
            \centering
            \centerline{
            \stackinset{l}{0pt}{t}{0pt}{%
                \colorbox{gray}{%
                    \textcolor{white}{\textbf{\footnotesize (k)}}%
                }%
            }{\includegraphics[width=\linewidth]{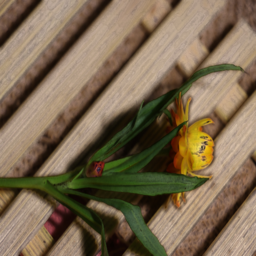}}}
        \end{minipage}
        \hfill
        \begin{minipage}[b]{0.137\linewidth}
            \centering
            \centerline{
            \stackinset{l}{0pt}{t}{0pt}{%
                \colorbox{gray}{%
                    \textcolor{white}{\textbf{\footnotesize (l)}}%
                }%
            }{\includegraphics[width=\linewidth]{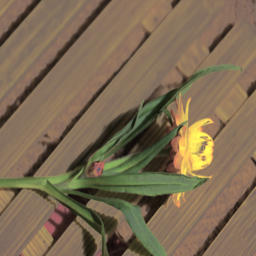}}}
        \end{minipage}
        \hfill
        \begin{minipage}[b]{0.137\linewidth}
            \centering
            \centerline{
            \stackinset{l}{0pt}{t}{0pt}{%
                \colorbox{gray}{%
                    \textcolor{white}{\textbf{\footnotesize (m)}}%
                }%
            }{\includegraphics[width=\linewidth]{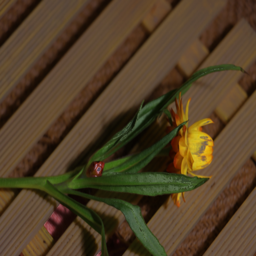}}}
        \end{minipage}   
        \hfill
        \begin{minipage}[b]{0.137\linewidth}
            \centering
            \centerline{
            \stackinset{l}{0pt}{t}{0pt}{%
                \colorbox{gray}{%
                    \textcolor{white}{\textbf{\footnotesize (n)}}%
                }%
            }{\includegraphics[width=\linewidth]{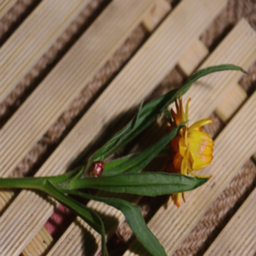}}}
        \end{minipage} 
    \end{minipage}

    \begin{minipage}[b]{1.0\linewidth}
        \begin{minipage}[b]{0.137\linewidth}
            \centering
            \centerline{
            \stackinset{l}{0pt}{t}{0pt}{%
                \colorbox{gray}{%
                    \textcolor{white}{\textbf{\footnotesize (o)}}%
                }%
            }{\includegraphics[width=\linewidth]{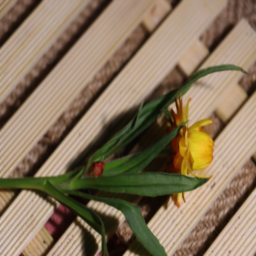}}}
        \end{minipage}   
        \hfill
        \begin{minipage}[b]{0.137\linewidth}
            \centering
            \centerline{
            \stackinset{l}{0pt}{t}{0pt}{%
                \colorbox{gray}{%
                    \textcolor{white}{\textbf{\footnotesize (p)}}%
                }%
            }{\includegraphics[width=\linewidth]{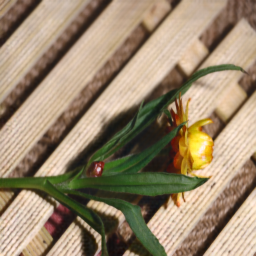}}}
        \end{minipage}   
        \hfill
        \begin{minipage}[b]{0.137\linewidth}
            \centering
            \centerline{
            \stackinset{l}{0pt}{t}{0pt}{%
                \colorbox{gray}{%
                    \textcolor{white}{\textbf{\footnotesize (q)}}%
                }%
            }{\includegraphics[width=\linewidth]{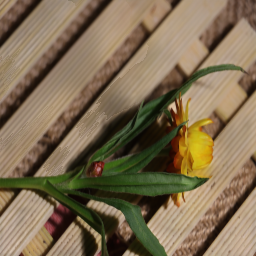}}}
        \end{minipage}
        \hfill
        \begin{minipage}[b]{0.137\linewidth}
            \centering
            \centerline{
            \stackinset{l}{0pt}{t}{0pt}{%
                \colorbox{gray}{%
                    \textcolor{white}{\textbf{\footnotesize (r)}}%
                }%
            }{\includegraphics[width=\linewidth]{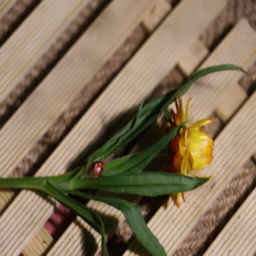}}}
        \end{minipage}
        \hfill
        \begin{minipage}[b]{0.137\linewidth}
            \centering
            \centerline{
            \stackinset{l}{0pt}{t}{0pt}{%
                \colorbox{gray}{%
                    \textcolor{white}{\textbf{\footnotesize (s)}}%
                }%
            }{\includegraphics[width=\linewidth]{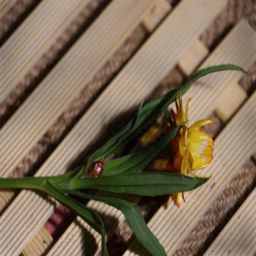}}}
        \end{minipage}
        \hfill
        \begin{minipage}[b]{0.137\linewidth}
            \centering
            \centerline{
            \stackinset{l}{0pt}{t}{0pt}{%
                \colorbox{gray}{%
                    \textcolor{white}{\textbf{\footnotesize (t)}}%
                }%
            }{\includegraphics[width=\linewidth]{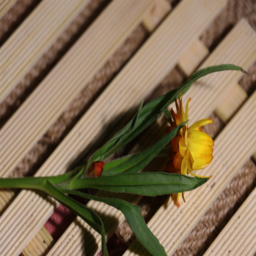}}}
        \end{minipage}
        \hfill
        \begin{minipage}[b]{0.137\linewidth}
            \centering
            \centerline{
            \stackinset{l}{0pt}{t}{0pt}{%
                \colorbox{gray}{%
                    \textcolor{white}{\textbf{\footnotesize (u)}}%
                }%
            }{\includegraphics[width=\linewidth]{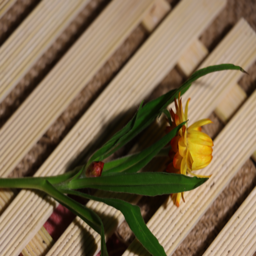}}}
        \end{minipage}
    \end{minipage}

    \begin{minipage}[b]{1.0\linewidth}
        \begin{minipage}[b]{0.137\linewidth}
            \centering
            \centerline{
            \stackinset{l}{0pt}{t}{0pt}{%
                \colorbox{gray}{%
                    \textcolor{white}{\textbf{\footnotesize (a)}}%
                }%
            }{\includegraphics[width=\linewidth]{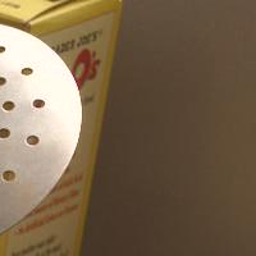}}}
        \end{minipage}   
        \hfill
        \begin{minipage}[b]{0.137\linewidth}
            \centering
            \centerline{
            \stackinset{l}{0pt}{t}{0pt}{%
                \colorbox{gray}{%
                    \textcolor{white}{\textbf{\footnotesize (b)}}%
                }%
            }{\includegraphics[width=\linewidth]{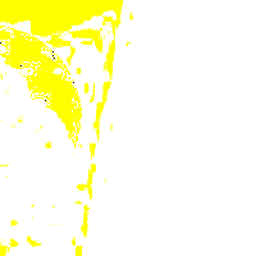}}}
        \end{minipage}   
        \hfill
        \begin{minipage}[b]{0.137\linewidth}
            \centering
            \centerline{
            \stackinset{l}{0pt}{t}{0pt}{%
                \colorbox{gray}{%
                    \textcolor{white}{\textbf{\footnotesize (c)}}%
                }%
            }{\includegraphics[width=\linewidth]{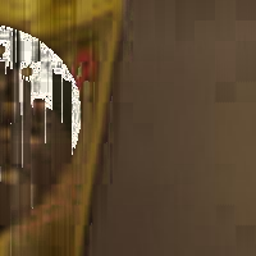}}}
        \end{minipage}   
        \hfill
        \begin{minipage}[b]{0.137\linewidth}
            \centering
            \centerline{
            \stackinset{l}{0pt}{t}{0pt}{%
                \colorbox{gray}{%
                    \textcolor{white}{\textbf{\footnotesize (d)}}%
                }%
            }{\includegraphics[width=\linewidth]{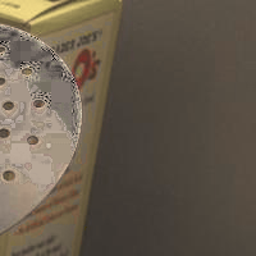}}}
        \end{minipage}   
        \hfill
        \begin{minipage}[b]{0.137\linewidth}
            \centering
            \centerline{
            \stackinset{l}{0pt}{t}{0pt}{%
                \colorbox{gray}{%
                    \textcolor{white}{\textbf{\footnotesize (e)}}%
                }%
            }{\includegraphics[width=\linewidth]{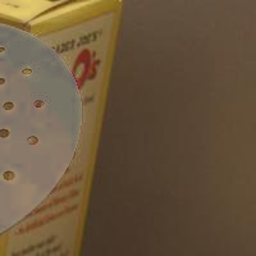}}}
        \end{minipage}
        \hfill
        \begin{minipage}[b]{0.137\linewidth}
            \centering
            \centerline{
            \stackinset{l}{0pt}{t}{0pt}{%
                \colorbox{gray}{%
                    \textcolor{white}{\textbf{\footnotesize (f)}}%
                }%
            }{\includegraphics[width=\linewidth]{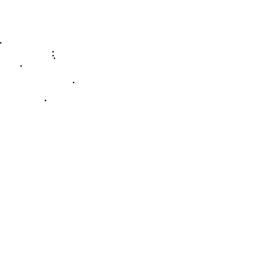}}}
        \end{minipage}
        \hfill
        \begin{minipage}[b]{0.137\linewidth}
            \centering
            \centerline{
            \stackinset{l}{0pt}{t}{0pt}{%
                \colorbox{gray}{%
                    \textcolor{white}{\textbf{\footnotesize (g)}}%
                }%
            }{\includegraphics[width=\linewidth]{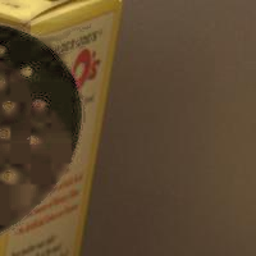}}}
        \end{minipage}   
    \end{minipage}

    \begin{minipage}[b]{1.0\linewidth}
        \begin{minipage}[b]{0.137\linewidth}
            \centering
            \centerline{
            \stackinset{l}{0pt}{t}{0pt}{%
                \colorbox{gray}{%
                    \textcolor{white}{\textbf{\footnotesize (h)}}%
                }%
            }{\includegraphics[width=\linewidth]{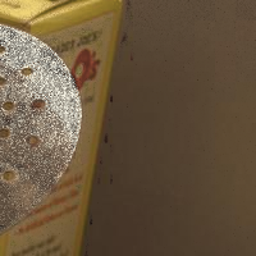}}}
        \end{minipage}   
        \hfill
        \begin{minipage}[b]{0.137\linewidth}
            \centering
            \centerline{
            \stackinset{l}{0pt}{t}{0pt}{%
                \colorbox{gray}{%
                    \textcolor{white}{\textbf{\footnotesize (i)}}%
                }%
            }{\includegraphics[width=\linewidth]{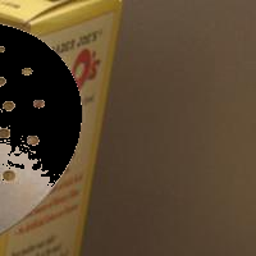}}}
        \end{minipage}   
        \hfill
        \begin{minipage}[b]{0.137\linewidth}
            \centering
            \centerline{
            \stackinset{l}{0pt}{t}{0pt}{%
                \colorbox{gray}{%
                    \textcolor{white}{\textbf{\footnotesize (j)}}%
                }%
            }{\includegraphics[width=\linewidth]{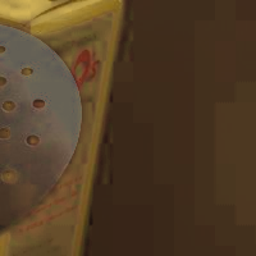}}}
        \end{minipage}   
        \hfill
        \begin{minipage}[b]{0.137\linewidth}
            \centering
            \centerline{
            \stackinset{l}{0pt}{t}{0pt}{%
                \colorbox{gray}{%
                    \textcolor{white}{\textbf{\footnotesize (k)}}%
                }%
            }{\includegraphics[width=\linewidth]{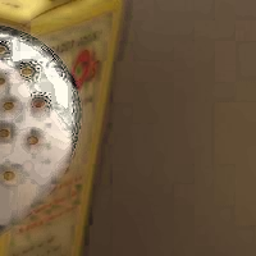}}}
        \end{minipage}
        \hfill
        \begin{minipage}[b]{0.137\linewidth}
            \centering
            \centerline{
            \stackinset{l}{0pt}{t}{0pt}{%
                \colorbox{gray}{%
                    \textcolor{white}{\textbf{\footnotesize (l)}}%
                }%
            }{\includegraphics[width=\linewidth]{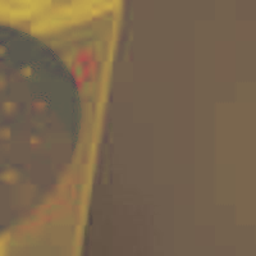}}}
        \end{minipage}
        \hfill
        \begin{minipage}[b]{0.137\linewidth}
            \centering
            \centerline{
            \stackinset{l}{0pt}{t}{0pt}{%
                \colorbox{gray}{%
                    \textcolor{white}{\textbf{\footnotesize (m)}}%
                }%
            }{\includegraphics[width=\linewidth]{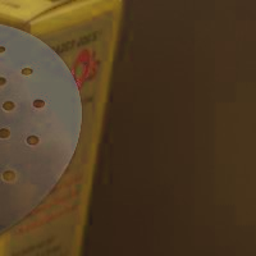}}}
        \end{minipage}   
        \hfill
        \begin{minipage}[b]{0.137\linewidth}
            \centering
            \centerline{
            \stackinset{l}{0pt}{t}{0pt}{%
                \colorbox{gray}{%
                    \textcolor{white}{\textbf{\footnotesize (n)}}%
                }%
            }{\includegraphics[width=\linewidth]{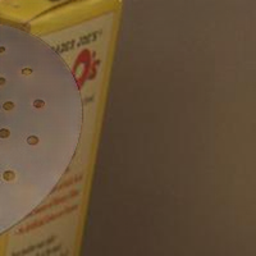}}}
        \end{minipage} 
    \end{minipage}

    \begin{minipage}[b]{1.0\linewidth}
        \begin{minipage}[b]{0.137\linewidth}
            \centering
            \centerline{
            \stackinset{l}{0pt}{t}{0pt}{%
                \colorbox{gray}{%
                    \textcolor{white}{\textbf{\footnotesize (o)}}%
                }%
            }{\includegraphics[width=\linewidth]{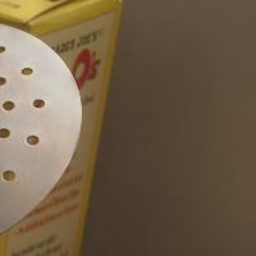}}}
        \end{minipage}   
        \hfill
        \begin{minipage}[b]{0.137\linewidth}
            \centering
            \centerline{
            \stackinset{l}{0pt}{t}{0pt}{%
                \colorbox{gray}{%
                    \textcolor{white}{\textbf{\footnotesize (p)}}%
                }%
            }{\includegraphics[width=\linewidth]{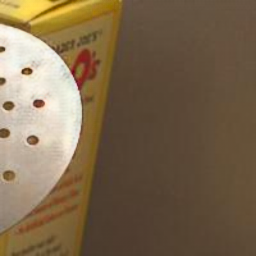}}}
        \end{minipage}   
        \hfill
        \begin{minipage}[b]{0.137\linewidth}
            \centering
            \centerline{
            \stackinset{l}{0pt}{t}{0pt}{%
                \colorbox{gray}{%
                    \textcolor{white}{\textbf{\footnotesize (q)}}%
                }%
            }{\includegraphics[width=\linewidth]{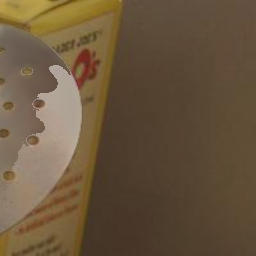}}}
        \end{minipage}
        \hfill
        \begin{minipage}[b]{0.137\linewidth}
            \centering
            \centerline{
            \stackinset{l}{0pt}{t}{0pt}{%
                \colorbox{gray}{%
                    \textcolor{white}{\textbf{\footnotesize (r)}}%
                }%
            }{\includegraphics[width=\linewidth]{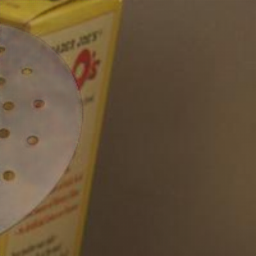}}}
        \end{minipage}
        \hfill
        \begin{minipage}[b]{0.137\linewidth}
            \centering
            \centerline{
            \stackinset{l}{0pt}{t}{0pt}{%
                \colorbox{gray}{%
                    \textcolor{white}{\textbf{\footnotesize (s)}}%
                }%
            }{\includegraphics[width=\linewidth]{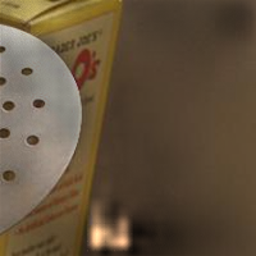}}}
        \end{minipage}
        \hfill
        \begin{minipage}[b]{0.137\linewidth}
            \centering
            \centerline{
            \stackinset{l}{0pt}{t}{0pt}{%
                \colorbox{gray}{%
                    \textcolor{white}{\textbf{\footnotesize (t)}}%
                }%
            }{\includegraphics[width=\linewidth]{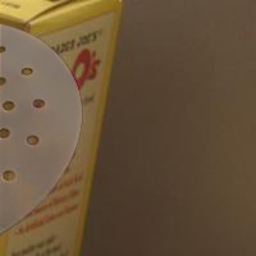}}}
        \end{minipage}
        \hfill
        \begin{minipage}[b]{0.137\linewidth}
            \centering
            \centerline{
            \stackinset{l}{0pt}{t}{0pt}{%
                \colorbox{gray}{%
                    \textcolor{white}{\textbf{\footnotesize (u)}}%
                }%
            }{\includegraphics[width=\linewidth]{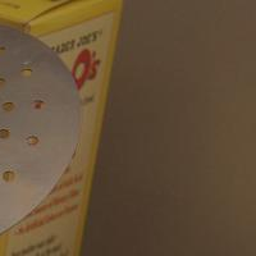}}}
        \end{minipage}
    \end{minipage}

    \begin{minipage}[b]{1.0\linewidth}
        \begin{minipage}[b]{0.137\linewidth}
            \centering
            \centerline{
            \stackinset{l}{0pt}{t}{0pt}{%
                \colorbox{gray}{%
                    \textcolor{white}{\textbf{\footnotesize (a)}}%
                }%
            }{\includegraphics[width=\linewidth]{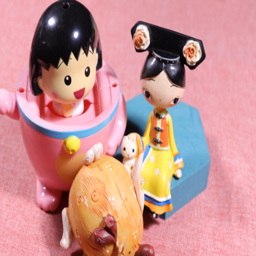}}}
        \end{minipage}   
        \hfill
        \begin{minipage}[b]{0.137\linewidth}
            \centering
            \centerline{
            \stackinset{l}{0pt}{t}{0pt}{%
                \colorbox{gray}{%
                    \textcolor{white}{\textbf{\footnotesize (b)}}%
                }%
            }{\includegraphics[width=\linewidth]{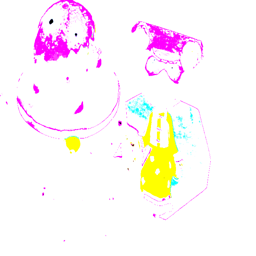}}}
        \end{minipage}   
        \hfill
        \begin{minipage}[b]{0.137\linewidth}
            \centering
            \centerline{
            \stackinset{l}{0pt}{t}{0pt}{%
                \colorbox{gray}{%
                    \textcolor{white}{\textbf{\footnotesize (c)}}%
                }%
            }{\includegraphics[width=\linewidth]{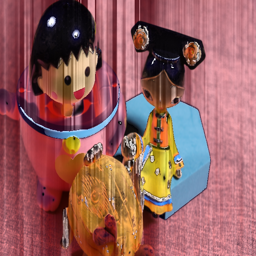}}}
        \end{minipage}   
        \hfill
        \begin{minipage}[b]{0.137\linewidth}
            \centering
            \centerline{
            \stackinset{l}{0pt}{t}{0pt}{%
                \colorbox{gray}{%
                    \textcolor{white}{\textbf{\footnotesize (d)}}%
                }%
            }{\includegraphics[width=\linewidth]{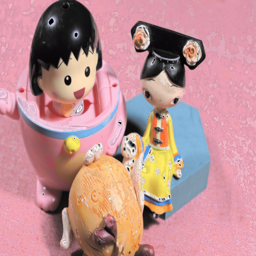}}}
        \end{minipage}   
        \hfill
        \begin{minipage}[b]{0.137\linewidth}
            \centering
            \centerline{
            \stackinset{l}{0pt}{t}{0pt}{%
                \colorbox{gray}{%
                    \textcolor{white}{\textbf{\footnotesize (e)}}%
                }%
            }{\includegraphics[width=\linewidth]{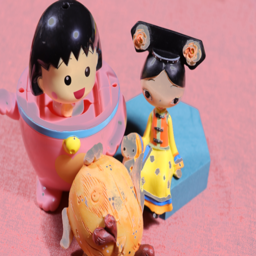}}}
        \end{minipage}
        \hfill
        \begin{minipage}[b]{0.137\linewidth}
            \centering
            \centerline{
            \stackinset{l}{0pt}{t}{0pt}{%
                \colorbox{gray}{%
                    \textcolor{white}{\textbf{\footnotesize (f)}}%
                }%
            }{\includegraphics[width=\linewidth]{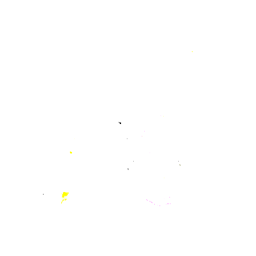}}}
        \end{minipage}
        \hfill
        \begin{minipage}[b]{0.137\linewidth}
            \centering
            \centerline{
            \stackinset{l}{0pt}{t}{0pt}{%
                \colorbox{gray}{%
                    \textcolor{white}{\textbf{\footnotesize (g)}}%
                }%
            }{\includegraphics[width=\linewidth]{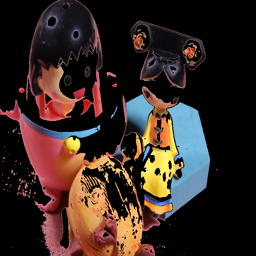}}}
        \end{minipage}   
    \end{minipage}

    \begin{minipage}[b]{1.0\linewidth}
        \begin{minipage}[b]{0.137\linewidth}
            \centering
            \centerline{
            \stackinset{l}{0pt}{t}{0pt}{%
                \colorbox{gray}{%
                    \textcolor{white}{\textbf{\footnotesize (h)}}%
                }%
            }{\includegraphics[width=\linewidth]{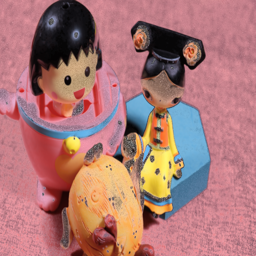}}}
        \end{minipage}   
        \hfill
        \begin{minipage}[b]{0.137\linewidth}
            \centering
            \centerline{
            \stackinset{l}{0pt}{t}{0pt}{%
                \colorbox{gray}{%
                    \textcolor{white}{\textbf{\footnotesize (i)}}%
                }%
            }{\includegraphics[width=\linewidth]{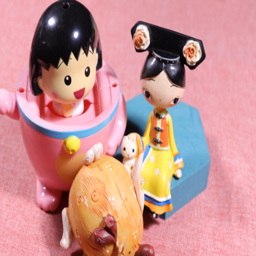}}}
        \end{minipage}   
        \hfill
        \begin{minipage}[b]{0.137\linewidth}
            \centering
            \centerline{
            \stackinset{l}{0pt}{t}{0pt}{%
                \colorbox{gray}{%
                    \textcolor{white}{\textbf{\footnotesize (j)}}%
                }%
            }{\includegraphics[width=\linewidth]{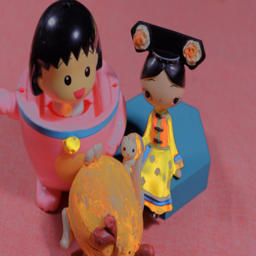}}}
        \end{minipage}   
        \hfill
        \begin{minipage}[b]{0.137\linewidth}
            \centering
            \centerline{
            \stackinset{l}{0pt}{t}{0pt}{%
                \colorbox{gray}{%
                    \textcolor{white}{\textbf{\footnotesize (k)}}%
                }%
            }{\includegraphics[width=\linewidth]{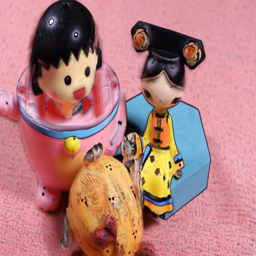}}}
        \end{minipage}
        \hfill
        \begin{minipage}[b]{0.137\linewidth}
            \centering
            \centerline{
            \stackinset{l}{0pt}{t}{0pt}{%
                \colorbox{gray}{%
                    \textcolor{white}{\textbf{\footnotesize (l)}}%
                }%
            }{\includegraphics[width=\linewidth]{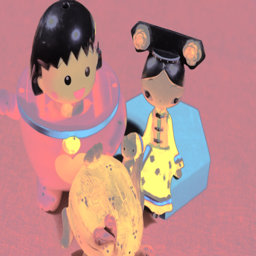}}}
        \end{minipage}
        \hfill
        \begin{minipage}[b]{0.137\linewidth}
            \centering
            \centerline{
            \stackinset{l}{0pt}{t}{0pt}{%
                \colorbox{gray}{%
                    \textcolor{white}{\textbf{\footnotesize (m)}}%
                }%
            }{\includegraphics[width=\linewidth]{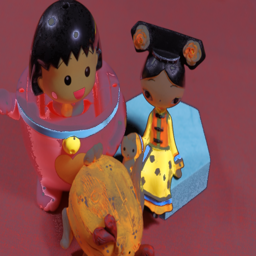}}}
        \end{minipage}   
        \hfill
        \begin{minipage}[b]{0.137\linewidth}
            \centering
            \centerline{
            \stackinset{l}{0pt}{t}{0pt}{%
                \colorbox{gray}{%
                    \textcolor{white}{\textbf{\footnotesize (n)}}%
                }%
            }{\includegraphics[width=\linewidth]{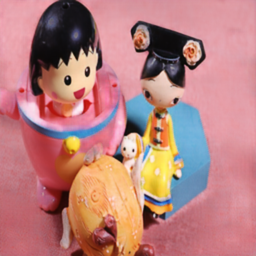}}}
        \end{minipage} 
    \end{minipage}

    \begin{minipage}[b]{1.0\linewidth}
        \begin{minipage}[b]{0.137\linewidth}
            \centering
            \centerline{
            \stackinset{l}{0pt}{t}{0pt}{%
                \colorbox{gray}{%
                    \textcolor{white}{\textbf{\footnotesize (o)}}%
                }%
            }{\includegraphics[width=\linewidth]{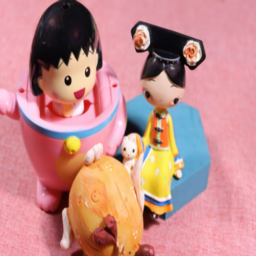}}}
        \end{minipage}   
        \hfill
        \begin{minipage}[b]{0.137\linewidth}
            \centering
            \centerline{
            \stackinset{l}{0pt}{t}{0pt}{%
                \colorbox{gray}{%
                    \textcolor{white}{\textbf{\footnotesize (p)}}%
                }%
            }{\includegraphics[width=\linewidth]{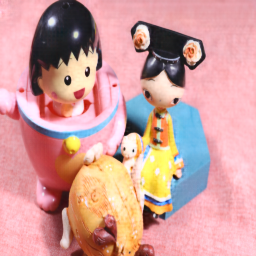}}}
        \end{minipage}   
        \hfill
        \begin{minipage}[b]{0.137\linewidth}
            \centering
            \centerline{
            \stackinset{l}{0pt}{t}{0pt}{%
                \colorbox{gray}{%
                    \textcolor{white}{\textbf{\footnotesize (q)}}%
                }%
            }{\includegraphics[width=\linewidth]{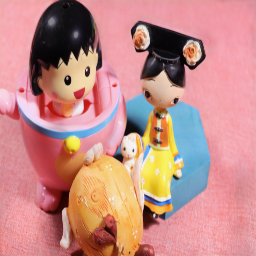}}}
        \end{minipage}
        \hfill
        \begin{minipage}[b]{0.137\linewidth}
            \centering
            \centerline{
            \stackinset{l}{0pt}{t}{0pt}{%
                \colorbox{gray}{%
                    \textcolor{white}{\textbf{\footnotesize (r)}}%
                }%
            }{\includegraphics[width=\linewidth]{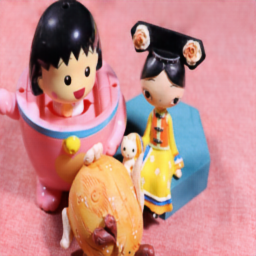}}}
        \end{minipage}
        \hfill
        \begin{minipage}[b]{0.137\linewidth}
            \centering
            \centerline{
            \stackinset{l}{0pt}{t}{0pt}{%
                \colorbox{gray}{%
                    \textcolor{white}{\textbf{\footnotesize (s)}}%
                }%
            }{\includegraphics[width=\linewidth]{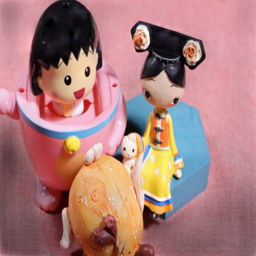}}}
        \end{minipage}
        \hfill
        \begin{minipage}[b]{0.137\linewidth}
            \centering
            \centerline{
            \stackinset{l}{0pt}{t}{0pt}{%
                \colorbox{gray}{%
                    \textcolor{white}{\textbf{\footnotesize (t)}}%
                }%
            }{\includegraphics[width=\linewidth]{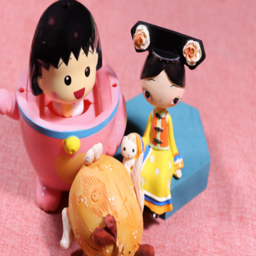}}}
        \end{minipage}
        \hfill
        \begin{minipage}[b]{0.137\linewidth}
            \centering
            \centerline{
            \stackinset{l}{0pt}{t}{0pt}{%
                \colorbox{gray}{%
                    \textcolor{white}{\textbf{\footnotesize (u)}}%
                }%
            }{\includegraphics[width=\linewidth]{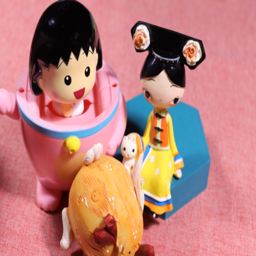}}}
        \end{minipage}
    \end{minipage}

    \caption{{\small Comprehensive visual comparison. (a) Input specular highlight image, (b) Tan [10], (c) Yoon [31], (d) Shen [11], (e) Shen [12], (f) Yang [13], (g) Shen [14], (h) Akashi [15], (i) Huo [32], (j) Fu [18], (k) Yamamoto [19], (l) Saha [20], (m) SLRR [22], (n) JSHDR [6], (o) SpecularityNet [5], (p) MG-CycleGAN [26], (q) Wu [25], (r) TSHRNet [7], (s) AHA [28], (t) Ours, (u) GT diffuse image. The reader is encouraged to zoom-in.}}
    \label{fig:full_compare3}
\end{figure*}

\begin{figure*}[hb]
    \begin{minipage}[b]{1.0\linewidth}
        \begin{minipage}[b]{0.137\linewidth}
            \centering
            \centerline{
            \stackinset{l}{0pt}{t}{0pt}{%
                \colorbox{gray}{%
                    \textcolor{white}{\textbf{\footnotesize (a)}}%
                }%
            }{\includegraphics[width=\linewidth]{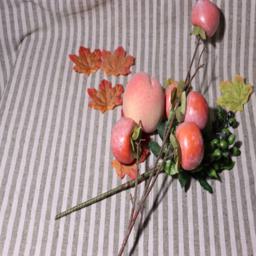}}}
        \end{minipage}   
        \hfill
        \begin{minipage}[b]{0.137\linewidth}
            \centering
            \centerline{
            \stackinset{l}{0pt}{t}{0pt}{%
                \colorbox{gray}{%
                    \textcolor{white}{\textbf{\footnotesize (b)}}%
                }%
            }{\includegraphics[width=\linewidth]{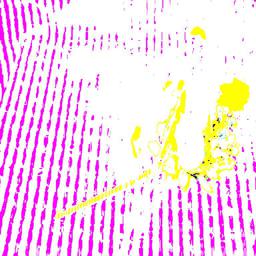}}}
        \end{minipage}   
        \hfill
        \begin{minipage}[b]{0.137\linewidth}
            \centering
            \centerline{
            \stackinset{l}{0pt}{t}{0pt}{%
                \colorbox{gray}{%
                    \textcolor{white}{\textbf{\footnotesize (c)}}%
                }%
            }{\includegraphics[width=\linewidth]{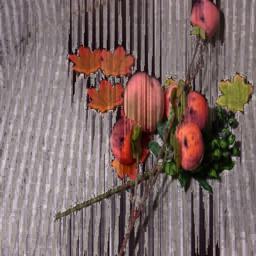}}}
        \end{minipage}   
        \hfill
        \begin{minipage}[b]{0.137\linewidth}
            \centering
            \centerline{
            \stackinset{l}{0pt}{t}{0pt}{%
                \colorbox{gray}{%
                    \textcolor{white}{\textbf{\footnotesize (d)}}%
                }%
            }{\includegraphics[width=\linewidth]{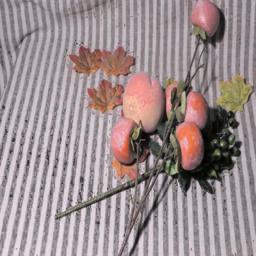}}}
        \end{minipage}   
        \hfill
        \begin{minipage}[b]{0.137\linewidth}
            \centering
            \centerline{
            \stackinset{l}{0pt}{t}{0pt}{%
                \colorbox{gray}{%
                    \textcolor{white}{\textbf{\footnotesize (e)}}%
                }%
            }{\includegraphics[width=\linewidth]{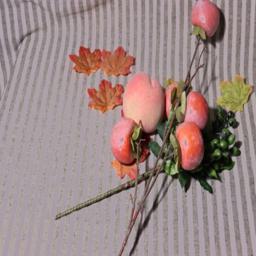}}}
        \end{minipage}
        \hfill
        \begin{minipage}[b]{0.137\linewidth}
            \centering
            \centerline{
            \stackinset{l}{0pt}{t}{0pt}{%
                \colorbox{gray}{%
                    \textcolor{white}{\textbf{\footnotesize (f)}}%
                }%
            }{\includegraphics[width=\linewidth]{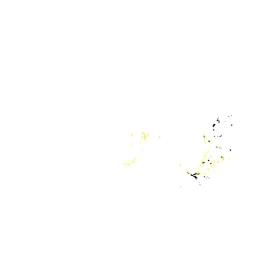}}}
        \end{minipage}
        \hfill
        \begin{minipage}[b]{0.137\linewidth}
            \centering
            \centerline{
            \stackinset{l}{0pt}{t}{0pt}{%
                \colorbox{gray}{%
                    \textcolor{white}{\textbf{\footnotesize (g)}}%
                }%
            }{\includegraphics[width=\linewidth]{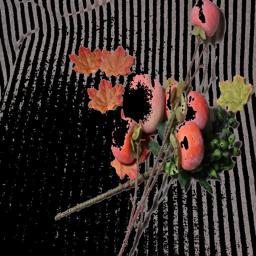}}}
        \end{minipage}   
    \end{minipage}

    \begin{minipage}[b]{1.0\linewidth}
        \begin{minipage}[b]{0.137\linewidth}
            \centering
            \centerline{
            \stackinset{l}{0pt}{t}{0pt}{%
                \colorbox{gray}{%
                    \textcolor{white}{\textbf{\footnotesize (h)}}%
                }%
            }{\includegraphics[width=\linewidth]{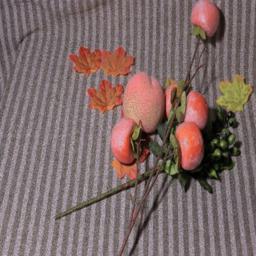}}}
        \end{minipage}   
        \hfill
        \begin{minipage}[b]{0.137\linewidth}
            \centering
            \centerline{
            \stackinset{l}{0pt}{t}{0pt}{%
                \colorbox{gray}{%
                    \textcolor{white}{\textbf{\footnotesize (i)}}%
                }%
            }{\includegraphics[width=\linewidth]{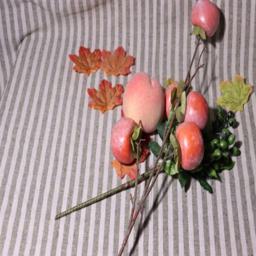}}}
        \end{minipage}   
        \hfill
        \begin{minipage}[b]{0.137\linewidth}
            \centering
            \centerline{
            \stackinset{l}{0pt}{t}{0pt}{%
                \colorbox{gray}{%
                    \textcolor{white}{\textbf{\footnotesize (j)}}%
                }%
            }{\includegraphics[width=\linewidth]{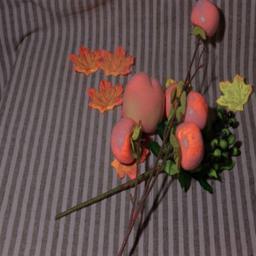}}}
        \end{minipage}   
        \hfill
        \begin{minipage}[b]{0.137\linewidth}
            \centering
            \centerline{
            \stackinset{l}{0pt}{t}{0pt}{%
                \colorbox{gray}{%
                    \textcolor{white}{\textbf{\footnotesize (k)}}%
                }%
            }{\includegraphics[width=\linewidth]{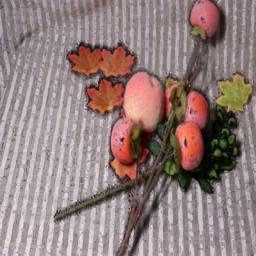}}}
        \end{minipage}
        \hfill
        \begin{minipage}[b]{0.137\linewidth}
            \centering
            \centerline{
            \stackinset{l}{0pt}{t}{0pt}{%
                \colorbox{gray}{%
                    \textcolor{white}{\textbf{\footnotesize (l)}}%
                }%
            }{\includegraphics[width=\linewidth]{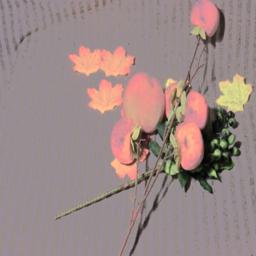}}}
        \end{minipage}
        \hfill
        \begin{minipage}[b]{0.137\linewidth}
            \centering
            \centerline{
            \stackinset{l}{0pt}{t}{0pt}{%
                \colorbox{gray}{%
                    \textcolor{white}{\textbf{\footnotesize (m)}}%
                }%
            }{\includegraphics[width=\linewidth]{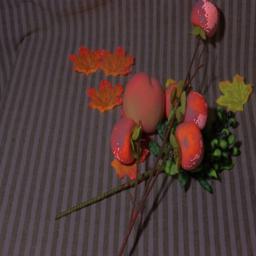}}}
        \end{minipage}   
        \hfill
        \begin{minipage}[b]{0.137\linewidth}
            \centering
            \centerline{
            \stackinset{l}{0pt}{t}{0pt}{%
                \colorbox{gray}{%
                    \textcolor{white}{\textbf{\footnotesize (n)}}%
                }%
            }{\includegraphics[width=\linewidth]{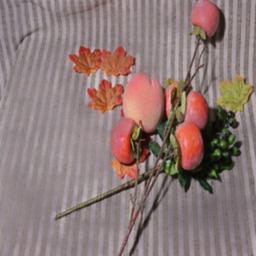}}}
        \end{minipage} 
    \end{minipage}

    \begin{minipage}[b]{1.0\linewidth}
        \begin{minipage}[b]{0.137\linewidth}
            \centering
            \centerline{
            \stackinset{l}{0pt}{t}{0pt}{%
                \colorbox{gray}{%
                    \textcolor{white}{\textbf{\footnotesize (o)}}%
                }%
            }{\includegraphics[width=\linewidth]{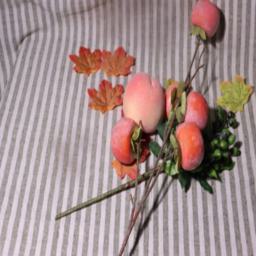}}}
        \end{minipage}   
        \hfill
        \begin{minipage}[b]{0.137\linewidth}
            \centering
            \centerline{
            \stackinset{l}{0pt}{t}{0pt}{%
                \colorbox{gray}{%
                    \textcolor{white}{\textbf{\footnotesize (p)}}%
                }%
            }{\includegraphics[width=\linewidth]{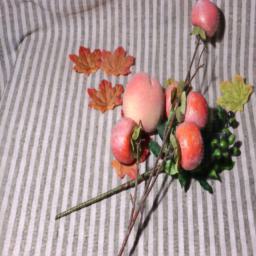}}}
        \end{minipage}   
        \hfill
        \begin{minipage}[b]{0.137\linewidth}
            \centering
            \centerline{
            \stackinset{l}{0pt}{t}{0pt}{%
                \colorbox{gray}{%
                    \textcolor{white}{\textbf{\footnotesize (q)}}%
                }%
            }{\includegraphics[width=\linewidth]{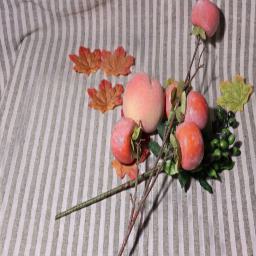}}}
        \end{minipage}
        \hfill
        \begin{minipage}[b]{0.137\linewidth}
            \centering
            \centerline{
            \stackinset{l}{0pt}{t}{0pt}{%
                \colorbox{gray}{%
                    \textcolor{white}{\textbf{\footnotesize (r)}}%
                }%
            }{\includegraphics[width=\linewidth]{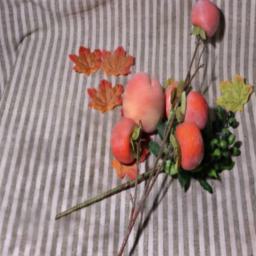}}}
        \end{minipage}
        \hfill
        \begin{minipage}[b]{0.137\linewidth}
            \centering
            \centerline{
            \stackinset{l}{0pt}{t}{0pt}{%
                \colorbox{gray}{%
                    \textcolor{white}{\textbf{\footnotesize (s)}}%
                }%
            }{\includegraphics[width=\linewidth]{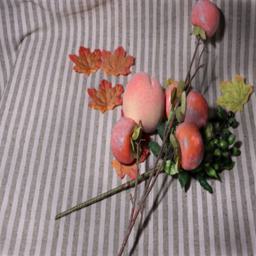}}}
        \end{minipage}
        \hfill
        \begin{minipage}[b]{0.137\linewidth}
            \centering
            \centerline{
            \stackinset{l}{0pt}{t}{0pt}{%
                \colorbox{gray}{%
                    \textcolor{white}{\textbf{\footnotesize (t)}}%
                }%
            }{\includegraphics[width=\linewidth]{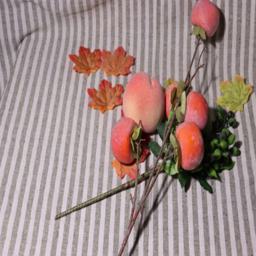}}}
        \end{minipage}
        \hfill
        \begin{minipage}[b]{0.137\linewidth}
            \centering
            \centerline{
            \stackinset{l}{0pt}{t}{0pt}{%
                \colorbox{gray}{%
                    \textcolor{white}{\textbf{\footnotesize (u)}}%
                }%
            }{\includegraphics[width=\linewidth]{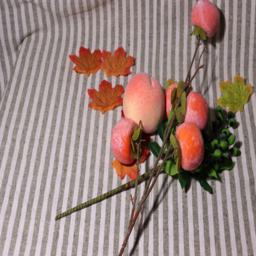}}}
        \end{minipage}
    \end{minipage}

    \begin{minipage}[b]{1.0\linewidth}
        \begin{minipage}[b]{0.137\linewidth}
            \centering
            \centerline{
            \stackinset{l}{0pt}{t}{0pt}{%
                \colorbox{gray}{%
                    \textcolor{white}{\textbf{\footnotesize (a)}}%
                }%
            }{\includegraphics[width=\linewidth]{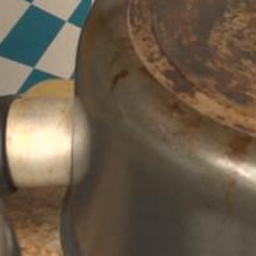}}}
        \end{minipage}   
        \hfill
        \begin{minipage}[b]{0.137\linewidth}
            \centering
            \centerline{
            \stackinset{l}{0pt}{t}{0pt}{%
                \colorbox{gray}{%
                    \textcolor{white}{\textbf{\footnotesize (b)}}%
                }%
            }{\includegraphics[width=\linewidth]{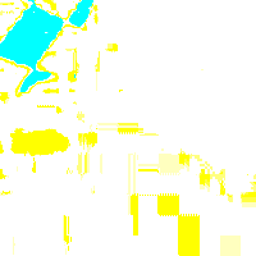}}}
        \end{minipage}   
        \hfill
        \begin{minipage}[b]{0.137\linewidth}
            \centering
            \centerline{
            \stackinset{l}{0pt}{t}{0pt}{%
                \colorbox{gray}{%
                    \textcolor{white}{\textbf{\footnotesize (c)}}%
                }%
            }{\includegraphics[width=\linewidth]{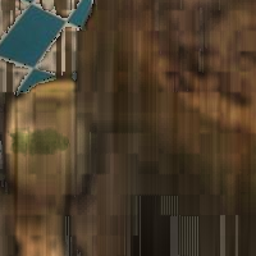}}}
        \end{minipage}   
        \hfill
        \begin{minipage}[b]{0.137\linewidth}
            \centering
            \centerline{
            \stackinset{l}{0pt}{t}{0pt}{%
                \colorbox{gray}{%
                    \textcolor{white}{\textbf{\footnotesize (d)}}%
                }%
            }{\includegraphics[width=\linewidth]{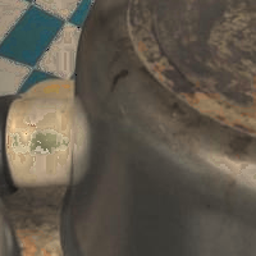}}}
        \end{minipage}   
        \hfill
        \begin{minipage}[b]{0.137\linewidth}
            \centering
            \centerline{
            \stackinset{l}{0pt}{t}{0pt}{%
                \colorbox{gray}{%
                    \textcolor{white}{\textbf{\footnotesize (e)}}%
                }%
            }{\includegraphics[width=\linewidth]{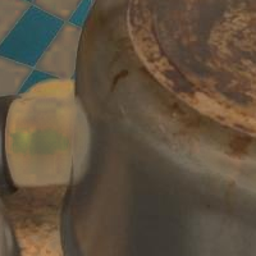}}}
        \end{minipage}
        \hfill
        \begin{minipage}[b]{0.137\linewidth}
            \centering
            \centerline{
            \stackinset{l}{0pt}{t}{0pt}{%
                \colorbox{gray}{%
                    \textcolor{white}{\textbf{\footnotesize (f)}}%
                }%
            }{\includegraphics[width=\linewidth]{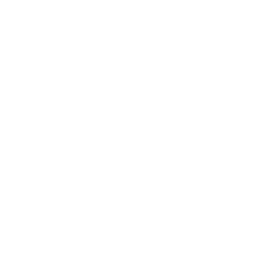}}}
        \end{minipage}
        \hfill
        \begin{minipage}[b]{0.137\linewidth}
            \centering
            \centerline{
            \stackinset{l}{0pt}{t}{0pt}{%
                \colorbox{gray}{%
                    \textcolor{white}{\textbf{\footnotesize (g)}}%
                }%
            }{\includegraphics[width=\linewidth]{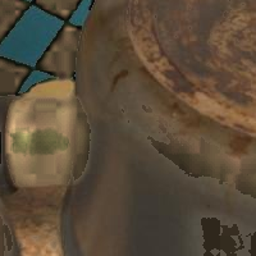}}}
        \end{minipage}   
    \end{minipage}

    \begin{minipage}[b]{1.0\linewidth}
        \begin{minipage}[b]{0.137\linewidth}
            \centering
            \centerline{
            \stackinset{l}{0pt}{t}{0pt}{%
                \colorbox{gray}{%
                    \textcolor{white}{\textbf{\footnotesize (h)}}%
                }%
            }{\includegraphics[width=\linewidth]{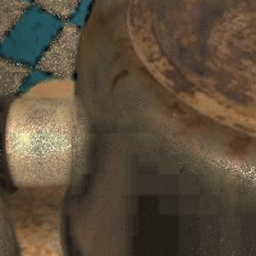}}}
        \end{minipage}   
        \hfill
        \begin{minipage}[b]{0.137\linewidth}
            \centering
            \centerline{
            \stackinset{l}{0pt}{t}{0pt}{%
                \colorbox{gray}{%
                    \textcolor{white}{\textbf{\footnotesize (i)}}%
                }%
            }{\includegraphics[width=\linewidth]{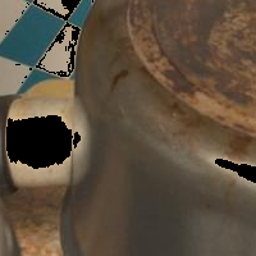}}}
        \end{minipage}   
        \hfill
        \begin{minipage}[b]{0.137\linewidth}
            \centering
            \centerline{
            \stackinset{l}{0pt}{t}{0pt}{%
                \colorbox{gray}{%
                    \textcolor{white}{\textbf{\footnotesize (j)}}%
                }%
            }{\includegraphics[width=\linewidth]{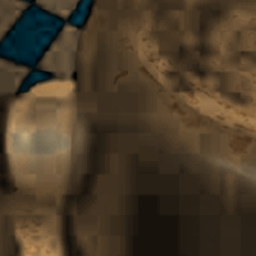}}}
        \end{minipage}   
        \hfill
        \begin{minipage}[b]{0.137\linewidth}
            \centering
            \centerline{
            \stackinset{l}{0pt}{t}{0pt}{%
                \colorbox{gray}{%
                    \textcolor{white}{\textbf{\footnotesize (k)}}%
                }%
            }{\includegraphics[width=\linewidth]{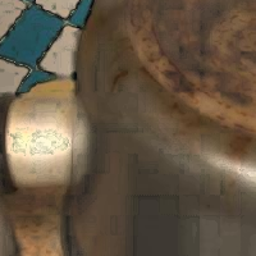}}}
        \end{minipage}
        \hfill
        \begin{minipage}[b]{0.137\linewidth}
            \centering
            \centerline{
            \stackinset{l}{0pt}{t}{0pt}{%
                \colorbox{gray}{%
                    \textcolor{white}{\textbf{\footnotesize (l)}}%
                }%
            }{\includegraphics[width=\linewidth]{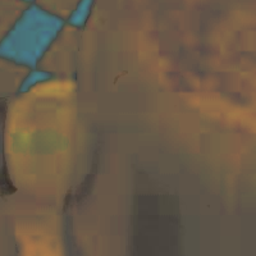}}}
        \end{minipage}
        \hfill
        \begin{minipage}[b]{0.137\linewidth}
            \centering
            \centerline{
            \stackinset{l}{0pt}{t}{0pt}{%
                \colorbox{gray}{%
                    \textcolor{white}{\textbf{\footnotesize (m)}}%
                }%
            }{\includegraphics[width=\linewidth]{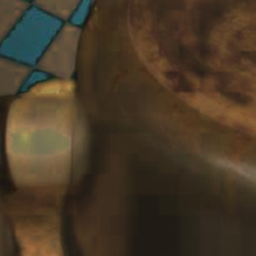}}}
        \end{minipage}   
        \hfill
        \begin{minipage}[b]{0.137\linewidth}
            \centering
            \centerline{
            \stackinset{l}{0pt}{t}{0pt}{%
                \colorbox{gray}{%
                    \textcolor{white}{\textbf{\footnotesize (n)}}%
                }%
            }{\includegraphics[width=\linewidth]{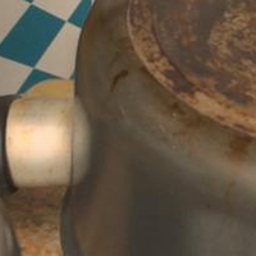}}}
        \end{minipage} 
    \end{minipage}

    \begin{minipage}[b]{1.0\linewidth}
        \begin{minipage}[b]{0.137\linewidth}
            \centering
            \centerline{
            \stackinset{l}{0pt}{t}{0pt}{%
                \colorbox{gray}{%
                    \textcolor{white}{\textbf{\footnotesize (o)}}%
                }%
            }{\includegraphics[width=\linewidth]{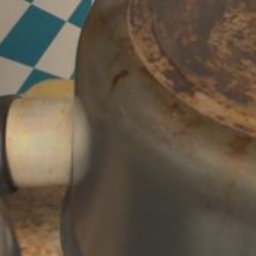}}}
        \end{minipage}   
        \hfill
        \begin{minipage}[b]{0.137\linewidth}
            \centering
            \centerline{
            \stackinset{l}{0pt}{t}{0pt}{%
                \colorbox{gray}{%
                    \textcolor{white}{\textbf{\footnotesize (p)}}%
                }%
            }{\includegraphics[width=\linewidth]{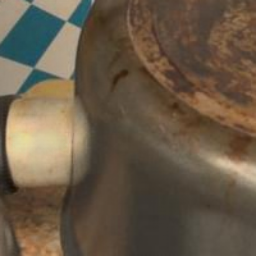}}}
        \end{minipage}   
        \hfill
        \begin{minipage}[b]{0.137\linewidth}
            \centering
            \centerline{
            \stackinset{l}{0pt}{t}{0pt}{%
                \colorbox{gray}{%
                    \textcolor{white}{\textbf{\footnotesize (q)}}%
                }%
            }{\includegraphics[width=\linewidth]{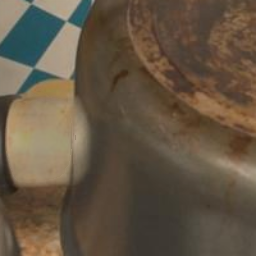}}}
        \end{minipage}
        \hfill
        \begin{minipage}[b]{0.137\linewidth}
            \centering
            \centerline{
            \stackinset{l}{0pt}{t}{0pt}{%
                \colorbox{gray}{%
                    \textcolor{white}{\textbf{\footnotesize (r)}}%
                }%
            }{\includegraphics[width=\linewidth]{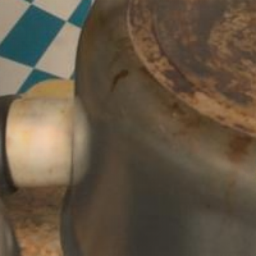}}}
        \end{minipage}
        \hfill
        \begin{minipage}[b]{0.137\linewidth}
            \centering
            \centerline{
            \stackinset{l}{0pt}{t}{0pt}{%
                \colorbox{gray}{%
                    \textcolor{white}{\textbf{\footnotesize (s)}}%
                }%
            }{\includegraphics[width=\linewidth]{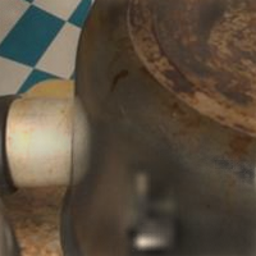}}}
        \end{minipage}
        \hfill
        \begin{minipage}[b]{0.137\linewidth}
            \centering
            \centerline{
            \stackinset{l}{0pt}{t}{0pt}{%
                \colorbox{gray}{%
                    \textcolor{white}{\textbf{\footnotesize (t)}}%
                }%
            }{\includegraphics[width=\linewidth]{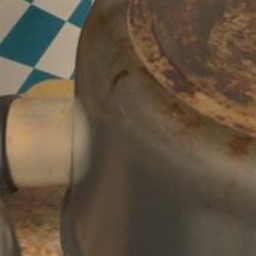}}}
        \end{minipage}
        \hfill
        \begin{minipage}[b]{0.137\linewidth}
            \centering
            \centerline{
            \stackinset{l}{0pt}{t}{0pt}{%
                \colorbox{gray}{%
                    \textcolor{white}{\textbf{\footnotesize (u)}}%
                }%
            }{\includegraphics[width=\linewidth]{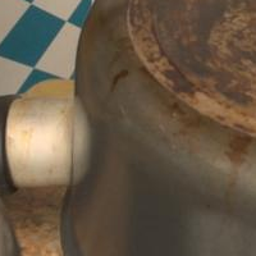}}}
        \end{minipage}
    \end{minipage}

    \begin{minipage}[b]{1.0\linewidth}
        \begin{minipage}[b]{0.137\linewidth}
            \centering
            \centerline{
            \stackinset{l}{0pt}{t}{0pt}{%
                \colorbox{gray}{%
                    \textcolor{white}{\textbf{\footnotesize (a)}}%
                }%
            }{\includegraphics[width=\linewidth]{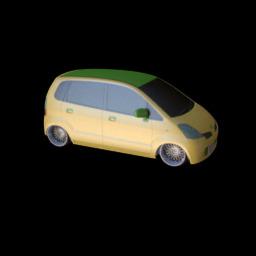}}}
        \end{minipage}   
        \hfill
        \begin{minipage}[b]{0.137\linewidth}
            \centering
            \centerline{
            \stackinset{l}{0pt}{t}{0pt}{%
                \colorbox{gray}{%
                    \textcolor{white}{\textbf{\footnotesize (b)}}%
                }%
            }{\includegraphics[width=\linewidth]{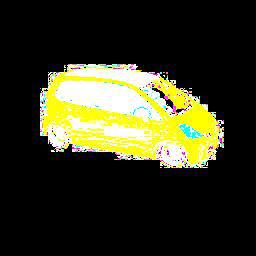}}}
        \end{minipage}   
        \hfill
        \begin{minipage}[b]{0.137\linewidth}
            \centering
            \centerline{
            \stackinset{l}{0pt}{t}{0pt}{%
                \colorbox{gray}{%
                    \textcolor{white}{\textbf{\footnotesize (c)}}%
                }%
            }{\includegraphics[width=\linewidth]{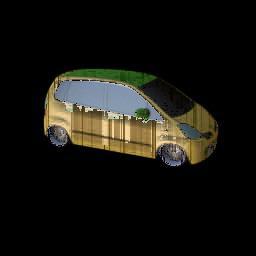}}}
        \end{minipage}   
        \hfill
        \begin{minipage}[b]{0.137\linewidth}
            \centering
            \centerline{
            \stackinset{l}{0pt}{t}{0pt}{%
                \colorbox{gray}{%
                    \textcolor{white}{\textbf{\footnotesize (d)}}%
                }%
            }{\includegraphics[width=\linewidth]{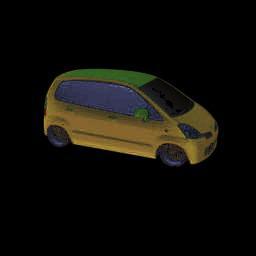}}}
        \end{minipage}   
        \hfill
        \begin{minipage}[b]{0.137\linewidth}
            \centering
            \centerline{
            \stackinset{l}{0pt}{t}{0pt}{%
                \colorbox{gray}{%
                    \textcolor{white}{\textbf{\footnotesize (e)}}%
                }%
            }{\includegraphics[width=\linewidth]{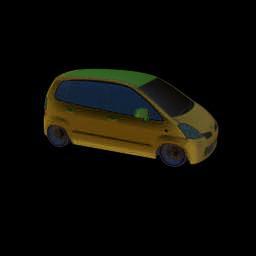}}}
        \end{minipage}
        \hfill
        \begin{minipage}[b]{0.137\linewidth}
            \centering
            \centerline{
            \stackinset{l}{0pt}{t}{0pt}{%
                \colorbox{gray}{%
                    \textcolor{white}{\textbf{\footnotesize (f)}}%
                }%
            }{\includegraphics[width=\linewidth]{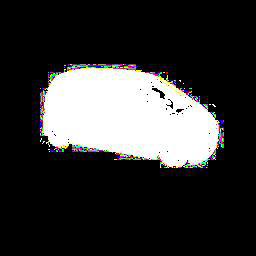}}}
        \end{minipage}
        \hfill
        \begin{minipage}[b]{0.137\linewidth}
            \centering
            \centerline{
            \stackinset{l}{0pt}{t}{0pt}{%
                \colorbox{gray}{%
                    \textcolor{white}{\textbf{\footnotesize (g)}}%
                }%
            }{\includegraphics[width=\linewidth]{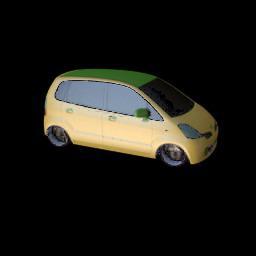}}}
        \end{minipage}   
    \end{minipage}

    \begin{minipage}[b]{1.0\linewidth}
        \begin{minipage}[b]{0.137\linewidth}
            \centering
            \centerline{
            \stackinset{l}{0pt}{t}{0pt}{%
                \colorbox{gray}{%
                    \textcolor{white}{\textbf{\footnotesize (h)}}%
                }%
            }{\includegraphics[width=\linewidth]{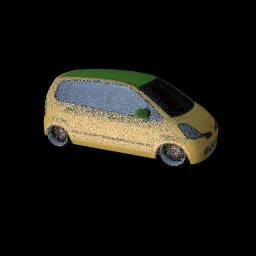}}}
        \end{minipage}   
        \hfill
        \begin{minipage}[b]{0.137\linewidth}
            \centering
            \centerline{
            \stackinset{l}{0pt}{t}{0pt}{%
                \colorbox{gray}{%
                    \textcolor{white}{\textbf{\footnotesize (i)}}%
                }%
            }{\includegraphics[width=\linewidth]{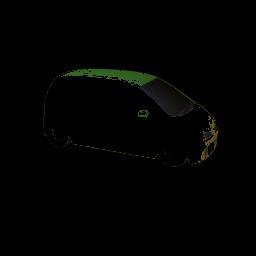}}}
        \end{minipage}   
        \hfill
        \begin{minipage}[b]{0.137\linewidth}
            \centering
            \centerline{
            \stackinset{l}{0pt}{t}{0pt}{%
                \colorbox{gray}{%
                    \textcolor{white}{\textbf{\footnotesize (j)}}%
                }%
            }{\includegraphics[width=\linewidth]{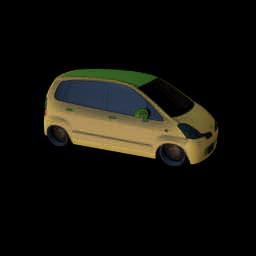}}}
        \end{minipage}   
        \hfill
        \begin{minipage}[b]{0.137\linewidth}
            \centering
            \centerline{
            \stackinset{l}{0pt}{t}{0pt}{%
                \colorbox{gray}{%
                    \textcolor{white}{\textbf{\footnotesize (k)}}%
                }%
            }{\includegraphics[width=\linewidth]{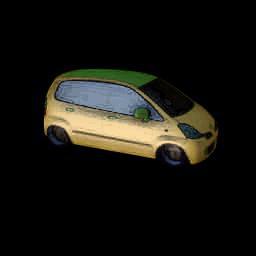}}}
        \end{minipage}
        \hfill
        \begin{minipage}[b]{0.137\linewidth}
            \centering
            \centerline{
            \stackinset{l}{0pt}{t}{0pt}{%
                \colorbox{gray}{%
                    \textcolor{white}{\textbf{\footnotesize (l)}}%
                }%
            }{\includegraphics[width=\linewidth]{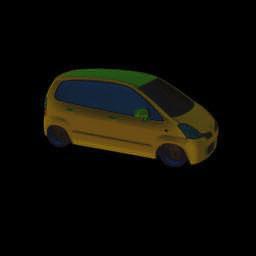}}}
        \end{minipage}
        \hfill
        \begin{minipage}[b]{0.137\linewidth}
            \centering
            \centerline{
            \stackinset{l}{0pt}{t}{0pt}{%
                \colorbox{gray}{%
                    \textcolor{white}{\textbf{\footnotesize (m)}}%
                }%
            }{\includegraphics[width=\linewidth]{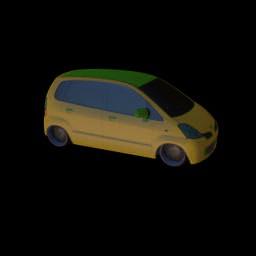}}}
        \end{minipage}   
        \hfill
        \begin{minipage}[b]{0.137\linewidth}
            \centering
            \centerline{
            \stackinset{l}{0pt}{t}{0pt}{%
                \colorbox{gray}{%
                    \textcolor{white}{\textbf{\footnotesize (n)}}%
                }%
            }{\includegraphics[width=\linewidth]{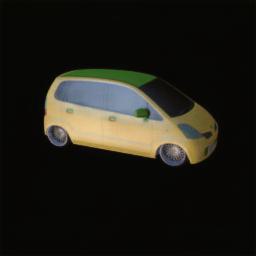}}}
        \end{minipage} 
    \end{minipage}

    \begin{minipage}[b]{1.0\linewidth}
        \begin{minipage}[b]{0.137\linewidth}
            \centering
            \centerline{
            \stackinset{l}{0pt}{t}{0pt}{%
                \colorbox{gray}{%
                    \textcolor{white}{\textbf{\footnotesize (o)}}%
                }%
            }{\includegraphics[width=\linewidth]{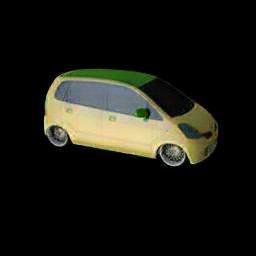}}}
        \end{minipage}   
        \hfill
        \begin{minipage}[b]{0.137\linewidth}
            \centering
            \centerline{
            \stackinset{l}{0pt}{t}{0pt}{%
                \colorbox{gray}{%
                    \textcolor{white}{\textbf{\footnotesize (p)}}%
                }%
            }{\includegraphics[width=\linewidth]{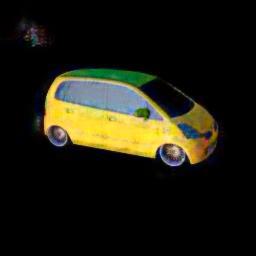}}}
        \end{minipage}   
        \hfill
        \begin{minipage}[b]{0.137\linewidth}
            \centering
            \centerline{
            \stackinset{l}{0pt}{t}{0pt}{%
                \colorbox{gray}{%
                    \textcolor{white}{\textbf{\footnotesize (q)}}%
                }%
            }{\includegraphics[width=\linewidth]{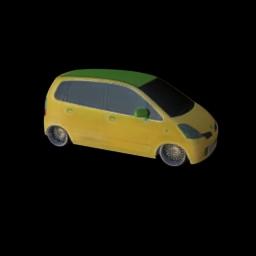}}}
        \end{minipage}
        \hfill
        \begin{minipage}[b]{0.137\linewidth}
            \centering
            \centerline{
            \stackinset{l}{0pt}{t}{0pt}{%
                \colorbox{gray}{%
                    \textcolor{white}{\textbf{\footnotesize (r)}}%
                }%
            }{\includegraphics[width=\linewidth]{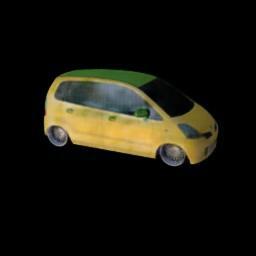}}}
        \end{minipage}
        \hfill
        \begin{minipage}[b]{0.137\linewidth}
            \centering
            \centerline{
            \stackinset{l}{0pt}{t}{0pt}{%
                \colorbox{gray}{%
                    \textcolor{white}{\textbf{\footnotesize (s)}}%
                }%
            }{\includegraphics[width=\linewidth]{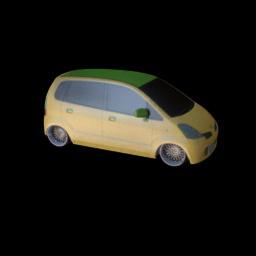}}}
        \end{minipage}
        \hfill
        \begin{minipage}[b]{0.137\linewidth}
            \centering
            \centerline{
            \stackinset{l}{0pt}{t}{0pt}{%
                \colorbox{gray}{%
                    \textcolor{white}{\textbf{\footnotesize (t)}}%
                }%
            }{\includegraphics[width=\linewidth]{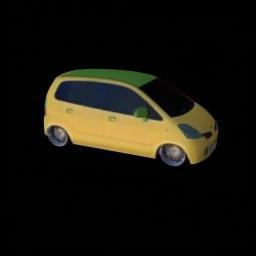}}}
        \end{minipage}
        \hfill
        \begin{minipage}[b]{0.137\linewidth}
            \centering
            \centerline{
            \stackinset{l}{0pt}{t}{0pt}{%
                \colorbox{gray}{%
                    \textcolor{white}{\textbf{\footnotesize (u)}}%
                }%
            }{\includegraphics[width=\linewidth]{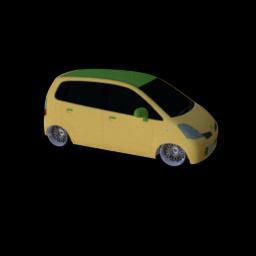}}}
        \end{minipage}
    \end{minipage}

    \caption{{\small Comprehensive visual comparison. (a) Input specular highlight image, (b) Tan [10], (c) Yoon [31], (d) Shen [11], (e) Shen [12], (f) Yang [13], (g) Shen [14], (h) Akashi [15], (i) Huo [32], (j) Fu [18], (k) Yamamoto [19], (l) Saha [20], (m) SLRR [22], (n) JSHDR [6], (o) SpecularityNet [5], (p) MG-CycleGAN [26], (q) Wu [25], (r) TSHRNet [7], (s) AHA [28], (t) Ours, (u) GT diffuse image. The reader is encouraged to zoom-in.}}
    \label{fig:full_compare4}
\end{figure*}

Figure \ref{fig:full_compare1}, \ref{fig:full_compare2}, \ref{fig:full_compare3}, and \ref{fig:full_compare4} provide a comprehensive visual comparison of all the methods discussed in the main paper, offering a more extensive performance evaluation. 
Overall, our method surpasses both traditional and deep learning specular highlight removal methods, excelling not only in effective highlight removal but also in the visual quality.

\section{User Study}
Although metrics such as Naturalness Image Quality Evaluator (NIQE) offers valuable insights into the visual quality of images, there remains a lack of unreferenced metrics specifically designed to evaluate the results of highlight removal. To address this gap and further validate the effectiveness of our DHAN-SHR in practical applications, we conducted a user study. This study aims to assess the perceptual quality of images processed by our method in comparison to other state-of-the-art techniques. It focuses not only on evaluating the effectiveness of highlight removal but also on the overall quality of the output images, providing a comprehensive analysis of our method's performance. 

Observing that deep learning methods significantly outperform traditional approaches, we limited participant evaluation to our method versus seven other learning-based methods to maintain the focus and reduce the burden on our study participants. We invited 20 participants to evaluate the visual quality of highlight removal images. To ensure a comprehensive assessment, we randomly selected 10 images from each of the three test sets (PSD, SHIQ, and SSHR), resulting in a total of 240 images for evaluation. 

We have meticulously prepared the user study form, as shown on the last page, in a Word document format and ensuring that there is no compression of the images. The images were organized into groups, each consisting of one original input image with specular highlights and eight corresponding highlight removal results, including our method and the seven other learning-based methods. The methods were anonymized to prevent bias, and the order of the methods within each group was randomized.

Participants were provided with a scoring table for each group of images, where they rated the eight methods based on the following criteria:
\begin{enumerate}
	\item \textbf{Highlight Reflection Area Detection Ability}: Assessing the effectiveness and accuracy of detecting highlight areas.
	\item \textbf{Highlight Removal Effect}: Evaluating the naturalness of highlight removal and the absence of color distortion.
	\item \textbf{Texture Restoration Level}: Assessing the consistency of texture in the highlight-removed area with nearby regions.
	\item \textbf{Diffuse Area Visual Quality}: Evaluating whether the diffuse areas were altered.
\end{enumerate}

The scoring scale ranged from 1 (worst) to 5 (best), allowing participants to capture a spectrum of perceptible quality levels in the highlight removal results:
\begin{itemize}
	\item 1 (Poor): The image significantly falls short in the specific criterion, marked by noticeable issues or distortions.
	\item 2 (Fair): The image, despite visible flaws, exhibits some elements of acceptable quality.
	\item 3 (Average): The image is satisfactory overall, with most elements adequately processed.
	\item 4 (Good): The image is well-processed, presenting only minor imperfections.
	\item 5 (Excellent): The image excels in the criterion, demonstrating exceptional quality.
\end{itemize}

Participants were instructed to assign a score for each criterion independently, ensuring a thorough evaluation of the various aspects of highlight removal. During the evaluation process, participants were able to zoom in on the images for a more detailed examination. For each de-highlighted image, we presented the original alongside the outputs from the eight methods, anonymizing the method names to prevent bias.

The final score for each image was determined by calculating the mean of the scores across the four criteria, with each criterion being equally weighted. This approach ensured a balanced and comprehensive assessment of each highlight removal result's overall quality. Table \ref{tab:userstudy} presents the final user study scores, illustrating that our method consistently achieves the highest average score across all three test sets.

% \usepackage[normalem]{ulem}
\begin{table}[]
\caption{Comparison of user study scores with seven learning-based methods. The highest-scored results are highlighted in bold, while the second-best are underlined for emphasis.}
\centering
\adjustbox{width=0.37\textwidth}{%
\begin{tabular}{lrrr}
\toprule
Method                 & PSD           & SHIQ          & SSHR          \\
\midrule
SLRR {[}22{]}          & 1.65          & 1.83          & 3.09          \\
JSHDR {[}6{]}          & 3.64          & \underline{4.8}     & 3.06          \\
SpecularityNet {[}5{]} & \underline{3.95}    & 3.96          & 3.72          \\
MG-CycleGAN {[}26{]}   & 3.26          & 3.21          & 2.46          \\
Wu {[}25{]}            & 3.65          & 3.95          & \underline{4.25}    \\
TSHRNet {[}7{]}        & \underline{3.95}    & 4.53          & 3.79          \\
AHA {[}28{]}           & 3.03          & 2.38          & 4.09          \\
Ours                   & \textbf{4.41} & \textbf{4.92} & \textbf{4.77}  \\
\bottomrule
\end{tabular}
}
\label{tab:userstudy}
\end{table}

\begin{figure*}[!htb]
\caption*{\Large\textbf{Specular Highlight Removal Methods Evaluation Form}}
\hrulefill

\textbf{Purpose:} This study aims to assess the effectiveness of various specular highlight removal methods. Your feedback will help improve the quality of specular highlight removal techniques.

\textbf{Instructions:}
\begin{itemize}
    \item You will see an original image with specular highlights followed by its processed versions.
    \item Please rate each processed image based on the criteria provided.
    \item Use the scale from 1 (Poor) to 5 (Excellent) for your rating.
\end{itemize}

\textbf{Evaluation Criteria:}
\begin{enumerate}
    \item \textbf{Highlight Reflection Area Detection Ability}: Assessing the effectiveness and accuracy of detecting highlight areas.
    \item \textbf{Highlight Removal Effect}: Evaluating the naturalness of highlight removal and the absence of color distortion.
    \item \textbf{Texture Restoration Level}: Assessing the consistency of texture in the highlight-removed area with nearby regions.
    \item \textbf{Diffuse Area Visual Quality}: Evaluating whether the diffusion areas were altered.
\end{enumerate}

\textbf{Image Evaluation:}
(**Below is a demonstration of one group of images to serve as an example for the evaluation process.)
\begin{minipage}[t]{0.68\linewidth}
        \centering
        \begin{minipage}[t]{0.245\linewidth}
            \centering            
            \centerline{\includegraphics[width=\linewidth]{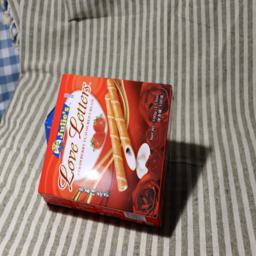}}
            \centerline{Input} \medskip
        \end{minipage}
    \end{minipage}

\begin{minipage}[t]{0.68\linewidth}
        \raggedleft
        \begin{minipage}[t]{0.24\linewidth}
            \centering
            \centerline{\includegraphics[width=\linewidth]{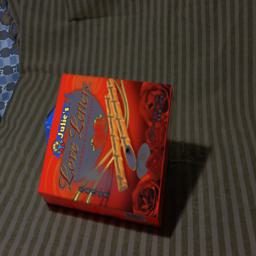}}
            \centerline{Method 1} \medskip
        \end{minipage}
        \hfill
        \begin{minipage}[t]{0.24\linewidth}
            \centering            \centerline{\includegraphics[width=\linewidth]{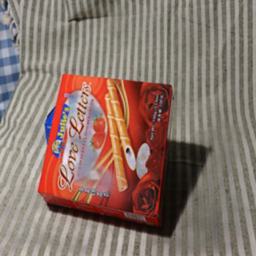}}
            \centerline{Method 2} \medskip
        \end{minipage}
        \hfill
        \begin{minipage}[t]{0.24\linewidth}
            \centering            \centerline{\includegraphics[width=\linewidth]{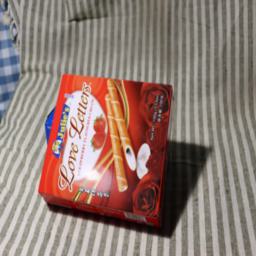}}
            \centerline{Method 3} \medskip
        \end{minipage}  
        \hfill
        \begin{minipage}[t]{0.24\linewidth}
            \centering            \centerline{\includegraphics[width=\linewidth]{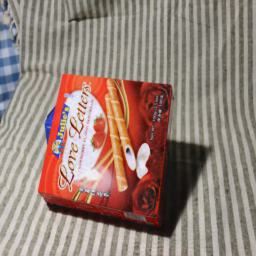}}
            \centerline{Method 4} \medskip
        \end{minipage} 
    \end{minipage}

\begin{minipage}[t]{0.68\linewidth}
        \raggedleft
        \begin{minipage}[t]{0.24\linewidth}
            \centering            \centerline{\includegraphics[width=\linewidth]{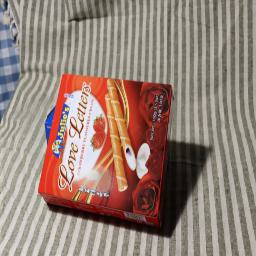}}
            \centerline{Method 5} \medskip
        \end{minipage}   
        \hfill
        \begin{minipage}[t]{0.24\linewidth}
            \centering
            \centerline{\includegraphics[width=\linewidth]{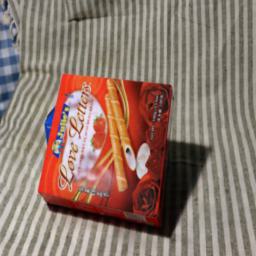}}
            \centerline{Method 6} \medskip
        \end{minipage}
        \hfill
        \begin{minipage}[t]{0.24\linewidth}
            \centering            \centerline{\includegraphics[width=\linewidth]{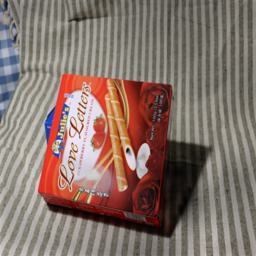}}
            \centerline{Method 7} \medskip
        \end{minipage}
        \hfill
        \begin{minipage}[t]{0.24\linewidth}
            \centering            \centerline{\includegraphics[width=\linewidth]{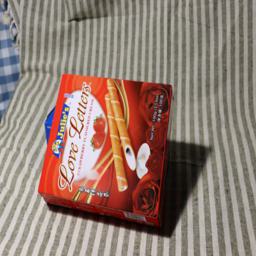}}
            \centerline{Method 8} \medskip
        \end{minipage}
    \end{minipage}
    
\vspace{0.2cm}

\adjustbox{width=0.9\textwidth}{%
        \begin{tabular}{|c|p{0.65cm}|p{0.65cm}|p{0.65cm}|p{0.65cm}|p{0.65cm}|p{0.65cm}|p{0.65cm}|p{0.65cm}|}
            \hline
            Method & \multicolumn{1}{c|}{1} & \multicolumn{1}{c|}{2} & \multicolumn{1}{c|}{3} & \multicolumn{1}{c|}{4} & \multicolumn{1}{c|}{5} & \multicolumn{1}{c|}{6} & \multicolumn{1}{c|}{7} & \multicolumn{1}{c|}{8} \\ \hline
            Highlight Detection Ability & & & & & & & &  \\
            \hline
            Highlight Removal Effect & & & & & & & &  \\
            \hline
            Texture Restoration Level & & & & & & & &  \\
            \hline
            Diffuse Area Visual Quality & & & & & & & &  \\
            \hline
        \end{tabular}    
    }
\vspace{0.2cm}  

{\Large Groups 2 to 30 have been omitted in this section for brevity.}

\vspace{0.8cm}
  
\hrulefill 

\textbf{Thank You Note:} Thank you for your participation. Your insights are invaluable to us.
\end{figure*}

\end{document}